\documentclass[twocolumn, final]{svjour3}
\usepackage[square,numbers]{natbib}
\usepackage{amsmath, epsfig}
\usepackage{amsfonts}
\usepackage{graphicx}
\usepackage{amsfonts}
\usepackage{algorithm}
\usepackage{algorithmic}
\usepackage{easybmat}
\usepackage{footmisc}
\usepackage[usenames,dvipsnames]{color}
\usepackage{subfig}

\newcommand{\ignore}[1]{}
\newcommand{\comment}[1]{}

\begin{document}


\title{Unsupervised Detection and Tracking of Arbitrary Objects with Dependent Dirichlet Process Mixtures}
\titlerunning{Unsupervised Detection and Tracking of Arbitrary Objects with Dependent Dirichlet Process Mixtures}
\author{Willie Neiswanger$^{*}$ and Frank Wood$^{\dagger}$}
\authorrunning{Willie Neiswanger and Frank Wood}
\institute{ $^{*}$Columbia University, Department of Applied Math and Applied Physics. Tel.: 503-464-6152. \email{wdn2101@columbia.edu} \and $^{\dagger}$Columbia University, Department of Statistics. Tel.: 212-851-2150. \email{fwood@stat.columbia.edu}}
\date{}  
\maketitle


\begin{abstract}
This paper proposes a technique for the unsupervised detection and tracking of arbitrary objects in videos. It is intended to reduce the need for detection and localization methods tailored to specific object types and serve as a general framework applicable to videos with varied objects, backgrounds, and image qualities. The technique uses a dependent Dirichlet process mixture (DDPM) known as the Generalized Polya Urn (GPUDDPM) to model image pixel data that can be easily and efficiently extracted from the regions in a video that represent objects. This paper describes a specific implementation of the model using spatial and color pixel data extracted via frame differencing and gives two algorithms for performing inference in the model to accomplish detection and tracking. This technique is demonstrated on multiple synthetic and benchmark video datasets that illustrate its ability to, without modification, detect and track objects with diverse physical characteristics moving over non-uniform backgrounds and through occlusion.
\end{abstract}


\section{Introduction}
\label{sec:introduction}

We define unsupervised detection and tracking of arbitrary objects in videos to be the task of automatically identifying the distinct objects present in a sequence of images and determining the path each object follows over time. Techniques that accomplish this task are useful in many fields that make use of video data, including robotics, video surveillance, time-lapse microscopy, and video summarization. By studying this task, we hope to help make progress towards general machine vision algorithms that can learn the positions, appearances, and number of objects present in any video scene.

This task can be broken down into three parts: data extraction, localization, and tracking. Data extraction is the act of extracting features from regions of video-frames that constitute objects, localization is the act of finding the positions and/or shapes of distinct objects, and tracking is the act of maintaining the identities of the detected objects over time. This paper introduces a new framework for carrying out these three actions based on a type of dependent Dirichlet process mixture model. This framework provides a foundation for a class of unsupervised algorithms that can detect and track arbitrary objects in a wide range of videos.

Research related to general detection and tracking of objects tends to focus on one of either extraction, localization, or tracking. Integrating all three tasks in a system for multiple arbitrary objects and diverse video types is not often a primary focus. A few attempts at accomplishing the three tasks in a cohesive manner have been studied in recent years \cite{brostow2006unsupervised, brox2010object, fragkiadaki2011detection, pece_2002}.
This paper furthers this line of work by providing a model that could give rise to a number of algorithms to detect arbitrary objects in videos---particularly in cases where frame-by-frame segmentation is difficult, video quality is low, and extraction is noisy---and maintain the isolation of distinct objects during tracking and through occlusion.

We begin by describing characteristics of the extracted data (Section~\ref{sec:dataextraction}), and giving the generalized form of the model (Section~\ref{sec:modeldefinition}). To implement this model, one must specify a data extraction procedure and distributions for representing objects, which may be chosen to allow for arbitrary object tracking or tailored to a specific object type for a given application. In our implementation, we extract data via a basic frame differencing procedure and specify distributions useful for representing arbitrary objects (Section~\ref{sec:modelspecification}). We describe inference algorithms for our model and show how the output of these algorithms can be interpreted as object localization and tracking results (Section~\ref{sec:inference}). Our implementation is demonstrated on multiple synthetic and benchmark datasets. Standard performance metric values are computed to quantify results on the benchmark datasets (Section~\ref{sec:experiments}). We compare the performance metrics of our method with those yielded by specialized detection and tracking algorithms tailored to specific objects in the benchmark datasets. Our results support our hypothesis that approaches combining simple data extraction and a powerful model can perform detection and tracking of arbitrary objects at a level comparable to state-of-the-art, object specific algorithms.


\section{Background}
\label{sec:priorwork}

A variety of methods in the fields of image processing, signal processing, and computer vision have been developed to solve aspects of the problem of unsupervised detection and tracking of arbitrary objects. These methods might be placed into a few broad categories: those that aim to distinguish the foreground regions of images from the background \cite{hong2007real,chien2002efficient, zhang2007moving, kim2002fast}, segment images into distinct regions to perform localization \cite{jain1997object, fei2005bayesian, sivic2005discovering}, track an object over a sequence of images (after its position has been specified in an initial image) \cite{raja_1998, mckenna_1999, jepson_2003, comaniciu_2003, perez_2002}, track multiple objects over a sequence of images (especially when the objects interact or occlude one another) \cite{senior2006appearance, cucchiara2004probabilistic, zhou2003background, han_2004, mckenna2000tracking, dockstader2001multiple}, segment a sequence of images into distinct spatiotemporal regions \cite{brox2003unsupervised, sista2000unsupervised, wang1998unsupervised}, and combine the previous methods in some way to create systems capable of both detecting and tracking specified objects \cite{Okuma04aboosted,eth_biwi_00633,4036928, khan_2004, leibe2008coupled}, or of discerning which regions of a video constitute distinct, arbitrary objects and tracking these \cite{brostow2006unsupervised, brox2010object, fragkiadaki2011detection, pece_2002,paragios2000geodesic}.

Methods that discern between the foreground and background regions of a video allow for data to be extracted from the areas in each frame where objects reside. Frame differencing and background subtraction are two such methods. Both record locations that exhibit motion relative to the background. Often, background subtraction refers to methods that compare an image containing targets with an image of the background only or with some model of the background that is learned as the video progresses \cite{piccardi2004background}, while frame differencing refers to methods that compare pairs of consecutive images in a video \cite{zhang2001segmentation}. Frame differencing has been used as the sole extraction method for object localization or tracking schemes with success \citep{pece_2002, beleznai_2006, chu_2007}, and also as a secondary data extraction method to help improve the accuracy of object tracking \citep{perez_2002}.

A great deal of research has focused on developing algorithms to track multiple objects simultaneously. There has been a particular emphasis on developing ways to deal with problems such as object occlusion (where one object blocks another from the view of a video camera) \cite{senior2006appearance, cucchiara2004probabilistic, zhou2003background}, complex object interactions \cite{khan_2004, mckenna2000tracking, dockstader2001multiple}, objects with similar appearances \cite{maccormick1999probabilistic, jepson_2003}, variable (and potentially high) numbers of objects \cite{reilly2010detection}, and objects that enter and exit a field of view at different times \cite{stauffer2003estimating, nedrich2010learning}. Multiple independent single-object trackers running simultaneously have been shown to be ineffective, as they will tend to coalesce and track the same object \cite{khan_2004}. To remedy this problem, methods have incorporated probabilistic principles for maintaining isolation of object trackers \cite{maccormick1999probabilistic}. An approach to this problem involving the use of a nonparametric mixture model has also found success in maintaining isolation of distinct objects \cite{vermaak_2003}.

Over the past decade, there have been attempts to provide general algorithms for the fully unsupervised detection and tracking of arbitrary objects in videos. Blob tracking, a basic method for carrying out this goal, has found success in videos where objects are easily isolated from the background and where localization and segmentation of distinct objects is possible \cite{francois2004real, isard_2001}. Blob tracking methods, however, run into problems when faced with videos where detection is difficult, object appearance or orientation varies heavily, and there exists object occlusion \cite{song2005model}. To improve the accuracy of these methods, techniques have been developed for performing extraction and segmentation in a joint manner, incorporating statistical methods for maintaining hypotheses of different numbers of detected objects, and introducing some of the multi-object tracking methods described previously to track distinct blobs after they have been segmented \cite{collins2003mean, isard_2001}. Another family of methods related to the task of unsupervised detection and tracking of video objects goes under the heading of video segmentation algorithms---these methods extend single-frame image segmentation to maintain coherence of image segments over time, and have had some success when used for the explicit purpose of detecting and tracking foreground objects in videos \cite{brox2003unsupervised, sista2000unsupervised, wang1998unsupervised}. Other attempts to perform unsupervised detection and tracking include methods for clustering short sequences of positions extracted by detecting the motion of objects \cite{brostow2006unsupervised, brox2010object}, which aim to return full-length distinct object tracks, and graph based methods that carry out a similar task using spectral clustering \cite{fragkiadaki2011detection}. Another approach uses a Gaussian mixture model to cluster data extracted from moving objects \cite{pece_2002}; this method also develops heuristics for the initialization and elimination of new object tracks. 

Nearly all high accuracy object detection and tracking methods are those tailored for specific object types. These methods rely on detection criteria that exploit knowledge about the appearance or behavior of the objects in a video. Some of these methods make use of state-of-the-art detectors designed to locate the specified objects of interest. In contrast, the method described in this paper is designed to track arbitrary objects without using any explicit detection criteria, and serve as a general strategy that can be used, without modification, to perform accurate detection and tracking of diverse objects in a wide range of videos. The technique we introduce falls into the category of clustering-based arbitrary object detection and tracking methods. Differing from previous work, we use a type of time dependent Bayesian nonparametric mixture model, and show how it can be applied to a variety of easily extracted data to perform detection and tracking. This method begins by performing a simple data extraction procedure that yields noisy data. Our model of this data serves as a general framework for which we can choose a variety of object appearance distributions and inference algorithms; each choice provides a new method for unsupervised detection and tracking of arbitrary objects in videos.


\begin{figure*}
  \centering
  \subfloat[]{\label{fig:ants_img}\includegraphics[width=0.33\textwidth]{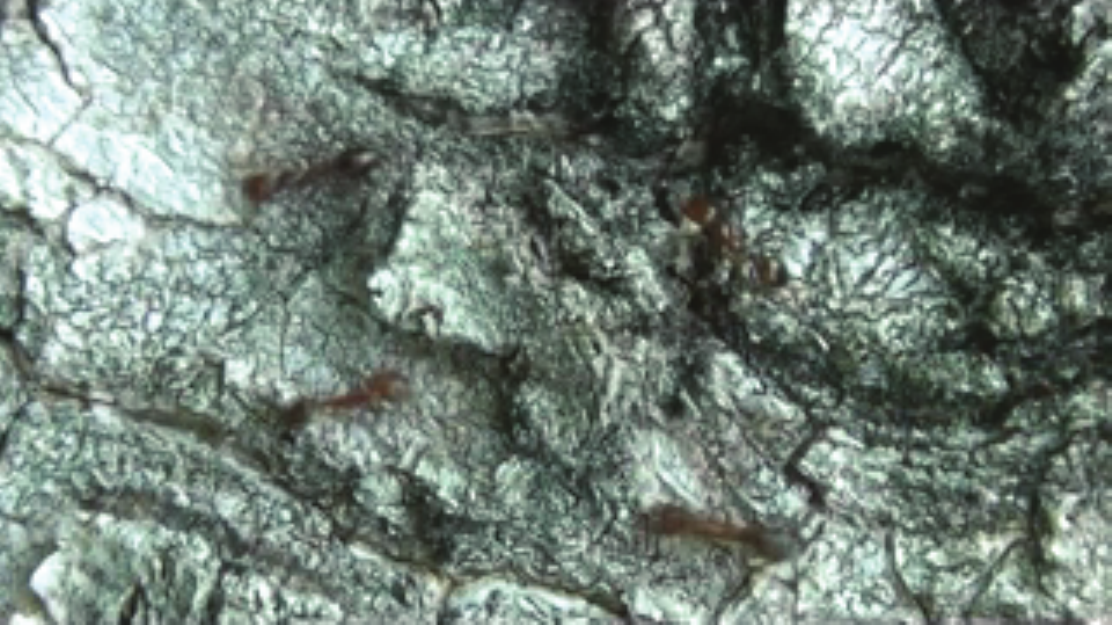}} \hspace{0.1mm}
  \subfloat[]{\label{fig:ants_img2}\includegraphics[width=0.33\textwidth]{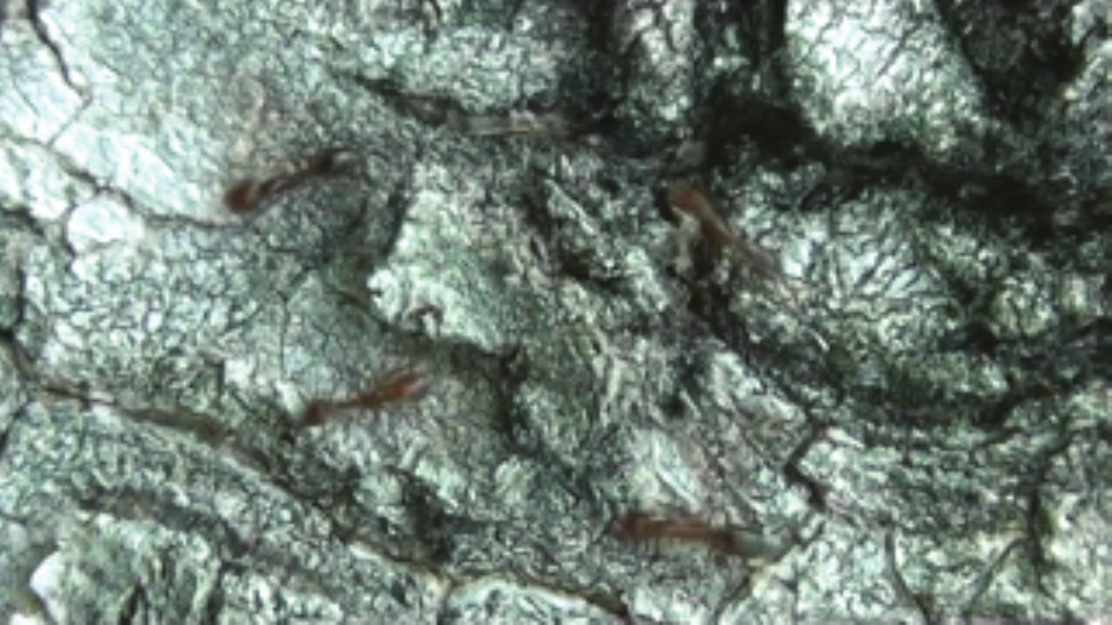}} \hspace{0.1mm}
  \subfloat[]{\label{fig:ants_img_framediff}\includegraphics[width=0.323\textwidth]{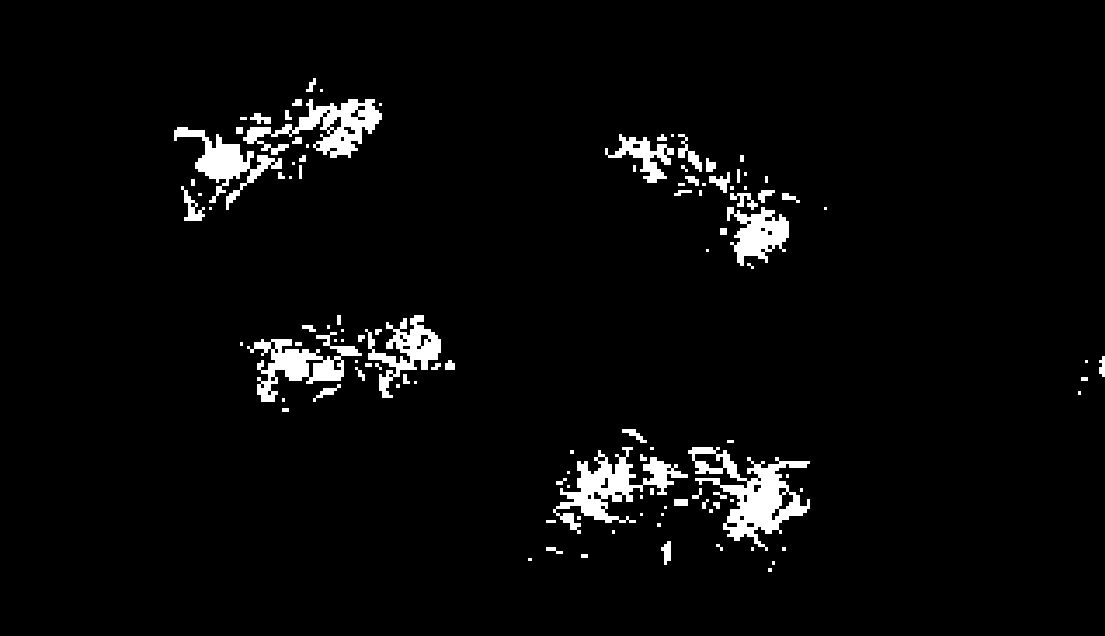}}\\
  \subfloat[]{\label{fig:traffic2_img}\includegraphics[width=0.328\textwidth]{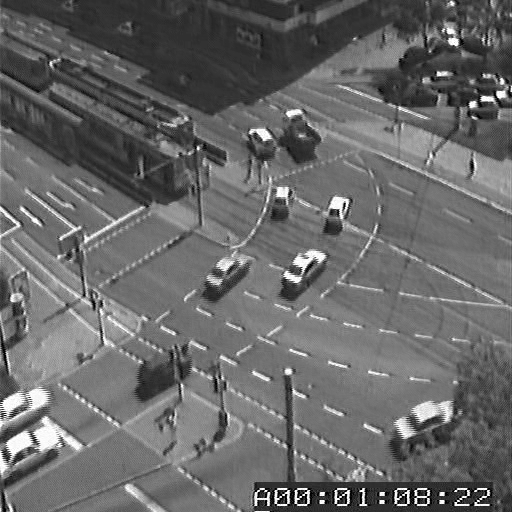}} \hspace{0.1mm}
  \subfloat[]{\label{fig:traffic2_img2}\includegraphics[width=0.328\textwidth]{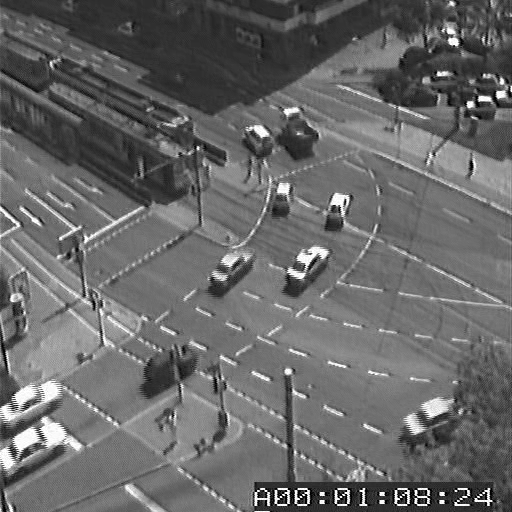}} \hspace{0.1mm}
  \subfloat[]{\label{fig:traffic2_img_framediff}\includegraphics[width=0.3273\textwidth]{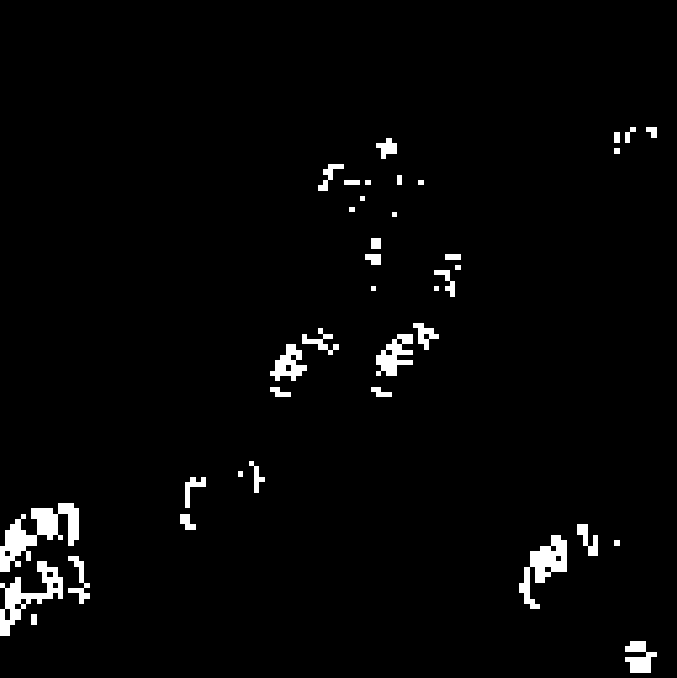}}\\
  \subfloat[]{\label{fig:pets2009_img}\includegraphics[width=0.328\textwidth]{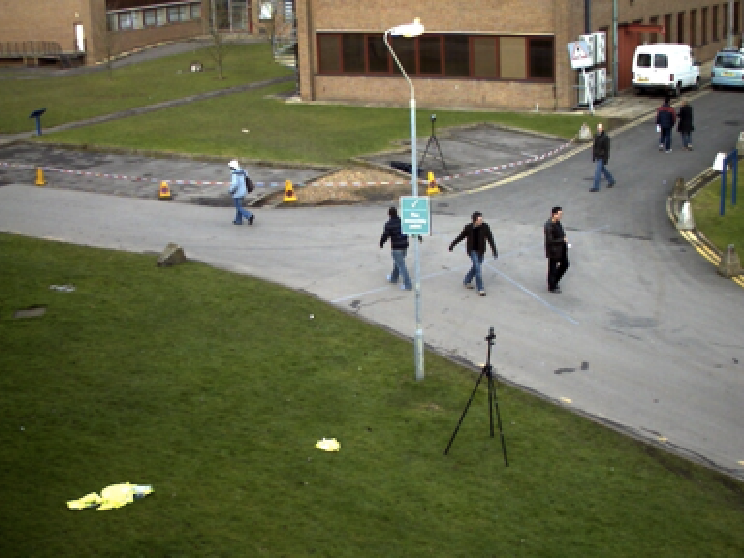}} \hspace{0.1mm}
  \subfloat[]{\label{fig:pets2009_img2}\includegraphics[width=0.328\textwidth]{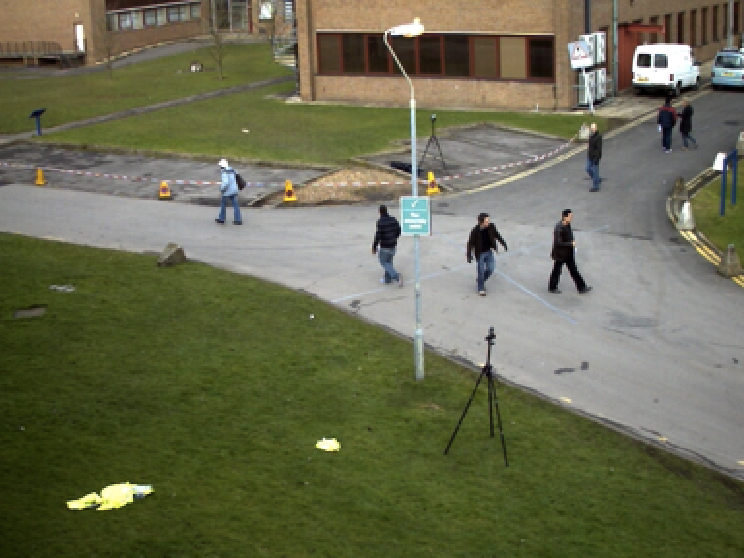}} \hspace{0.1mm}
  \subfloat[]{\label{fig:pets2009_img_framediff}\includegraphics[width=0.328\textwidth]{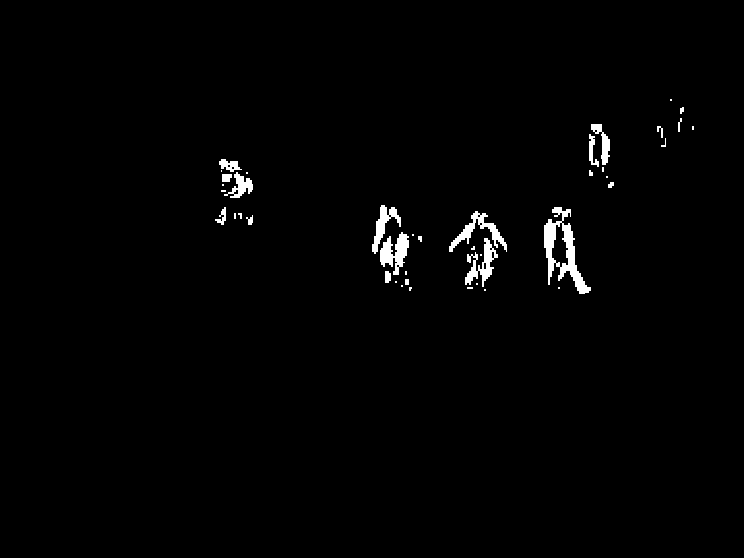}}\\
  \subfloat[]{\label{fig:longimg_all}\includegraphics[width=1\textwidth]{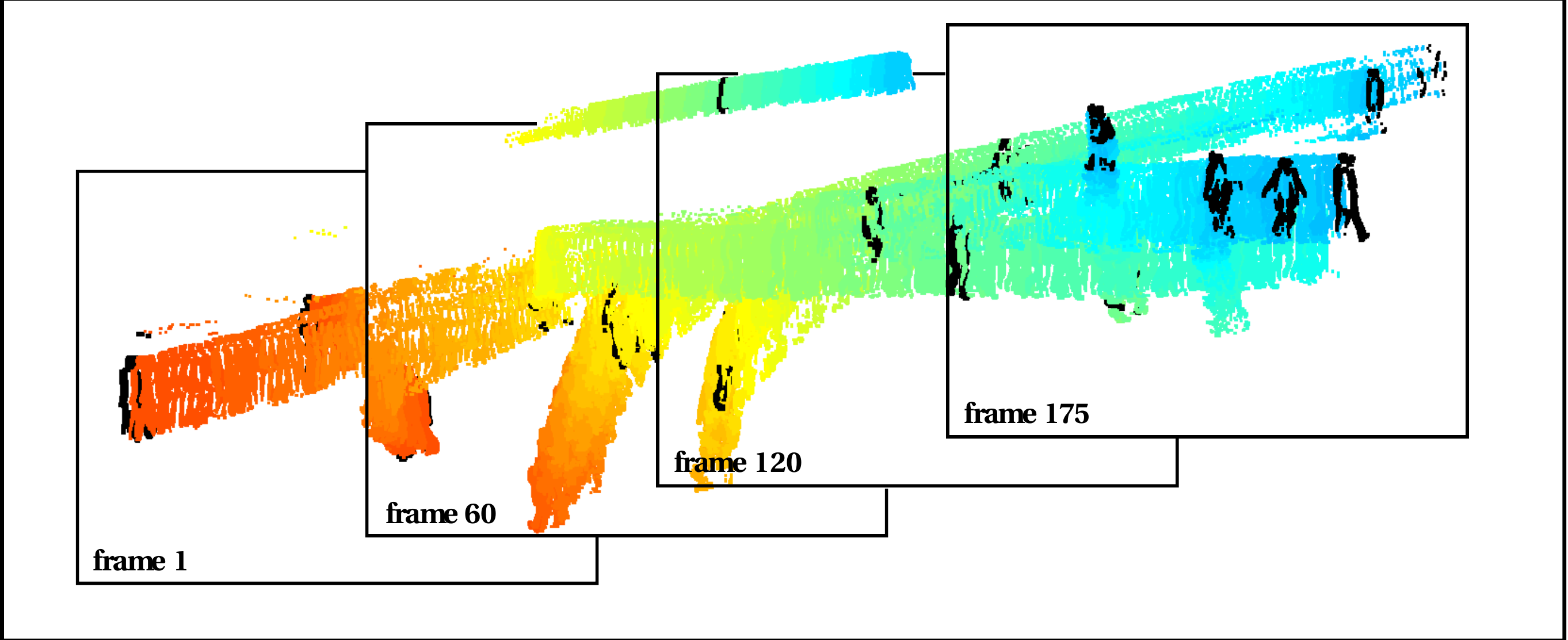}}
  \caption{Three pairs of consecutive frames and the results produced by taking the pixel-wise frame difference between each pair (a - i). The final image shows the results of frame differencing over a sequence of images (from the PETS2009/2010 dataset.}
  \label{fig:img_and_framediff}
\end{figure*}


\section{Data Extraction}
\label{sec:dataextraction}

We desire a data extraction procedure that yields observations of the form
\begin{equation}
\centering
\bold{x} = ( \bold{x}^{s}, \bold{x}^{c}, t ) = ( x^{s_{1}}, x^{s_{2}}, x^{c_{1}}, \ldots, x^{c_{V}}, t )
\end{equation}
where each $\bold{x}$ corresponds to a point within an image region where an object (or foreground element)  is believed to reside, $\bold{x}^{s} \in \mathbb{R}^{2}$ denotes the spatial location of this point, $\bold{x}^{c} \in S_{1} \times \ldots \times S_{V}$ denotes some collection of local image features in the vicinity of this point, and $t \in \{1, \ldots, T \}$ denotes the time index.

We'd like to use an extraction procedure that is as unsophisticated as possible. Consequently, we use frame differencing. This procedure locates the pixels in image regions that undergo change. Specifically, at each frame the pixels that differ from the previous frame beyond some threshold are recorded. Here, each pixel corresponds to an observation $\bold{x}$. Frame differencing is simple, computationally inexpensive, and able to be applied to a wide range of static, single-camera videos (the videos used in experiments are stationary; moving-camera videos require data extraction methods useful for non-stationary video \cite{chien2002efficient, zhang2007moving}). Examples of pixel location data extracted via frame differencing are shown in Figure~\ref{fig:img_and_framediff}(a-j).

We also extract features $\bold{x}^{c}$ that capture image information in the vicinity of each extracted pixel. Examples of possible features include color distributions, pixel intensity values, feature point (such as corner, shape, or edge) locations or spatial characteristics, and texture representations. In principle, we can extract any image features that may be used to characterize the appearance of objects. In our implementation, we choose to extract only color information in the vicinity of each pixel. Incorporating color features allows our method to infer a distribution over color for each detected object; this improves its ability to distinguish between adjacent objects and track objects through occlusion. To add this information, we let $\bold{x}^{c}$ represent a $V$ dimensional discrete distribution over some aspect of color (such as the hue) in the pixel's immediate vicinity. Details on this vector and how it is computed are given in Section~\ref{sec:dataextractioninexperiments}. We refer to the components of this vector as the ``color counts'' of the pixel.


\section{Model Framework}
\label{sec:modeldefinition}

We use a type of dependent Dirichlet process mixture (DDPM) model known as the Generalized Polya Urn dependent Dirichlet process mixture (GPUDDPM). We define a general form of this model in Section~\ref{sec:gpuddpm}, and specify the distributions used in our implementation of this model in Section~\ref{sec:modelspecification}. We also define a secondary form of this model in Section~\ref{sec:MCMC}, which is used in one of the two inference algorithms. We provide a brief introduction to mixture models and the Dirichlet process in Appendix~\ref{sec:modelbackground}.

Dirichlet process mixture (DPM) models (Section \ref{sec:dpmixture}) fall under the heading of Bayesian nonparametric models. These models have been widely used in the past decade to perform nonparametric density estimation and cluster analysis. Data extracted from videos comprises spatiotemporal clusters, each corresponding to a distinct object. Consequently, we are interested in using a class of models known as dependent Dirichlet process mixture (DDPM) models (Section~\ref{sec:ddpmixture}), which are particularly useful for estimating the number of latent classes (clusters) in time dependent data.

Since objects can enter and exit a scene, the number of clusters present throughout the video may not be constant (i.e., clusters may be created, or be ``born'', and may disappear, or ``die'', at intermediate time steps). To cluster data with these properties, we choose to use a DDPM known as the Generalized Polya Urn dependent Dirichlet process mixture \cite{caron_2007}. This model may be viewed intuitively as a sequence of DPMs, where there exist dependencies between the parameters and number of clusters in adjacent time steps. Like the DPM, the GPUDDPM allows for a distribution over the number of clusters within a dataset---which, in this work, corresponds to the number of objects in a video---to be inferred.


\subsection{Generalized Polya Urn Dependent Dirichlet Process Mixture Model}
\label{sec:gpuddpm}

In a GPUDDPM, each observation $\bold{x}_{i,t}$ is associated with an assignment variable $c_{i,t}$ that represents its assignment to a cluster $\theta_{k,t}$.
The sizes of clusters in the GPUDDPM increase when observations are assigned to them, and decrease at later time points when these observations become ``unassigned''. We define a distribution over the size of cluster $k$ at time $t$ ($m_{k,t}$), conditioned on the cluster's previous size ($m_{k,t-1}$), the assignments at time $t$ ($c_{1:N_{t}, t}$), and a deletion parameter ($\rho$),
\begin{equation}
\begin{split}
\label{D_distro}
\text{D} &(m_{k,t} | m_{k,t-1}, \hspace{1mm} c_{1:N_{t}, t}, \hspace{1mm} \rho) := \hspace{2mm} \text{Binomial}( m_{k,t-1}  \\
&- m_{k,t} + \sum_{i=1}^{N_{t}} \mathbb{I}(c_{i,t} = k) | \hspace{2mm} {m_{k,t-1}, \hspace{1mm} \rho})
\end{split}
\end{equation}
$\forall k \in \{1, \ldots, K_{t} \}$, where $K_{t}$ is the number of clusters at time $t$ and $\mathbb{I}(c_{i,t} = k)$ is an indicator function whose value is 1 if $c_{i,t} = k$ and 0 otherwise.

We also define a distribution over the assignment of observation $i$ at time $t$ ($c_{i,t}$), conditioned on the sizes of all clusters at time $t$ ($m_{1:K_{t},t}$) and a concentration parameter ($\alpha$),
\begin{equation}
\begin{split}
\label{C_distro}
\text{C} ( c_{i,t} = k | &m_{1:K_{t},t}, \hspace{1mm} \alpha)\\ :=
& \begin{cases}
\frac{m_{k,t}}{\sum_{k=1}^{K_{t}} m_{k,t} + \alpha}, \hspace{1mm} \text{if} \hspace{1mm} k \in \{ 1, \ldots, K_{t} \} \\
\frac{\alpha}{\sum_{k=1}^{K_{t}} m_{k,t} + \alpha}, \hspace{1mm} \text{if} \hspace{1mm} k = K_{t} + 1
\end{cases}
\end{split}
\end{equation}
$\forall i \in \{1, \ldots, N_{t} \}$, where there exists $K_{t}$ clusters at time $t$ and we give a newly created cluster the index $K_{t+1}$.

\begin{figure}[h]
        \center{\includegraphics[width=80mm]{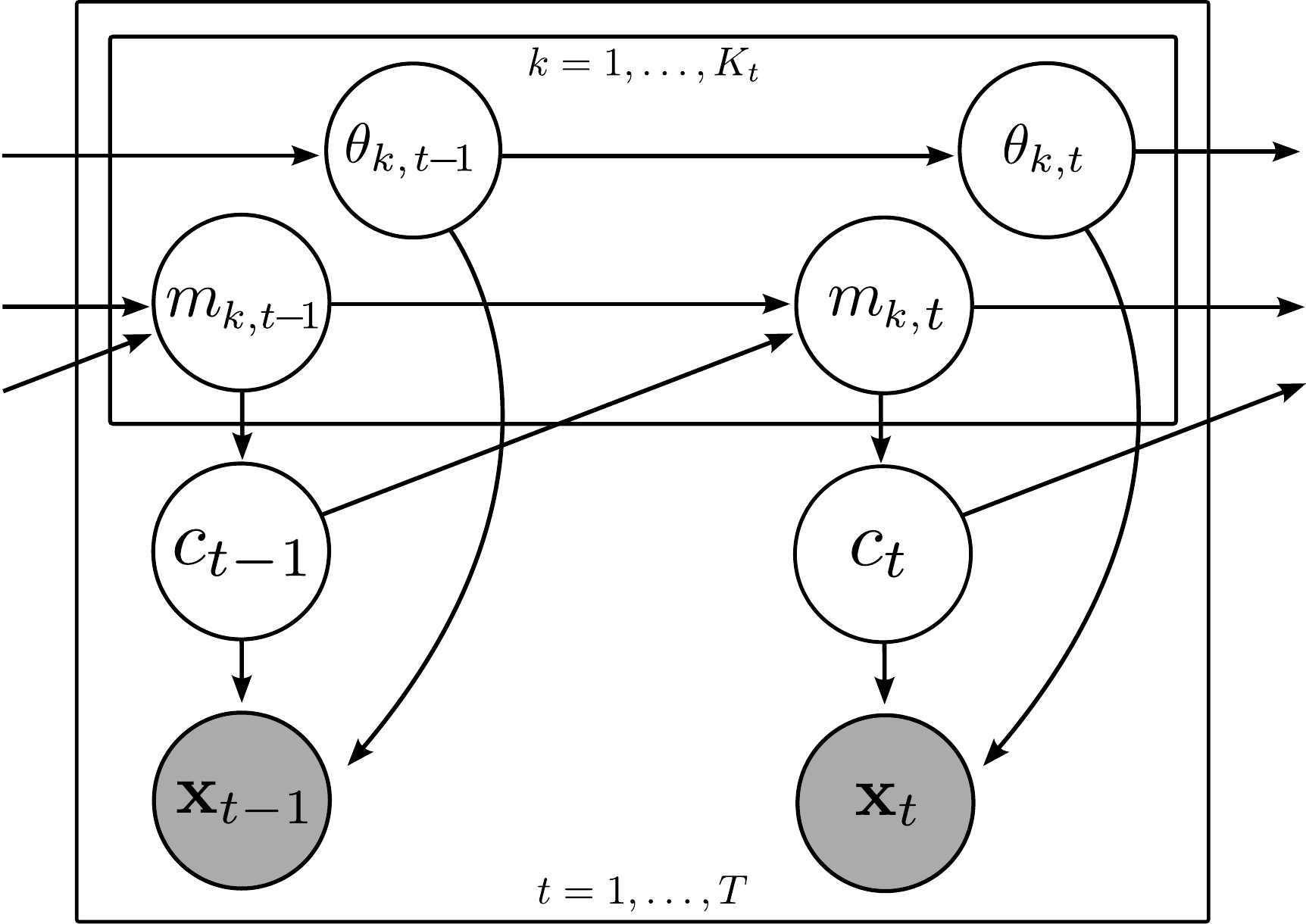}}
        \caption{\label{fig:gpuddpm_gm_1} Graphical Model of the Generalized Polya Urn dependent Dirichlet process mixture. The observations at time $t$, $\bold{x}_{1:N_{t},t}$, and their associated assignments $c_{1:N_{t},t}$ are denoted respectively as $\bold{x}_{t}$ and $c_{t}$ (likewise for those at time $t-1$).}
\end{figure}

Distributions \eqref{D_distro} and \eqref{C_distro} together comprise what is referred to as the ``Generalized Polya Urn'' \cite{caron_2007}. We can now define the GPUDDPM generatively as
\begin{align}
\label{gpuddpm_def}
\begin{split}
m_{k,t} | m_{k,t-1}, \hspace{1mm} c_{1:N_{t}, t}, \hspace{1mm} \rho  &\sim \text{D}(m_{k,t-1}, \hspace{1mm} c_{1:N_{t}, t}, \hspace{1mm} \rho) \\
\theta_{k, t} | \theta_{k, t-1}   & \sim
\begin{cases}
  P(\theta_{k, t} | \theta_{k, t-1}) \hspace{1mm} \text{if} \hspace{1mm} k \leq K_{t} \\ 
  \mathbb{G}_{0} \hspace{1mm} \text{if} \hspace{1mm} k = K_{t+1}
\end{cases} \\
c_{i,t} | m_{1:K_{t},t}, \hspace{1mm} \alpha  &\sim  \text{C} (m_{1:K_{t},t}, \hspace{1mm} \alpha) \\
\bold{x}_{i,t} | c_{i,t}, \theta_{1:K_{t}, t} &\sim \text{F}(\theta_{c_{i,t}, t})
\end{split}
\end{align}
$\forall$ times $t \in \{1, \ldots, T\}$ and each cluster $k \in \{ 1, \ldots, K_{t} \}$ at time $t$, where we choose application-specific distributions for F, $\mathbb{G}_{0}$, and $P(\theta_{k, t} | \theta_{k, t-1})$ in Section~\ref{sec:modelspecification}. The graphical model associated with this formulation of the GPUDDPM is shown in Figure~\ref{fig:gpuddpm_gm_1}.


\section{Model Specifics}
\label{sec:modelspecification}

Sections~\ref{sec:objectappearance}-\ref{sec:motionmodel} detail the object representation distributions chosen to fully specify our implementation of the GPUDDPM. This specification is used for all experiments in Section~\ref{sec:experiments}. The distributions F, $\mathbb{G}_{0}$, and $P(\theta_{k, t} | \theta_{k, t-1})$ represent object appearance, the appearance prior, and object movement, respectively. Our specification is kept general to allow for wide applicability, though one could choose to incorporate known object appearance or motion information for a specific tracking application in future studies.

\subsection{Object Appearance and Mixture Component, F}
\label{sec:objectappearance}

At a given time $t$, we model each observation $\bold{x} \in \bold{X}$ as a draw from the product of a multivariate normal and multinomial distribution
\begin{equation}
\label{likelihood}
\text{F}(\bold{x}|\theta) = \mathcal{N}(\bold{x}^{s} | \boldsymbol{\mu}, \Sigma)  \mathcal{M}n(\bold{x}^{c} | \bold{p})
\end{equation}
where $\theta = \{ \boldsymbol{\mu}, \Sigma, \bold{p} \}$ denotes the parameters of a cluster at time $t$, with mean $\boldsymbol{\mu} \in \mathbb{R}^{2}$, covariance matrix $\Sigma \in \mathbb{R}^{2\times2}$, and discrete probability vector $\bold{p} = (p_{1}, \ldots, p_{V})$ such that $\sum_{i=1}^{V}p_{i} = 1$. Additionally, $\mathcal{N}$ denotes the multivariate normal distribution and $\mathcal{M}n$ denotes the multinomial distribution. Note that we withhold writing the subscripts specifying the cluster number k and time t in this section when writing them is unnecessary.

The multivariate normal distribution over the spatial features $\bold{x}^{s}$ models the position and spatial extent of an object. It can be thought intuitively to represent the shape of each object as an oval. We model the color features $\bold{x}^{c}$ as draws from a multinomial distribution. Incorporating this distribution into our cluster likelihood allows us to exploit our observation that the pixels associated with distinct objects tend to have similar color count vectors.

\subsection{Appearance Prior and Base Distribution, $\mathbb{G}_{0}$}
\label{sec:appearanceprior}

$\mathbb{G}_{0}$ denotes the base distribution of the DDPM; it also serves as a prior distribution for the parameters $\theta = \{ \boldsymbol{\mu}, \Sigma, \bold{p} \}$ of the mixture components (i.e. of the object appearance distributions). We use conjugate priors in the base distribution to allow for more efficient computation. Specifically, in our implementation, a normal-inverse-Wishart prior is placed on the multivariate normal parameters $\{ \boldsymbol{\mu}, \Sigma \}$, and a Dirichlet prior is placed on the multinomial parameter $\bold{p}$. The prior can therefore be written
\begin{equation}
\label{basedistro}
\mathbb{G}_{0}(\theta) = \mathcal{N}i\mathcal{W}(\boldsymbol{\mu}, \Sigma | \boldsymbol{\mu}_{0}, \kappa_{0}, \nu_{0}, \Lambda_{0})  \mathcal{D}ir( \bold{p} | \bold{q}_{0})
\end{equation}
where $\mathcal{N}i\mathcal{W}$ denotes the normal-inverse-Wishart distribution, $\mathcal{D}ir$ denotes the Dirichlet distribution, and the prior has the hyperparameters $\boldsymbol{\mu}_{0}, \kappa_{0}, \nu_{0}, \Lambda_{0}$ and $\bold{q}_{0}$.

\subsection{Motion Model and Transition Kernel, $P(\theta_{t} | \theta_{t-1})$}
\label{sec:motionmodel}

The transition kernel $P(\theta_{t} | \theta_{t-1})$ represents how we expect tracked objects to move over time. Since our implementation is intended for tracking arbitrary objects, we do not wish to make sophisticated assumptions about object motion. For example, we choose not to incorporate complex objects dynamics, though they are often used with success in certain object-specific tracking tasks, such as people tracking \cite{choo2001people, conte2010performance}. We assume only that the position of an object at a given time is close to its position at the previous time, and that the position varies in all directions equally between time steps.

The base distribution $\mathbb{G}_{0}$ must be the invariant distribution of the transition kernel $P(\theta_{t} | \theta_{t-1})$ in order for the the cluster parameters to remain marginally distributed according to the base distribution, and for the model to be a valid GPUDDPM. In other words, the transition kernel must satisfy
\begin{equation}
\int \mathbb{G}_{0}(\theta_{t-1})P(\theta_{t} | \theta_{t-1}) d\theta_{t-1} = \mathbb{G}_{0}(\theta_{t})
\end{equation}
for a given cluster with parameters $\theta$. One way to achieve this is through the use of auxiliary variables. These are a set of $M$ variables $\bold{z}_{t} = (z_{t,1}, \ldots, z_{t,M})$ associated with each cluster at each time $t$ that satisfy
\begin{eqnarray}
\label{auxiliaryvariablecriteria}
P(\theta_{t} | \theta_{t-1}) = \int P(\theta_{t} | \bold{z}_{t}) P(\bold{z}_{t} | \theta_{t-1}) d \bold{z}_{t}
\end{eqnarray}

With the addition of these variables, the parameters of a cluster at a given time do not depend directly on their value at the previous time; they are instead dependent through an intermediate sequence of variables. This allows the cluster parameters at each time step to be marginally distributed according to the base distribution $\mathbb{G}_{0}$ while maintaining simple time varying behavior.

Each of the auxiliary variables $z_{t,m}$ is drawn from the product of a multivariate normal and multinomial with the associated cluster parameters $\theta_{t} = \{ \boldsymbol{\mu}_{t}, \Sigma_{t}, \bold{p}_{t} \}$
\begin{equation}
\label{transker1}
z_{t, m} | \boldsymbol{\mu}_{t}, \Sigma_{t}, \bold{p}_{t}  \sim  \mathcal{N}(\boldsymbol{\mu}_{t}, \Sigma_{t}) \mathcal{M}n(\bold{p}_{t})   \hspace{15pt}   
\end{equation}
$\forall m \in \{ 1, \ldots, M \}$. To satisfy \eqref{auxiliaryvariablecriteria}, we specify the dependencies of a given cluster on its associated set of auxiliary variables at each time $t$ by
\begin{equation}
\label{transker2}
\boldsymbol{\mu}_{t}, \Sigma_{t}, \bold{p}_{t} | \bold{z}_t  \sim  \mathcal{N}i\mathcal{W}(\boldsymbol{\mu}_{M}, \kappa_{M}, \nu_{M}, \Lambda_{M})  \mathcal{D}ir(\bold{q}_{M})
\end{equation}
where $\boldsymbol{\mu}_{M}, \kappa_{M}, \nu_{M}, \Lambda_{M},$ and $\bold{q}_{M}$ are
\begin{eqnarray}
\kappa_{M} &=& \kappa_{0} + M \\
\nu_{M} &=& \nu_{0} + M \\
\boldsymbol{\mu}_{M} &=& \frac{\kappa_{0}}{\kappa_{0}+M} \boldsymbol{\mu}_{0}  +  \frac{M}{\kappa_{0}+M} \overline{\bold{z}_t}^{s}\\
\Lambda_{M} &=& \Lambda_{0} + S_{\bold{z}_{t}^{s}}\\
\bold{q}_{M} &=& \bold{q}_{0} + \sum_{m=1}^{M} z_{t,m}^{c}
\end{eqnarray}
and where $M$ is the number of auxiliary variables, $\{ \boldsymbol{\mu}_{0},$ $\kappa_{0}$, $\nu_{0}$, $\Lambda_{0} \}$ are the $\mathcal{N}i\mathcal{W}$ prior parameters, and $\bold{q}_{0}$ is the $\mathcal{D}ir$ prior parameter. We use $\bold{z}^{s}$ and $\bold{z}^{c}$ to respectively denote the spatial and color features of an auxiliary variable $\bold{z}$, and $\overline{\bold{z}}$ and $S_{\bold{z}}$ to respectively denote the sample mean and sample covariance for a set $\bold{z} = \{ z_{1}, \ldots, z_{M} \}$ (of auxiliary variables, in this case), which we can write as
\begin{eqnarray}
\label{samplemean}
\overline{\bold{z}}  &=&  \left( \sum_{m=1}^{M} z_{m} \right) / M\\
\label{samplecov}
S_{\bold{z}}  &=&  \sum_{m=1}^{M} (z_{m} - \overline{\bold{z}}) (z_{m} - \overline{\bold{z}})^{T}
\end{eqnarray}

\subsection{Recap of Model Parameters}
\label{sec:recapofparameters}

The multivariate normal-multinomial GPUDDPM for object tracking has a number of parameters, which are used to specify the object appearance prior distribution, transition kernel, and Generalized Polya Urn distributions (C and D). 
The object appearance prior parameters include:
\begin{center}
\begin{tabular}[c]{| p{2cm} |  p{5.5cm} | }
\hline
$\boldsymbol{\mu}_{0} \in \mathbb{R}^{2}$  &  Mean position prior. In experiments performed in Section~\ref{sec:experiments} the data was recentered to the origin, and this parameter was set to $(0,0)$. \\ \hline
$\kappa_{0} \in \mathbb{R}$  &  Scale factor of the mean prior.\\ \hline
$\Lambda_{0} \in \mathbb{R}^{2 \times 2}$  &  Shape factor of the covariance prior.\\ \hline
$\nu_{0} \in \mathbb{Z}_{+} > 1$  &  Scale factor of the covariance prior.\\ \hline
$\bold{q}_{0} \in \mathbb{R}_{+}^{V}$  &  Scale factor of the multinomial prior. \\
\hline
\end{tabular}
\end{center} \vspace{3mm}
The following parameter dictates characteristics of object movement.
\begin{center}
\begin{tabular}[c]{ | p{2cm} | p{5.5cm} | }
\hline
$M \in \mathbb{Z}_{+}$  &  Number of auxiliary variables. A larger number will produce a smoother object path.\\
\hline
\end{tabular}
\end{center} \vspace{3mm}
Additionally, one can tune the model's tendency to detect new objects and maintain the existence of these objects (both dictated by distributions C and D) with the following parameters.
\begin{center}
\begin{tabular}[c]{ | p{2cm} | p{5.5cm} | }
\hline
$\alpha \in \mathbb{R}_{+}$  &  The concentration parameter for the Dirichlet process. A higher value will increase the tendency for new objects to be detected.\\ \hline
$\rho \in ( 0, 1 ]$  &  The deletion parameter. A higher value will give objects an increased tendency to die off. \\
\hline
\end{tabular}
\end{center}


\section{Inference}
\label{sec:inference}

Bayesian inference is used to achieve detection and tracking results. Previously developed inference strategies can be applied to the generative model defined in Section~\ref{sec:gpuddpm} and Section~\ref{sec:modelspecification}. We provide details on the two Bayesian inference algorithms implemented in this stu-\\dy. The first is a type of Markov Chain Monte Carlo (MCMC) batch inference, which uses Gibbs sampling to generate samples from the posterior distribution of the model. The second is a type of Sequential Monte Carlo (SMC) inference, also known as a particle filter, which generates samples from the posterior distribution of the model in a sequential manner.

\subsection{MCMC: Batch Inference}
\label{sec:MCMC}

This section details the MCMC sampler used to perform inference.
A secondary formulation of the GPUDDPM, which we refer to as the ``Deletion Variable Formulation'' (defined in Section~\ref{sec:deletionvariableformulation}), is used here. This formulation is equivalent to the formulation given in Section~\ref{sec:gpuddpm}, but allows for easier sampling.

\subsubsection{Deletion Variable Formulation}
\label{sec:deletionvariableformulation}

Instead of incorporating the cluster size variable $m_{k,t}$ directly in the GPUDDPM model (as it was in the definition given by \eqref{gpuddpm_def}), we can formulate an equivalent model which makes use of new set of variables called deletion variables. We introduce a deletion variable $d_{i,t}$ for each observation $\bold{x}_{i,t}$, which denotes the time at which the observation is removed from its assigned cluster. At each time, cluster sizes $m_{k,t}$ can be reconstructed from all previous assignments and deletion variables by
\begin{equation}
\label{compute_clust_size}
m_{k,t} = \sum_{t' = 1}^{t} \mathbb{I}[(c_{t'}=k) \wedge (t < d_{t'})]
\end{equation}
where $\mathbb{I}[\cdot]$ is an indicator function that evaluates to 1 if its argument is true, and 0 otherwise. Additionally, we can define the deletion variable $d_{i,t}$ to be $d_{i,t} = t + l_{i,t}$, where $l_{i,t}$ can be thought of as the lifetime of an assignment. From the definition of distribution D (given by \eqref{D_distro}), the lifetime can be shown to be distributed geometrically, and can be written as
\begin{equation}
\label{del_rho_form}
l_{i,t} | \rho  \sim  \rho(1 - \rho)^{l_{i,t}}
\end{equation}
where the parameter $\rho$ is the same as that in \eqref{D_distro}.

\begin{figure}[h]
        \center{\includegraphics[width=82mm]{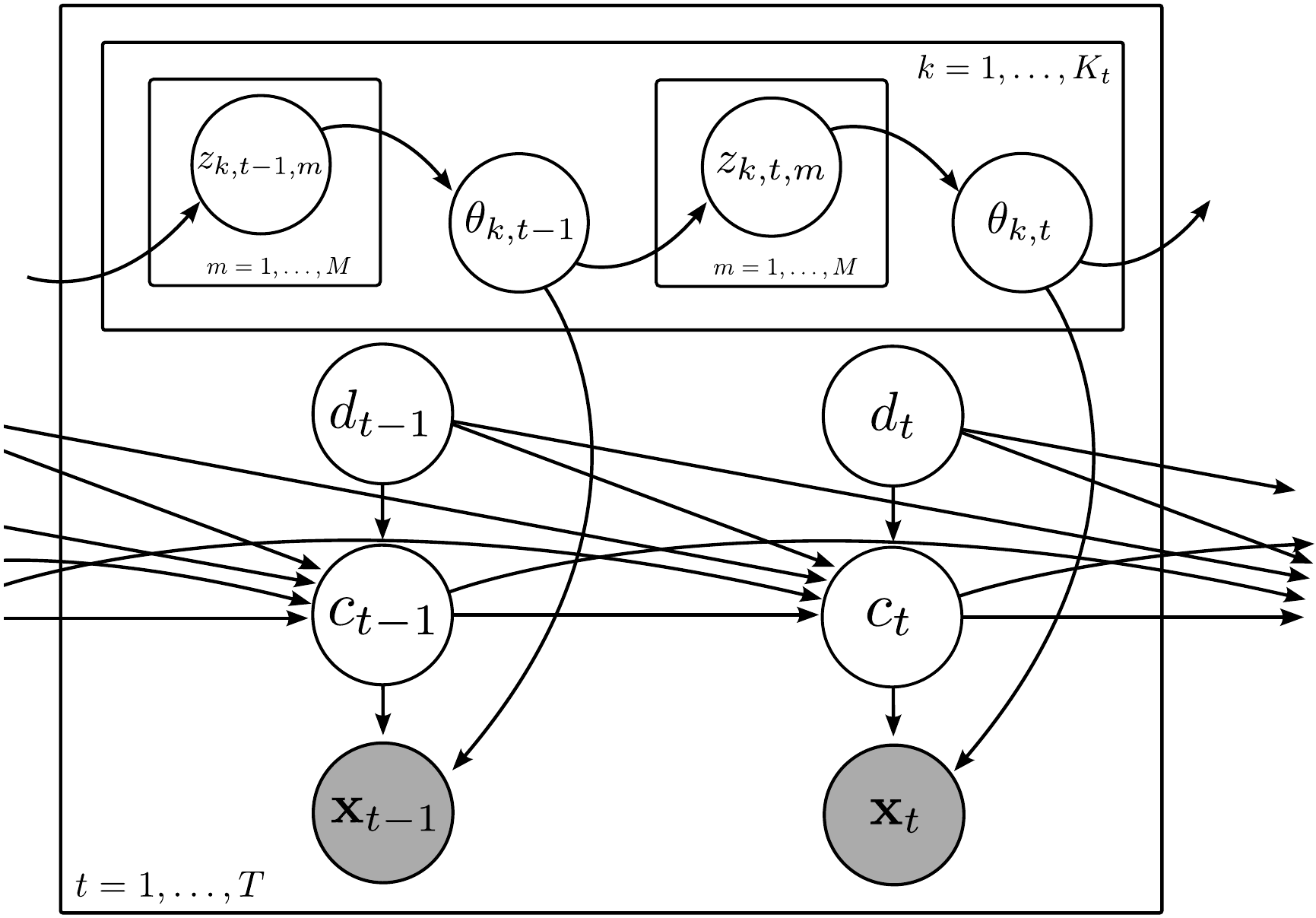}}
        \caption{Graphical model of the Deletion Variable Formulation of the GPUDDPM, showing auxiliary variables.}
        \label{fig:gpuddpm_gm_2}
\end{figure}

We can now define the Deletion Variable Formulation of the GPUDDPM generatively as
\begin{align}
\begin{split}
d_{i,t} | \rho  &\sim \text{Geo}(\rho) + t + 1 \\
\theta_{k, t} | \theta_{k, t-1}  &\sim
\begin{cases}
P(\theta_{k,t} | \theta_{k,t-1}) \hspace{1mm} \text{if} \hspace{1mm} k \leq K_{t} \\ 
\mathbb{G}_{0} \hspace{1mm} \text{if} \hspace{1mm} k = K_{t} + 1
\end{cases} \\
c_{i,t} | \bold{c}_{1:t-1}, \bold{d}_{1:t-1}, \alpha  &\sim  \text{C}(\bold{c}_{1:t-1}, \bold{d}_{1:t-1}, \alpha) \\
\bold{x}_{i,t} | c_{i,t}, \theta_{c_{i,t},t}  &\sim  \text{F}(\theta_{c_{i,t}, t})
\end{split}
\end{align}
$\forall$ times $t \in \{1, \ldots, T\}$ and clusters $k \in \{ 1, \ldots, K_{t} \} $ at time $t$,
where $\bold{c}_{t} = c_{1:N_{t},t}$, $\bold{d}_{t} = d_{1:N_{t},t}$, and the distributions F, $\mathbb{G}_{0}$, and $P(\theta_{k, t} | \theta_{k, t-1})$ are described in Section~\ref{sec:modelspecification}. This formulation of the GPUDDPM is also used by \cite{gasthaus_thesis} and \cite{caron_2007}. The associated graphical model for this formulation is given in Figure~\ref{fig:gpuddpm_gm_2}.

For easier notation, we define $\boldsymbol{x}_{t} = \bold{x}_{1:N_{t},t}$, $\bold{c}_{t} = c_{1:N_{t},t}$, $\bold{d}_{t} = d_{1:N_{t},t}, \Theta_{t} = \theta_{1:K_{t}, t}$, and $\bold{z}_{t} = z_{1:K_{t},t,1:M}$.
At each time $t$, the sampler moves sequentially through the $N_{t}$ observations, sampling the assignment $c_{i,t}$ and the deletion variable $d_{i,t}$ for each. Afterwards, the cluster parameters for all active clusters at $t$ are sampled, and the $M$ auxiliary variables for all active clusters at $t$ are sampled using Metropolis Hastings (MH). The following sections detail the distributions from which each of these samples is drawn.


\subsubsection{Sampling Assignment Variables, $c_{i,t}$}
\label{sec:sample_assignments}

A value proportional to the posterior probability can be computed for each possible value that $c_{i,t}$ may take on. These values allow us to construct a discrete probability distribution from which we can draw samples from the posterior distribution over assignments. The possible values that the $c_{i,t}$ may take on are $k \in \{ 1 , \ldots ,  K_{t}^{'}+1 \}$, where $K_{t}^{'}$ denotes the number clusters with a non-zero size at any time $t' \in \{ t, \ldots, d_{i,t} \}$, and $K_{t}'+1$ denotes a ``new'' cluster. The probability that $x_{i,t}$ is assigned to cluster $k$, i.e. that $c_{i,t}=k$, given values for all other variables in the model (which we denote as ``$\ldots$"), is given by
\begin{equation}
\label{mcmc_assig_vars}
\begin{split}
p(c_{i,t} = k | \ldots ) & \propto
\prod_{i'=i}^{N_{t}}  \text{C}(c_{i',t} | \bold{m}_{t}, \alpha) \\
& \times \prod_{t' = t+1}^{d_{i,t}}  \prod_{i'=1}^{N_{t'}}   \text{C}(c_{i',t'} | \bold{m}_{t'}, \alpha) \\
 & \times
\begin{cases}
  F(\bold{x}_{i,t} | \theta_{k,t}) \hspace{2mm} \text{if} \hspace{1mm} k \leq K_{t}' \\ 
  \int P(\bold{x}_{i,t} | \theta) \mathbb{G}_{0}(\theta) d\theta \hspace{2mm} \text{if} \hspace{1mm}  k = K_{t}'+1
\end{cases}
\end{split}
\end{equation}
where C is given by \eqref{C_distro}, and the cluster sizes $\bold{m}_{t'}$ are calculated under the assumption that $c_{i,t} = k$. Note that the above integral has an analytic solution for our specific model
\begin{equation}
\begin{split}
\label{marginal_obs_prob}
\int & P(\bold{x}_{i} | \theta) \mathbb{G}_{0}(\theta)d\theta \\
&= t_{\nu_{0} - 1}  \left ( \bold{x}_{i}^{s}  \hspace{1mm} \big{|} \hspace{1mm} \boldsymbol{\mu}_{0}, \frac{\Lambda_{0}(\kappa_{0}+1)}{\kappa_{0}(\nu_{0}-1)} \right ) \\
& \times \prod_{v=1}^{V} \frac{ \Gamma(\bold{x}_{i}^{c}) }  { \Gamma(\bold{q}_{0})}
\times \frac{\Gamma(\sum_{j=1}^{V} \bold{q}_{0}) }{\Gamma(\sum_{j=1}^{V} \bold{x}_{i}^{c}) }
\end{split}
\end{equation}
where t denotes the multivariate t-distribution, where we follow the three-value parameterization (location parameter, scale parameter, and degrees of freedom) given in \cite{gelman2004bayesian, Wood2008nonparametric}, and $\{ \boldsymbol{\mu}_{0},$ $\kappa_{0},$ $\Lambda_{0},$ $\nu_{0},$ $\bold{q}_{0} \}$ are prior parameters.

If a new cluster is sampled as an assignment, the cluster parameters and auxiliary variables for this new cluster must be initialized for all time steps before sampling can proceed. In our implementation, newly sampled clusters were initialized by iteratively sampling forward to time T and backwards to time 1 via the transition kernel.


\subsubsection{Sampling Cluster Parameters, $\theta_{k, t}$}

The conjugacy of appearance model and transition kernel distributions allow us to easily sample from the posterior distribution over the cluster parameters, which we can write
\begin{align}
\begin{split}
P(\theta_{k,t} | \ldots ) = & \hspace{1mm} P(\bold{x}_{i,t} | \theta_{k,t}) P(z_{k,t+1,1:M} | \theta_{k,t})
\\ & \times P(\theta_{k,t} | z_{k,t,1:M}) \\
 = & \hspace{1mm} \mathcal{N}i\mathcal{W}(\boldsymbol{\mu}_{k,t}, \Sigma_{k,t} | \boldsymbol{\mu}_{N}, k_{N}, v_{N}, \Lambda_{N}) \\
  & \times \mathcal{D}ir( \bold{p}_{k,t} | \bold{q}_{N})
\end{split}
\end{align}
Where the parameters in the above distribution are given when the observations $\bold{x}_{1:N_{t},t}$, and auxiliary variables $z_{k,t-1:t,1:M}$ for cluster k at time $t-1$ and $t$, are taken to be the ``observations'' for the following Bayesian updates
\begin{eqnarray}
\label{bayesian_update_1}
\kappa_{N} &=& \kappa_{0} + N \\
\label{bayesian_update_2}
\nu_{N} &=& \nu_{0} + N \\
\label{bayesian_update_3}
\boldsymbol{\mu}_{N} &=& \frac{\kappa_{0}}{\kappa_{0}+N} \boldsymbol{\mu}_{0}  +  \frac{N}{\kappa_{0}+N} \overline{\bold{x}}^{s}\\
\Lambda_{N} &=& \Lambda_{0} + S_{\bold{x}^{s}}\\
\label{bayesian_update_4}
\bold{q}_{N} &=& \bold{q}_{0} + \sum_{i=1}^{N} \bold{x}_{i}^{c}
\end{eqnarray}
where $N$ is the number of observations, $\{ \boldsymbol{\mu}_{0}, \kappa_{0}, \nu_{0}, \Lambda_{0} \}$ are the $\mathcal{N}i\mathcal{W}$ prior parameters, $\bold{q}_{0}$ is the $\mathcal{D}ir$ prior parameter, $\bold{x}^{s}$ and $\bold{x}^{c}$ respectively denote the spatial and color features of the observations, and $\overline{\bold{x}}$ and $S_{\bold{x}}$ respectively denote the sample mean and sample covariance for the set of observations $\bold{x}$, defined in \eqref{samplemean} and \eqref{samplecov}.


\subsubsection{Sampling Auxiliary Variables, $z_{k,t,m}$}
\label{sec:sample_aux_vars}

The posterior distribution over each of the auxiliary variables $z_{k,t,m}$ can be written as
\begin{equation}
\label{stationary_pdf}
P(z_{k,t,m} | \ldots) \propto  P(z_{k,t,m} | \theta_{k,t-1}) P(\theta_{k,t} | z_{k,t,1:M})
\end{equation}
We sample a new value for all $M$ auxiliary variables, denoted $z_{k,t,m}^{*}$, using MH, with the proposal distribution
\begin{align}
\begin{split}
\label{proposal_distro}
z_{k,t,m}^{*}  &\sim  P(z_{k,t,m} | \theta_{k,t}) \\
&= \mathcal{N}(z_{k,t,m}^{s} | \boldsymbol{\mu}_{k,t}, \Sigma_{k,t}) \mathcal{M}n(z_{k,t,m}^{c} | \bold{p}_{k,t})
\end{split}
\end{align}
and compute the standard MH acceptance ratio, which in this case simplifies to 
\begin{equation}
\label{accept_ratio}
r_{accept} = \frac{P(\theta_{k,t-1} | z_{k,t,m}^{*})}{P(\theta_{k,t-1} | z_{k,t,m})}
\end{equation}


\subsubsection{Sampling Deletion Variables, $d_{i,t}$}
Sampling deletion variables could be performed in a manner similar to how we sample the assignment variables in Section~\ref{sec:sample_assignments}, but this may be computationally expensive due to the large number of possible deletion times. To remedy this, the MH algorithm may again be used to generate samples $d_{i,t}^{*}$ from the posterior distribution over possible deletion times, where we use the proposal distribution
\begin{align}
\begin{split}
l_{i,t}  &\sim  \text{Geo}(\rho)  \\
d_{i,t}^{*}  &= l_{i,t} + t + 1
\end{split}
\end{align}
where $l_{i,t}$ denotes the geometrically distributed ``lifetime'', and we accept or reject this sample using the process described in Section~\ref{sec:sample_aux_vars}.

\subsection{SMC: Sequential Inference}
\label{sec:SMC}

This section details a Sequential Monte Carlo (SMC) sampler---also known as a particle filter---used to perform inference. SMC inference operates in the orginal GPUDDPM formulation (Section~\ref{sec:gpuddpm}). The algorithm is shown in Algorithm~\ref{alg:SMC}. At each time step $t \in \{ 1, \ldots, T \}$, a number of samples referred to as ``particles'' are generated; each particle consists of a sample from the posterior distribution over the assignment for each observation, $c_{1,t}, \ldots, c_{N_{t}, t}$, parameters for each cluster $\theta_{1,t}, \ldots,$ $\theta_{K_{t}, t}$, and size after deletion for each cluster $m_{1,t},$ $\ldots,$ $m_{K_{t},t}$. A set of particles is sampled at each time step from relevant proposal distributions (described in Sections~\ref{sec:smc_proposal_1}, \ref{sec:smc_proposal_2}, and \ref{sec:smc_proposal_3}), a weight is computed for each particle, and a new set of particles are sampled from the set of weighted particles via a resampling process. Within each time step, Gibbs sampling is used to generate the samples associated with each particle.

The sequence of target distributions for the SMC algorithm may be written as
\begin{equation}
\begin{split}
\pi_{t}(\bold{c}_{1:t}, \Theta_{1:t}, \bold{m}_{1:t}) & =  \pi_{t-1}(\bold{c}_{1:t-1}, \Theta_{1:t-1}, \bold{m}_{1:t-1})\\
& \times \prod_{i=1}^{N_{t}} P(c_{i,t} | \bold{m}_{t}, \Theta_{t}, \bold{c}_{1:t}, \bold{x}_{1:N_{t}}) \\
& \times \prod_{k=1}^{K_{t}} 
\begin{cases}
\begin{split}
P(\theta_{k,t} | \theta_{k,t-1}) \hspace{2mm} & \text{if} \hspace{2mm} k \leq K_{t-1} \\
\mathbb{G}_{0} \hspace{18mm} & \text{if} \hspace{2mm} k > K_{t-1}
\end{split}
\end{cases} \\
& \times \prod_{k=1}^{K_{t}} \text{D}(m_{k,t} | m_{k,t-1}, c_{1:N_{t-1}, t-1}, \rho)
\end{split}
\end{equation}
where $\bold{c}_{t} = c_{1:N_{t},t}$, $\Theta_{t} = \theta_{1:K_{t},t}$, and $\bold{m}_{t} = m_{1:K_{t},t} $, and D is given by \eqref{D_distro}.

\begin{algorithm*}[!]
\caption{Sequential Monte Carlo Inference for the GPUDDPM}
\label{alg:SMC}
\begin{algorithmic}[1]
\FOR{$l = 1 : L$}
\STATE $w_{0}^{(l)} \leftarrow 1/L$ \hfill $\triangleright$ initialize weights
\ENDFOR
\STATE $K_{0}^{(l)} \leftarrow 0$ \hfill $\triangleright$ initialize \# of clusters
\FOR{$t = 1 : T  \text{  (\# of frames})$}
\FOR{$l = 1 : L  \text{  (\# of particles})$}
\STATE $K_{t}^{(l)}  \leftarrow K_{t-1}^{(l)}$
\STATE $m_{1:K_{t}^{(l)}, t}^{(l)} \leftarrow m_{1:K_{t-1}^{(l)}, t-1}^{(l)}$
\FOR{$s = 1 : S \text{  (\# of Gibbs samples})$}
\FOR{$i = 1 : N_{t} \text{  (\# of observations at frame $t$})$}
\IF{$s=1$}
\STATE Sample $c_{i,t}^{(l)} \sim P \left( c_{i,t}^{(l)} | m_{1:K_{t-1}^{(l)}, t-1}^{(l)}, \theta_{1:K_{t-1}^{(l)}, t-1}^{(l)}, \alpha \right)$  \hfill $\triangleright$ eq.~\eqref{smc_assig_eqn}
\STATE $m_{c_{i,t}^{(l)},t}^{(l)} \leftarrow m_{c_{i,t}^{(l)},t}^{(l)} + 1$
\ELSE
\STATE $m_{c_{i,t}^{(l)},t}^{(l)} \leftarrow m_{c_{i,t}^{(l)},t}^{(l)} - 1$
\STATE Sample $c_{i,t}^{(l)} \sim P \left( c_{i,t}^{(l)} | m_{1:K_{t}^{(l)}, t}^{(l)}, \theta_{1:K_{t}^{(l)}, t}^{(l)}, \alpha \right)$  \hfill $\triangleright$ eq.~\eqref{smc_assig_eqn}
\STATE $m_{c_{i,t}^{(l)},t}^{(l)} \leftarrow m_{c_{i,t}^{(l)},t}^{(l)} + 1$
\ENDIF
\IF{$c_{i,t}^{(l)} = K_{t}^{(l)} + 1  \text{  (a new cluster)}$}
\STATE Sample $\theta_{c_{i,t}^{(l)}, t}^{(l)} \sim q_{1}(\bold{x}_{i,t})$ \hfill $\triangleright$ eq.~\eqref{smc_q1_eqn}
\STATE $K_{t}^{(l)} \leftarrow K_{t}^{(l)} + 1$
\STATE $m_{K_{t}^{(l)},t}^{(l)} \leftarrow 1$
\ENDIF
\ENDFOR
\FOR{$k = 1 : K_{t}^{(l)}  \text{  (\# of clusters at frame $t$})$}
\IF{$k > K_{t-1}^{(l)}$}
\STATE Sample $\theta_{k,t}^{(l)} \sim q_{1}(\{ \bold{x}_{1:N_{t}, t} = k \})$  \hfill $\triangleright$ eq.~\eqref{smc_q1_eqn}
\ELSIF{$k \leq K_{t-1}^{(l)}$ and $ \#\{ \bold{x}_{1:N_{t}, t} = k \} > 0$}
\STATE Sample $\theta_{k,t}^{(l)} \sim q_{2}(\theta_{k,t-1}^{(l)}, \{ \bold{x}_{1:N_{t}, t} = k \})$  \hfill $\triangleright$ eq.~\eqref{smc_q2_eqn}
\ELSIF{$m_{k,t}^{(l)} > 0$}
\STATE Sample $\theta_{k,t}^{(l)} \sim P(\theta_{k,t}^{(l)} | \theta_{k,t-1}^{(l)})$  \hfill $\triangleright$ eqs.~\eqref{transker1} $\&$ \eqref{transker2}
\ENDIF
\IF{$s=S$}
\STATE Sample $m_{k,t+1}^{(l)} \sim \text{D}(m_{k,t}^{(l)}, c_{1:N_{t}, t}^{(l)}, \rho)$  \hfill $\triangleright$ eq.~\eqref{D_distro}
\ENDIF
\ENDFOR
\ENDFOR
\STATE $\tilde{w}_{t}^{(l)} \leftarrow w_{t-1}^{(l)} \times \frac{ P(\bold{x}_{1:N_{t},t}, c_{1:N_{t}, t}^{(l)} | \theta_{1:K_{t}^{(l)}}, m_{1:K_{t},t}^{(l)})} {P(c_{1:N_{t}, t}^{(l)} | m_{t}^{(l)}, \theta_{t-1}^{(l)}, \bold{x}_{1:N_{t},t} ) } $
\ENDFOR
\FOR {l = 1 : L}
\STATE $w_{t}^{(l)} \leftarrow \frac{\tilde{w}_{t}^{(l)}}{\sum_{l=1}^{L} \tilde{w}_{t}^{(l)} }$   \hfill $\triangleright$ normalize weights
\ENDFOR
\STATE Resample particles $1, \ldots, L$ and weights $w_{t}^{(1)}, \ldots, w_{t}^{(L)}$   \hfill $\triangleright$ Section~\ref{sec:resample}
\ENDFOR
\end{algorithmic}
\end{algorithm*}

\subsubsection{Proposal Distribution for Assignments}
\label{sec:smc_proposal_1}

The probability of assignments given current cluster sizes, cluster parameters, and the Dirichlet process concentration parameter $\alpha$ can be written as
\begin{align}
\begin{split}
\label{smc_assig_eqn}
P &\left( c_{i,t} | m_{1:K_{t}, t}, \theta_{1:K_{t}, t}, \alpha \right) \propto
\text{C}(m_{1:K_{t}, t}, \alpha) \\ 
& \hspace{15mm}\times
\begin{cases}
F(\bold{x}_{i,t} | \theta_{c_{i,t},t}) \hspace{2mm} \text{if} \hspace{2mm} k \leq K_{t-1} \\
\int P(\bold{x}_{i,t} | \theta) \mathbb{G}_{0}(\theta)d\theta \hspace{2mm} k > K_{t-1}
\end{cases}
\end{split}
\end{align}
where C is defined in \eqref{C_distro}, and
$\int P(\bold{x}_{i,t} | \theta) \mathbb{G}_{0}(\theta)d\theta$ can be determined analytically, and is given in \eqref{marginal_obs_prob}.

\subsubsection{Proposal Distribution $q_{1}$}
\label{sec:smc_proposal_2}

The following is a distribution over cluster parameters $\theta_{k,t}$ given a set of $N$ observations $\bold{x}_{1:N, t}$. We define $q_{1}$ to be
\begin{equation}
\label{smc_q1_eqn}
q_{1}(\theta_{k,t} | \bold{x}_{1:N,t}) = P(\theta_{k,t} | \bold{x}_{1:N,t})
\end{equation}
where samples can be drawn from $P(\theta_{k,t} | \bold{x}_{1:N,t})$ using the Bayesian updates found in \eqref{bayesian_update_1}, \eqref{bayesian_update_2}, \eqref{bayesian_update_3}, and \eqref{bayesian_update_4}, where the $\bold{x}_{1:N,t}$ are taken to be the observations.

\subsubsection{Proposal Distribution $q_{2}$}
\label{sec:smc_proposal_3}

The following is a distribution over cluster parameters $\theta_{k,t}$ given a set of $N$ observations $\bold{x}_{1:N, t}$ and the cluster parameters at a previous time, $\theta_{k,t-1}$. We define $q_{2}$ to be
\begin{equation}
\label{smc_q2_eqn}
q_{2}(\theta_{k,t} | \theta_{k,t-1}, \bold{x}_{1:N,t}) = P(\theta_{k,t} | \theta_{k,t-1}, \bold{x}_{1:N,t})
\end{equation}
where samples can be drawn from $P(\theta_{k,t} | \theta_{k,t-1} \bold{x}_{1:N,t})$ using the Bayesian updates found in \eqref{bayesian_update_1}, \eqref{bayesian_update_2}, \eqref{bayesian_update_3}, and \eqref{bayesian_update_4}, where both the $\bold{x}_{1:N,t}$ and auxiliary variables $z_{k,t,1:M}$ are taken to be the observations.

\subsection{Resampling Particles and Particle Weights}
\label{sec:resample}

At each time step $t \in \{ 1, \ldots, T \}$, after all of the $L$ particles have been sampled and their associated weights computed, a resampling step is carried out. In this step, $L$ new particles are sampled from current set of $L$ particles.
Resampling strategies such as those described in \cite{gasthaus_thesis} and \cite{douc2005comparison} can be used.

\subsection{From Inference to Tracking Results}
\label{sec:inferencetotrackingresults}
 
Each inferred cluster is taken to be a distinct object, and the sequence of means and covariance matrices for a given cluster are used to determine the position and spatial region, respectively, of a given object over a sequence of time steps. In particular, the mean parameter is taken to be the centroid of an object, and a 2-dimensional oval centered on the mean that contains a specified percentage of the normal distribution mass (where we refer to the specified percentage as the confidence value) is taken to be the spatial region of the object. We report the maximum a posteriori (MAP) sample as our result.


\section{Experiments}
\label{sec:experiments}

This section provides details on performance evaluation metrics that have been developed to quantify results in object detection and tracking studies (and adopted in this paper), synthetic video experiments that verify aspects of the developed technique, and benchmark video experiments that demonstrate the performance of this technique in relation to other strategies (including state-of-the-art, object-specific strategies) that have been developed in recent years.

\subsection{Performance Evaluation Metrics}
\label{sec:performanceevaluationmetrics}

Performance evaluation metrics, which provide a standardized way of quantifying the success of a detection and tracking procedure on a given video, have started to become consistently used in the past four years. The metrics presented in \cite{kasturi_2008} and used in \cite{ellis_2010, taj_2007, lee_2009} have become well established for evaluating the performance of object detection and tracking in videos and have been adopted by the Video Analysis and Content Extraction (VACE) program and the Classification of Events, Activities, and Relationships (CLEAR) consortium, two large-scale efforts concerned with video tracking and interaction analysis. The two metrics used to quantify the experimental results in this study are known as the Sequence Frame Detection Accuracy (SFDA) and Average Tracking Accuracy (ATA). Details on how these metrics are defined and computed are given in Appendix~\ref{sec:performancemetricdetails}.

The above metrics are dependent upon ground-truth data specifying the positions of each object in each frame throughout a video sequence. In the experiments described below, we recorded the synthetic video ground-truth during construction of the videos (described in Section~\ref{sec:syntheticvideodatasets}), and we used the Video Performance Evaluation Resource (ViPER) ground-truth software \cite{doermann_2000}, an open source tool commonly used in the video tracking community, to author ground-truth data for each of the benchmark datasets.

The ground-truth authored by the ViPER tool took the form of bounding boxes denoting the spatial position of each object at each time step. Consequentially, to find the spatial overlap between results and ground-truth, which is intrinsic to both metrics, a rectangular bounding box was needed per object per time step from the results of the algorithm. We took the maximal and minimal axially aligned values of the oval inferred by our algorithm (as described in Section~\ref{sec:inferencetotrackingresults}) to be the sides of a representative bounding box for a given object at a given frame.

\subsection{Data Extraction in Experiments}
\label{sec:dataextractioninexperiments}

Frame differencing was used in all experiments to identify pixels exhibiting motion. 
For each pixel $\bold{x} = ( x_{1},$ $x_{2},$ $t )$ recorded during frame differencing, we also extracted color information. Specifically, we specified a square, $L$ pixels in length, centered on $(x_{1}, x_{2})$, that contained a set of pixels surrounding $\bold{x}$ in frame $t$. We chose to capture the hue for each pixel. The set of possible hue values (i.e. the range of hues to which a pixel may be assigned) was partitioned into $V$ bins, and the number of pixels with a color value lying in each of the bins yielded the $V$ dimensional vector of color counts. For all experiments, we chose $V=10$.

\subsection{Synthetic Video Datasets}
\label{sec:syntheticvideodatasets}

\begin{figure*}[!]
  \centering               
  \subfloat[]{\includegraphics[width=0.32\textwidth]{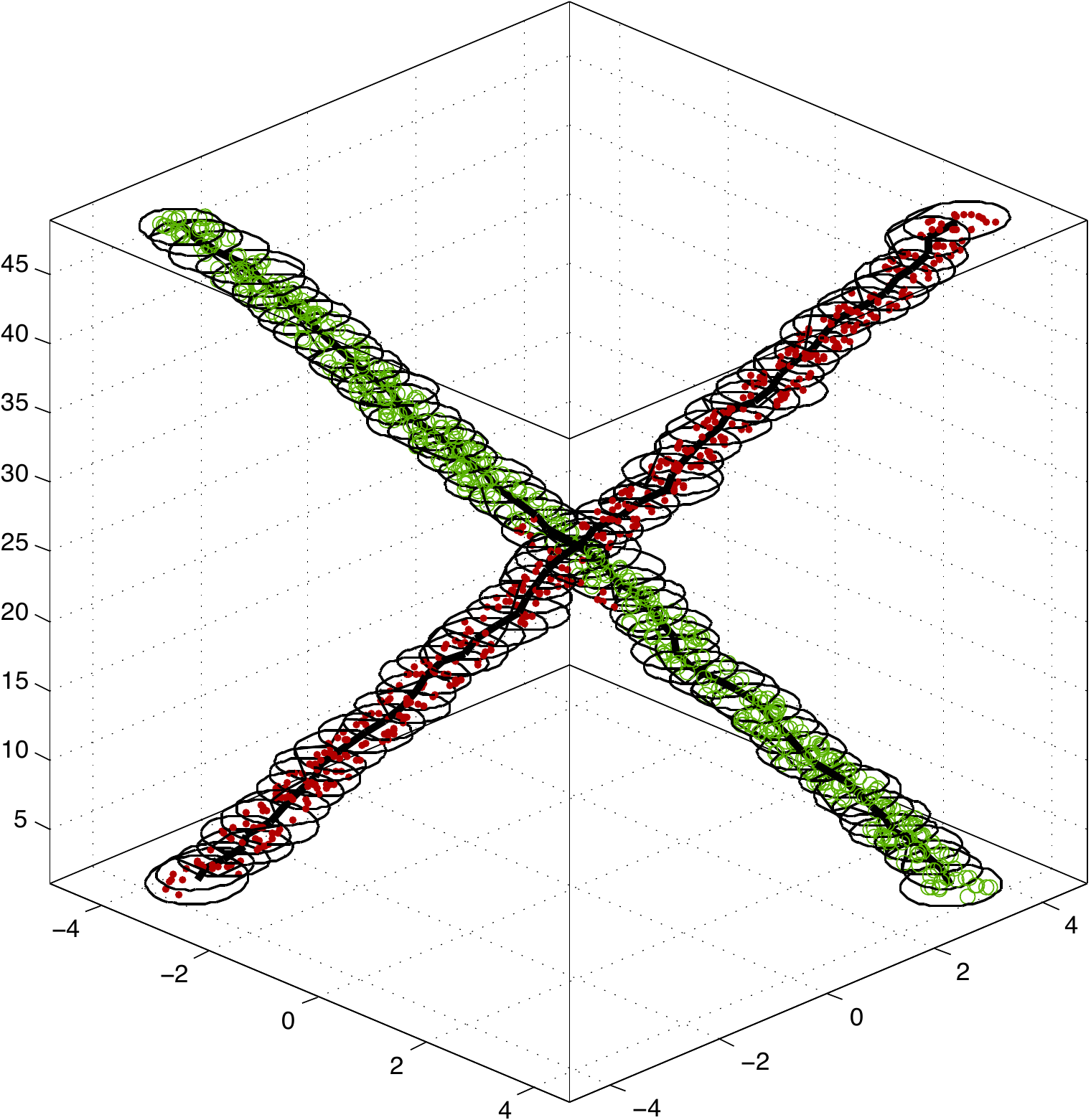}} \hspace{0.5mm}
  \subfloat[]{\includegraphics[width=0.32\textwidth]{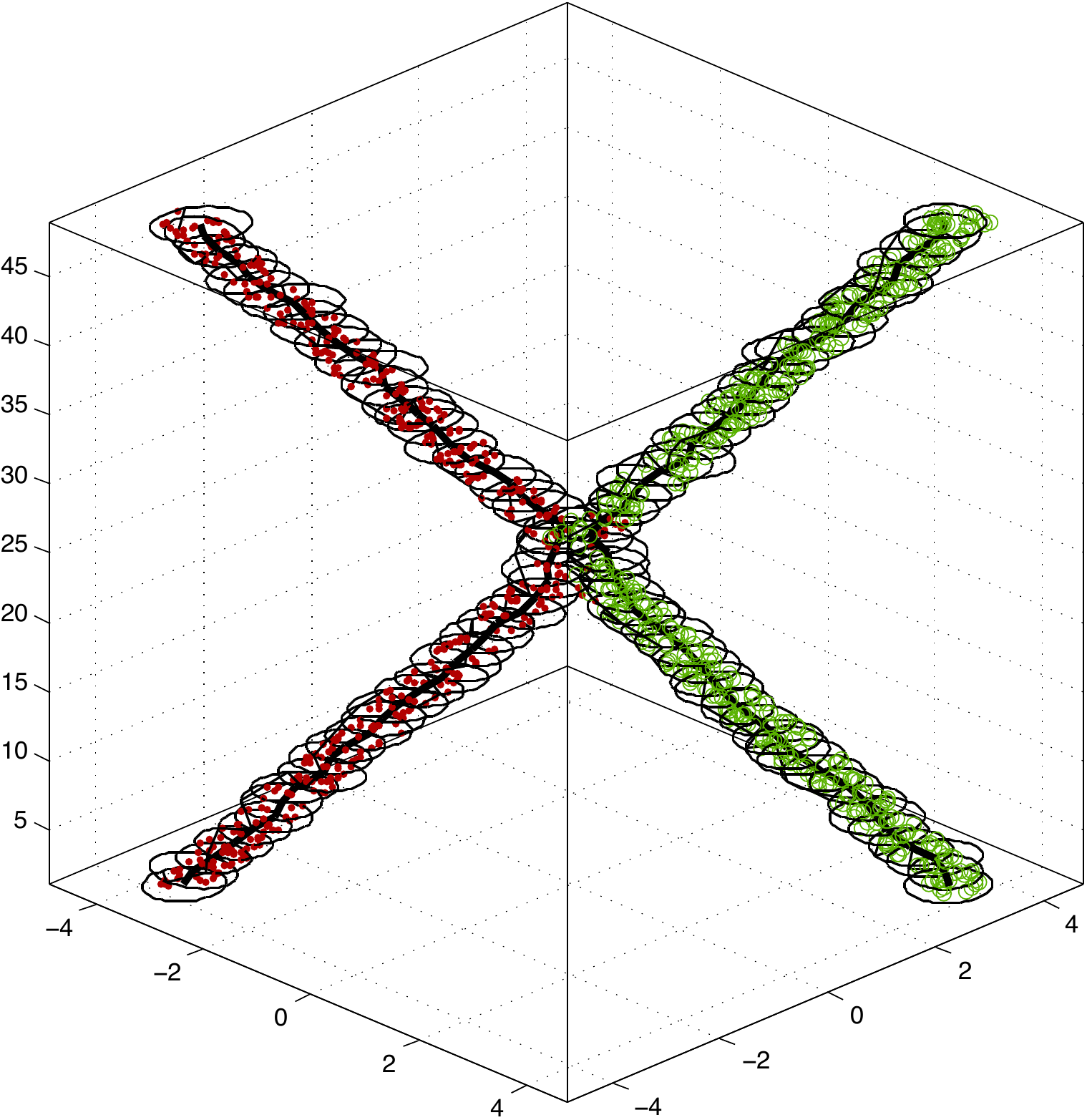}} \hspace{0.5mm}
  \subfloat[]{\includegraphics[width=0.32\textwidth]{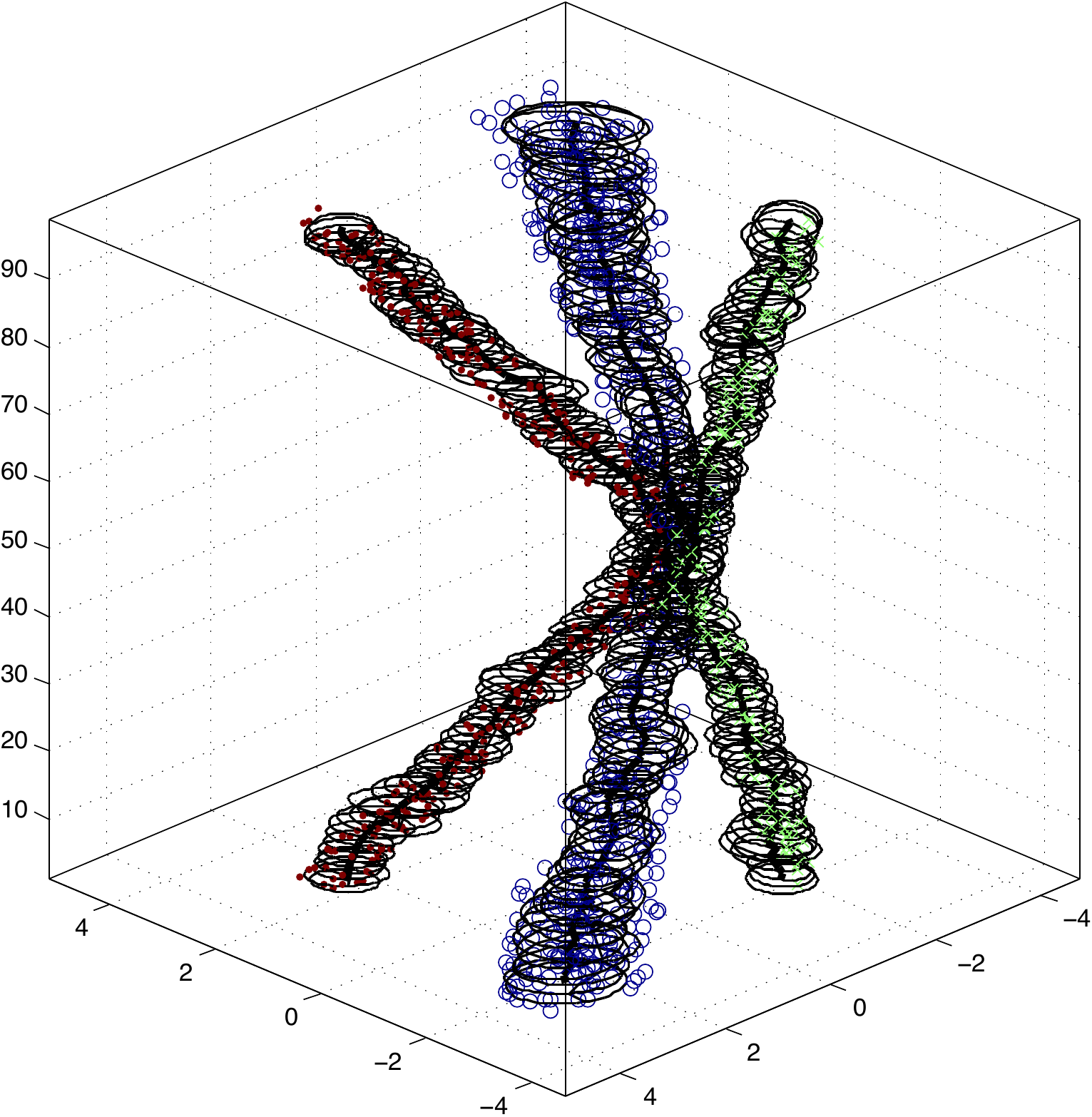}}
  \caption{Each plot shows a sample from the posterior distribution of the model for synthetic experiments one (a and b) and two (c), where the vertical axis represents frame number, the horizontal axes represent spatial position, objects are denoted by marker colors and marker types, and the mean and standard deviation are shown. In all cases, the objects are successfully tracked through occlusion, whether they travel in a straight line (a), reverse direction (b), or do a combination of both (c).}
  \label{fig:synth_one_plot}
\end{figure*}

Each of the following synthetic videos consists of a sequence of 200 images (each of size $500$ $\times$ $500$ pixels) containing a number of smaller colored squares of different (and potentially time-varying) sizes moving at varied speeds and trajectories over a black background. The synthetic videos contain instances of occlusion (where one or more objects are briefly hidden) and objects with time-varying appearances and behaviors, as these notoriously decrease the accuracy of detection and tracking. After each video was constructed, the extraction procedure described in Section~\ref{sec:dataextractioninexperiments} (using $L=3$) and inference procedures described in Section~\ref{sec:inference} were carried out to return a sequence of multivariate-normal-parameters (means and covariance matrices), which are used to determine a sequence of positions and ovals approximating, respectively, the locations and shapes of a tracked object over each frame that it is present in the video (as outlined in Section~\ref{sec:inferencetotrackingresults}).

The first synthetic video experiment aimed to test the ability of the model and inference procedure to maintain the identity of independent objects based on color information alone. Two videos were constructed, both containing a red square (rgb value $[255,0,0]$ and size $20$ $\times$ $20$ pixels) and a blue square (rgb value $[0,0,255]$ and size $20$ $\times$ $20$ pixels). In both videos, the squares begin at opposite sides of the scene at frame $f=1$ and travel towards each other, arriving at the same location at $f=100$ (where the blue square occludes the red square). The second half of the two videos differ in that both squares in the first video continue in the same direction and end at the other's starting position at $f=200$, and both squares in the second video reverse directions and end at their initial starting positions at $f=200$. The frame difference extraction yields identical spatial features in both videos; hence, successful tracking depends fully on the incorporation of color information into the model.

Parameters were set to the same values for inference on both videos: $\alpha = 0.1, \rho = 0.3, M = 10, \boldsymbol{\mu}_{0} = (0,0), \kappa_{0} = 0.05, \nu_{0} = 5, \Lambda_{0} = \left( \begin{smallmatrix} 1&0\\ 0&1 \end{smallmatrix} \right)$, and $\bold{q}_{0} = (5, \ldots, 5)$. Inference was carried out using the MCMC algorithm (Section~\ref{sec:MCMC}); the MAP sample correctly tracked both colored squares through occlusion in both videos, and is shown in Figure~\ref{fig:synth_one_plot}. 

The second synthetic video experiment aimed to test tracking performance under occlusion, object appearance change, and motion change. A video was constructed showing a red square (rgb value $[255,0,0]$), a green square (rgb value $[0,255,0]$), and a blue square (rgb value $[0,0,255]$). The red square was of size $20$ $\times$ $20$ pixels, the blue square was of size $15$ $\times$ $15$ pixels, and the green square began at size $50$ $\times$ $50$ pixels at frame $f=1$, linearly shrinks to $10$ $\times$ $10$ pixels at $f=100$, then linearly grows back to $50$ $\times$ $50$ pixels by the end of the video, $f=200$. Furthermore, the red and blue squares display the same behavior as in the second video of the first synthetic experiment (they begin at opposite sides of the scene traveling towards each other, cross at the center of the scene at $f=100$, and reverse direction, ending at their initial positions at $f=200$). The green square begins at a point equidistant from the other two squares, intersects with them as they overlap (causing the blue square to occlude the other two), and continues on in a direction at a 20 degree angle from its initial trajectory.

Parameters were set to the same values chosen in the first synthetic experiment. The MCMC inference algorithm correctly tracked all three objects through occlusion and inferred the appearance and size shifts. Figure~\ref{fig:synth_one_plot} shows a sample from the posterior distribution of the cluster parameters, where the mean and oval representation of the covariance matrix (with $0.5$ confidence value) are overlayed on the data. The data is plotted with time on the vertical axis, and the assignment of each data point to one of the three inferred clusters is denoted by color and marker type.

\begin{figure*}[!]
  \centering
  \subfloat[]{\label{fig:pets2000_results} \includegraphics[width=0.48\textwidth]{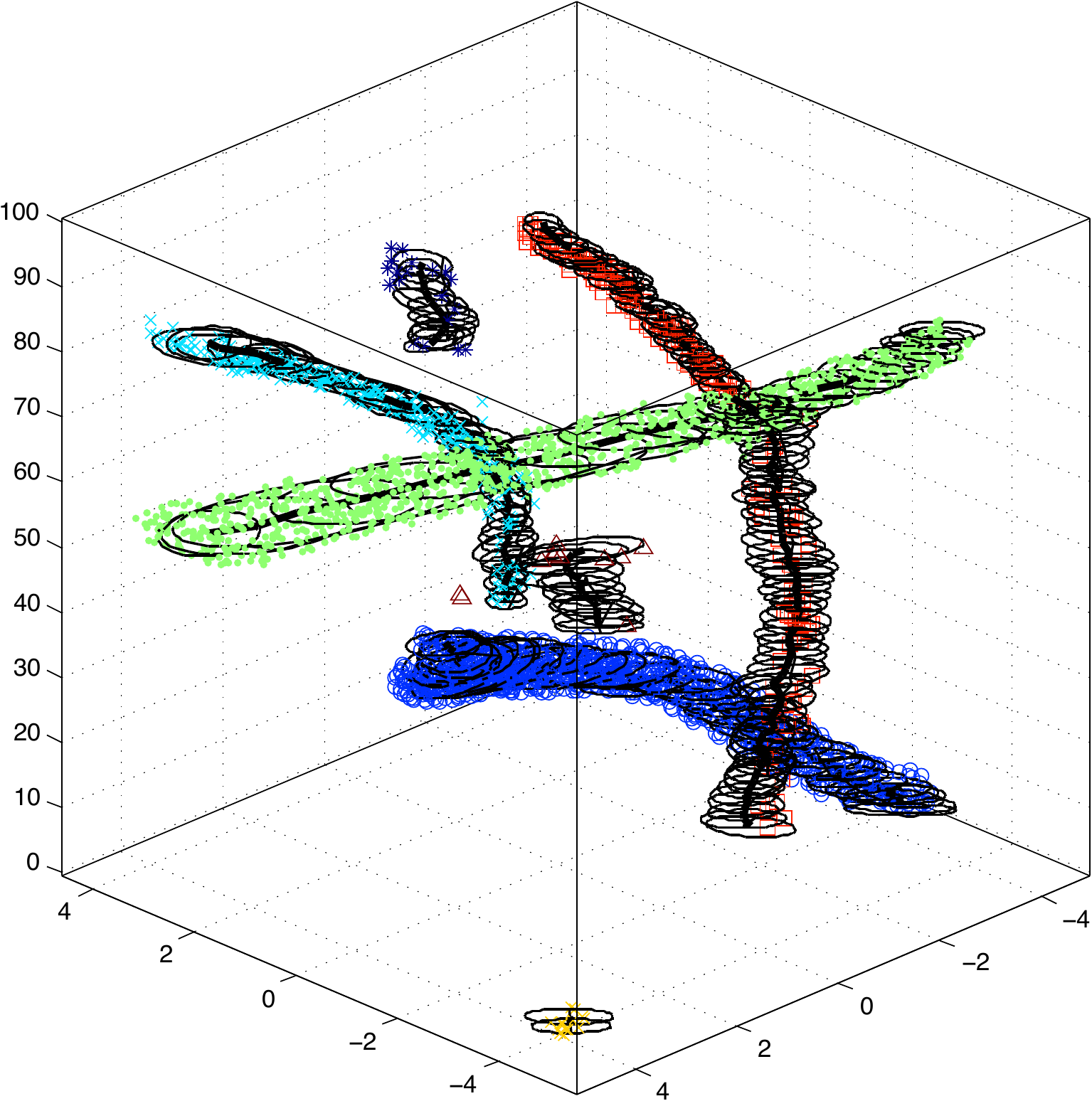}} \hspace{1mm}           
  \subfloat[]{\label{fig:pets2001_results} \includegraphics[width=0.48\textwidth]{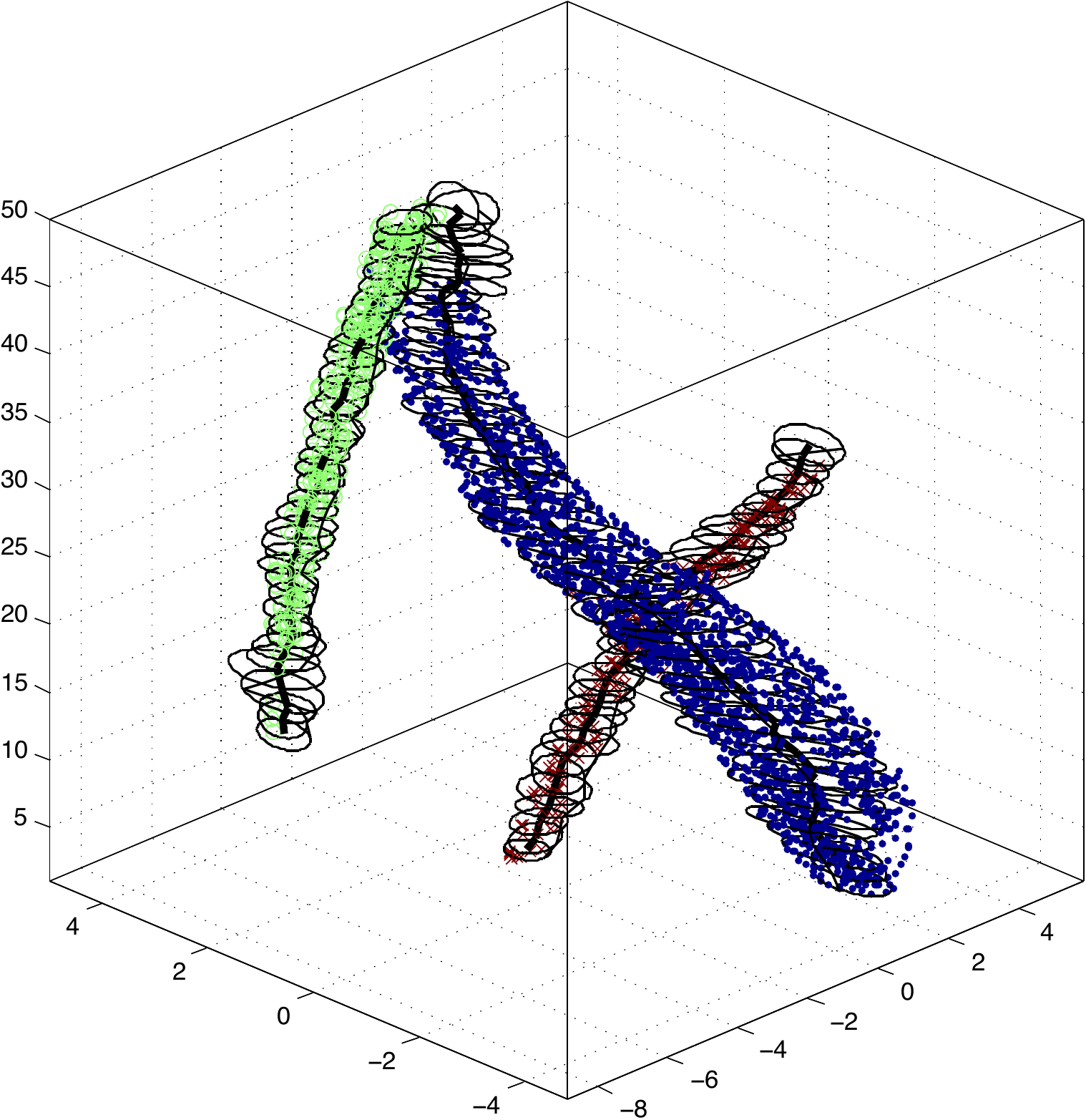}} \\
  \subfloat[]{\includegraphics[width=0.24\textwidth]{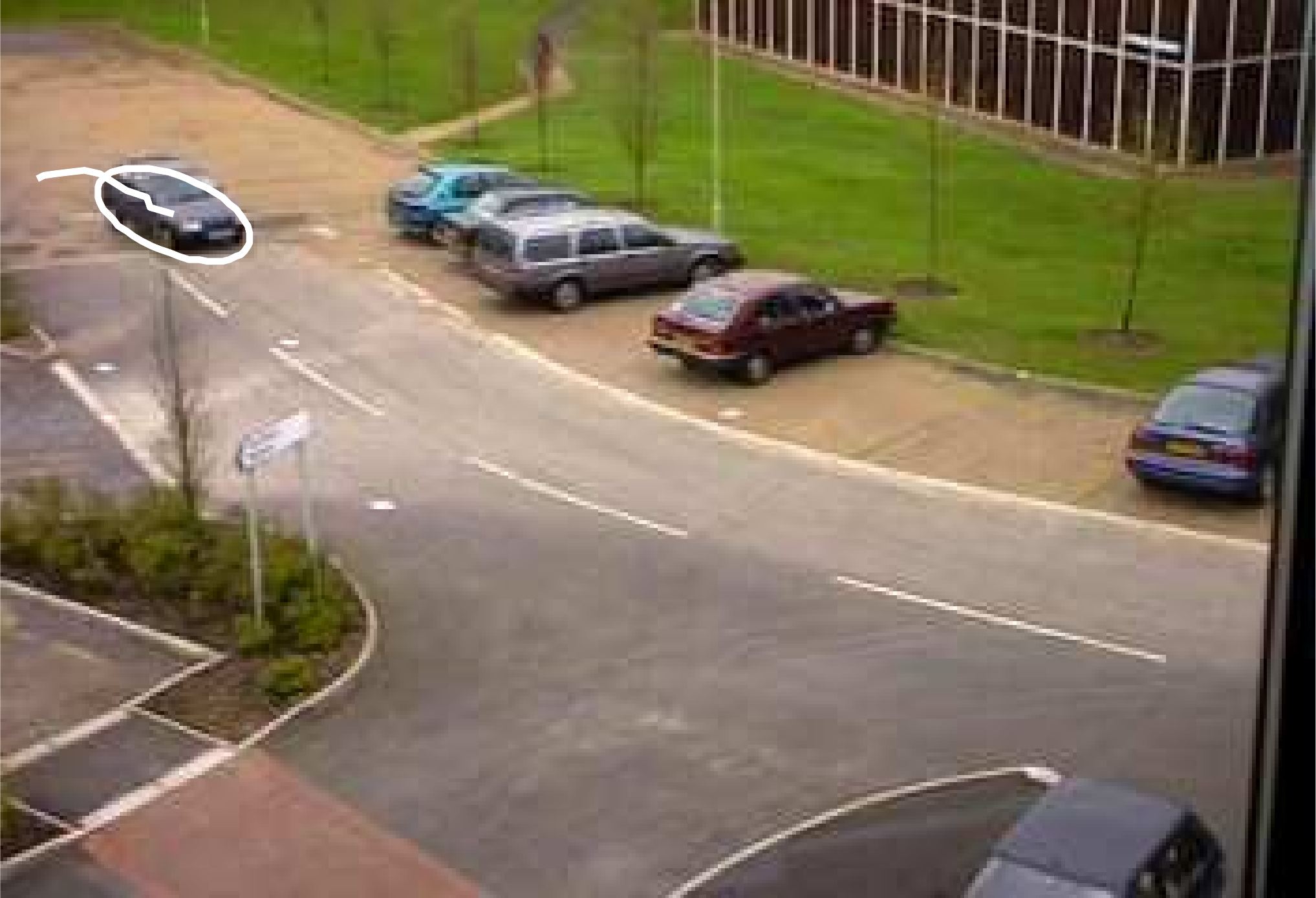}} \hspace{1pt}
  \subfloat[]{\includegraphics[width=0.24\textwidth]{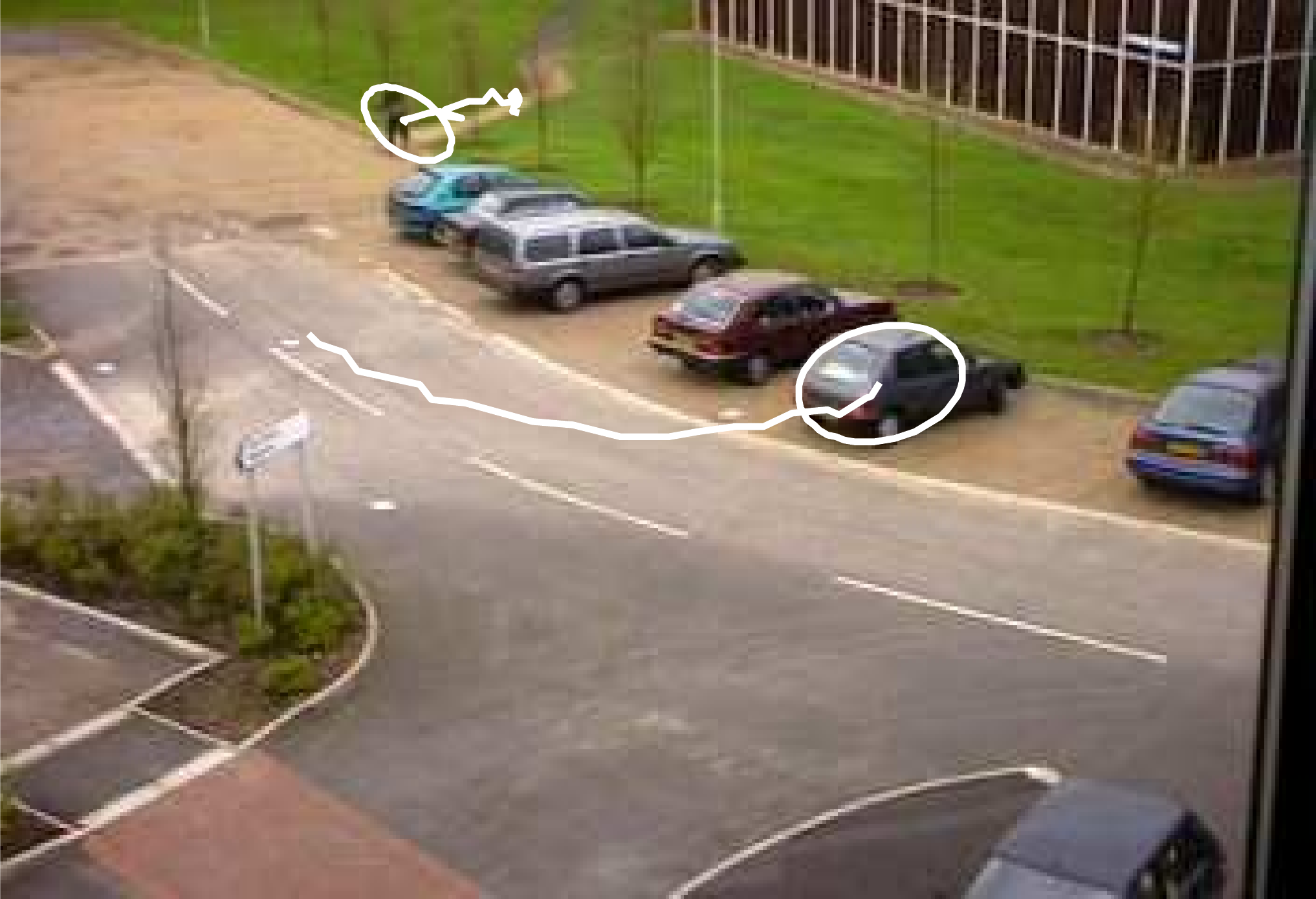}} \hspace{1pt}
  \subfloat[]{\includegraphics[width=0.24\textwidth]{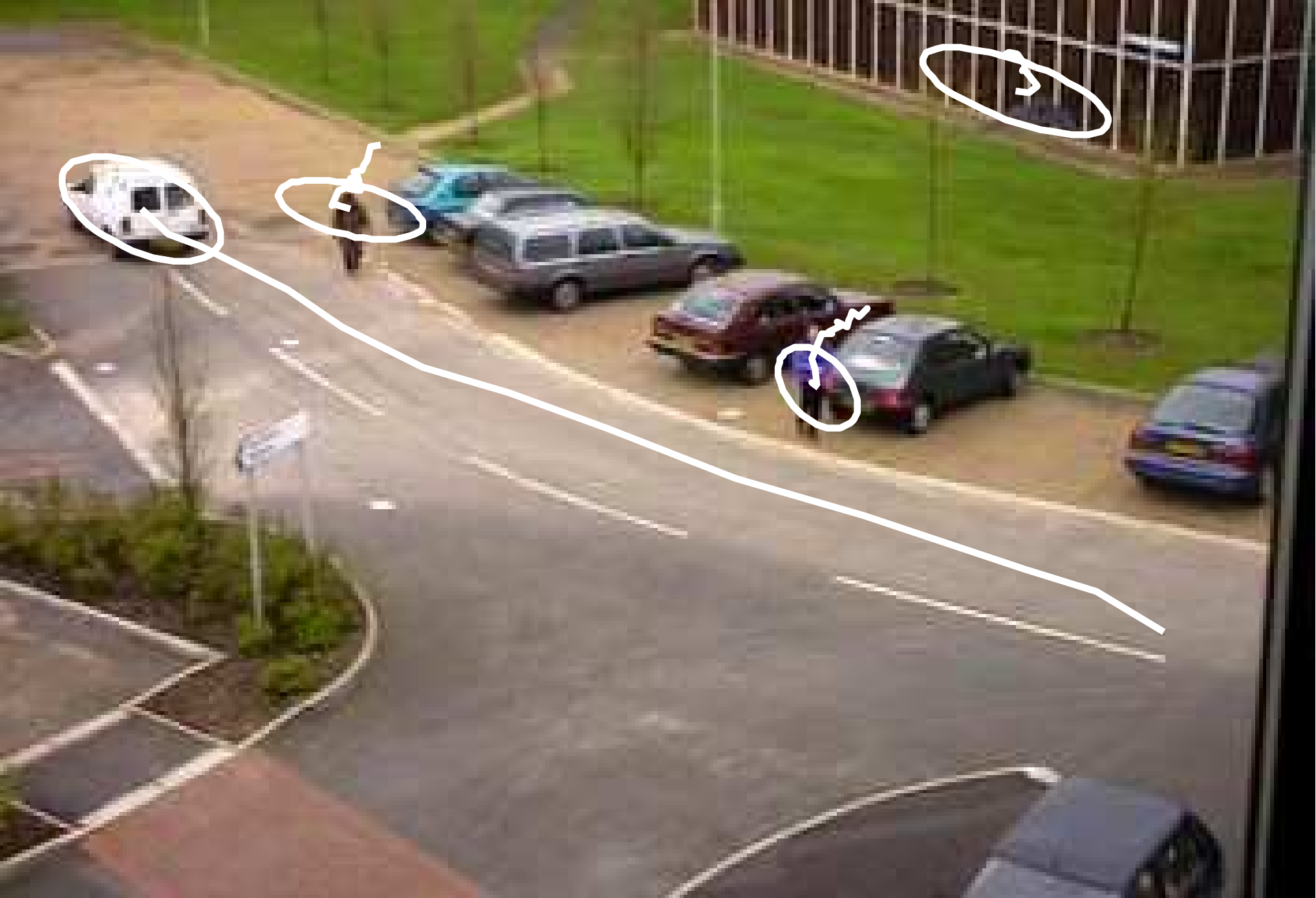}} \hspace{1pt}
  \subfloat[]{\includegraphics[width=0.24\textwidth]{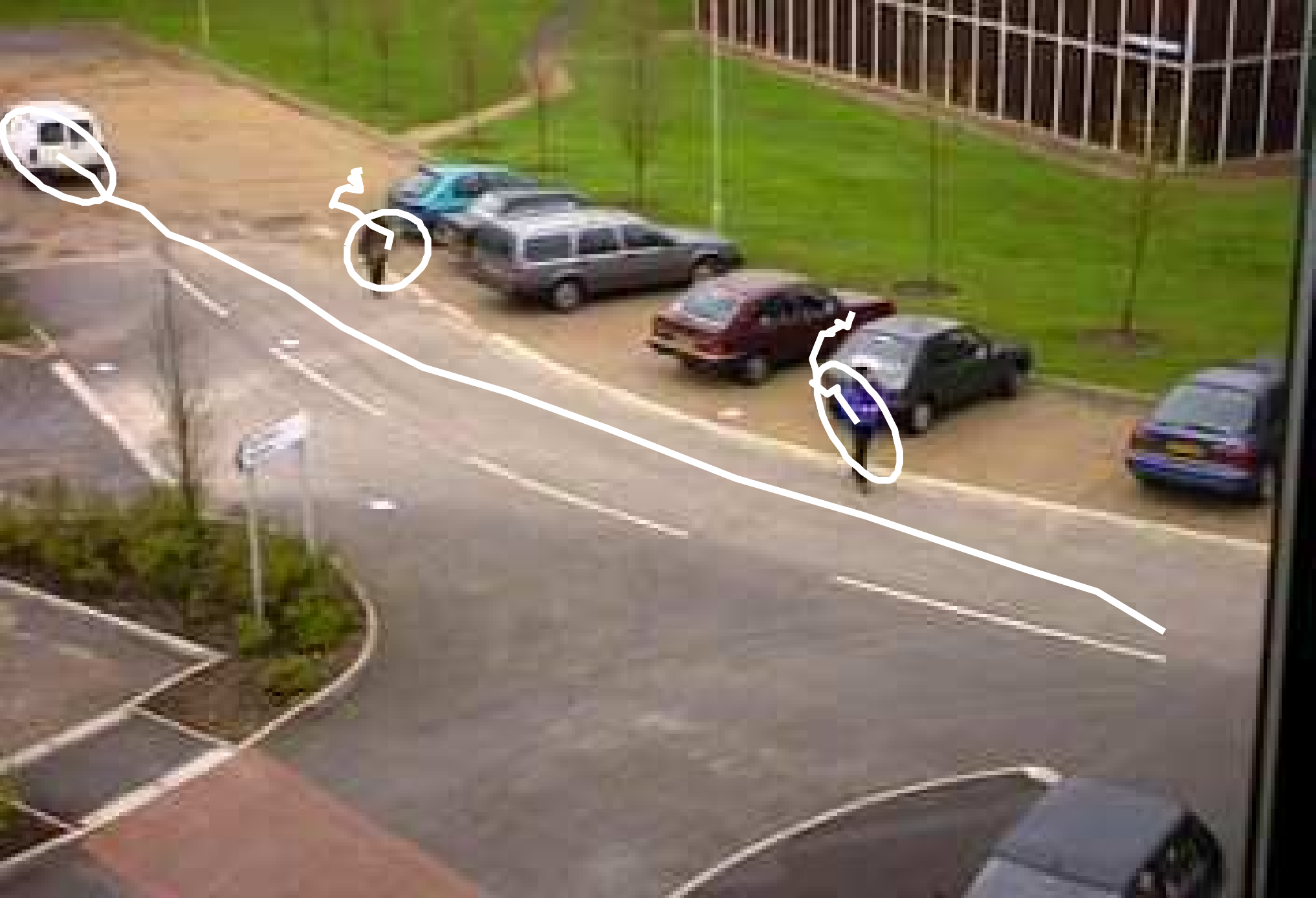}} \\           
  \subfloat[]{\includegraphics[width=0.24\textwidth]{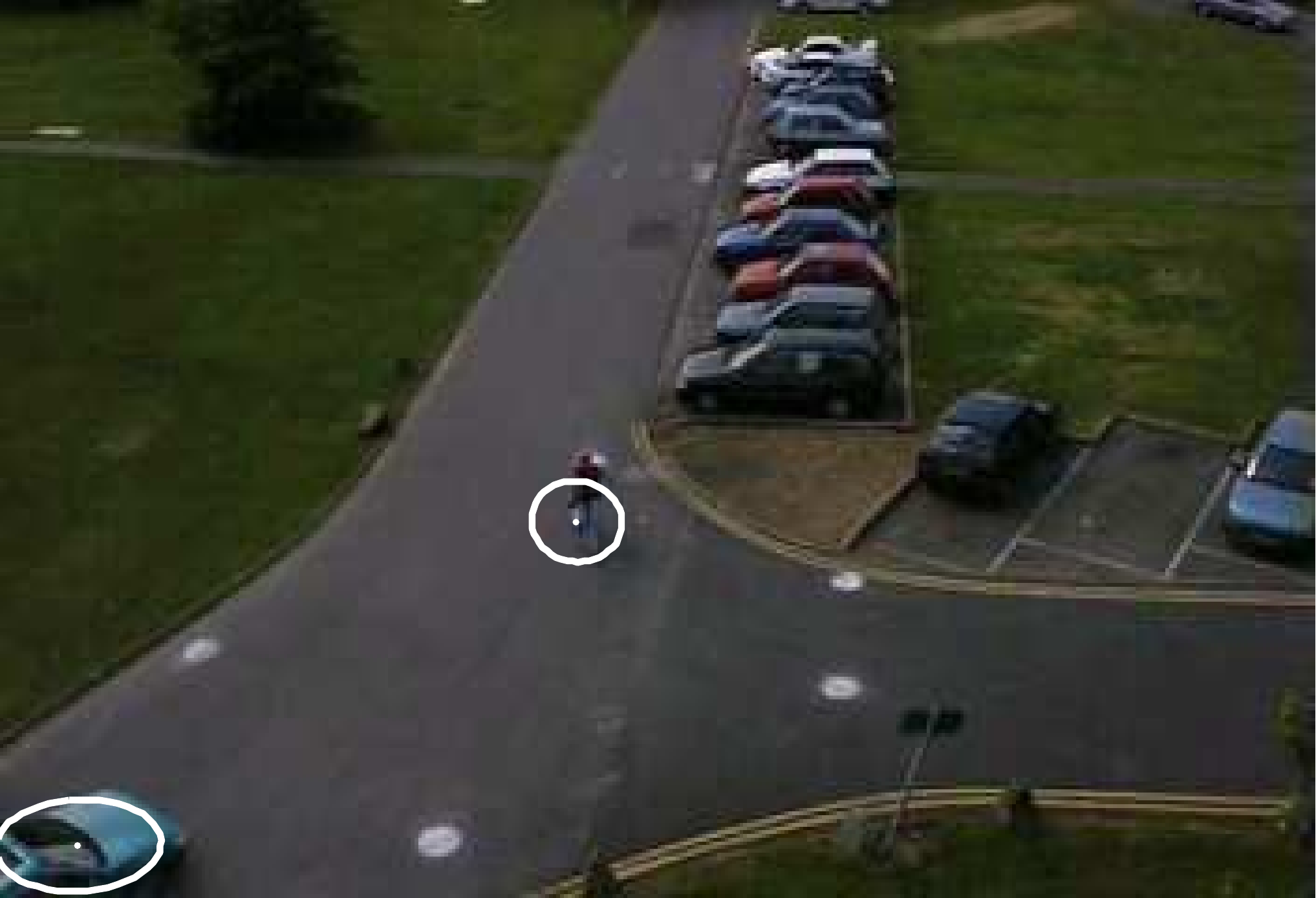}} \hspace{1pt}
  \subfloat[]{\includegraphics[width=0.24\textwidth]{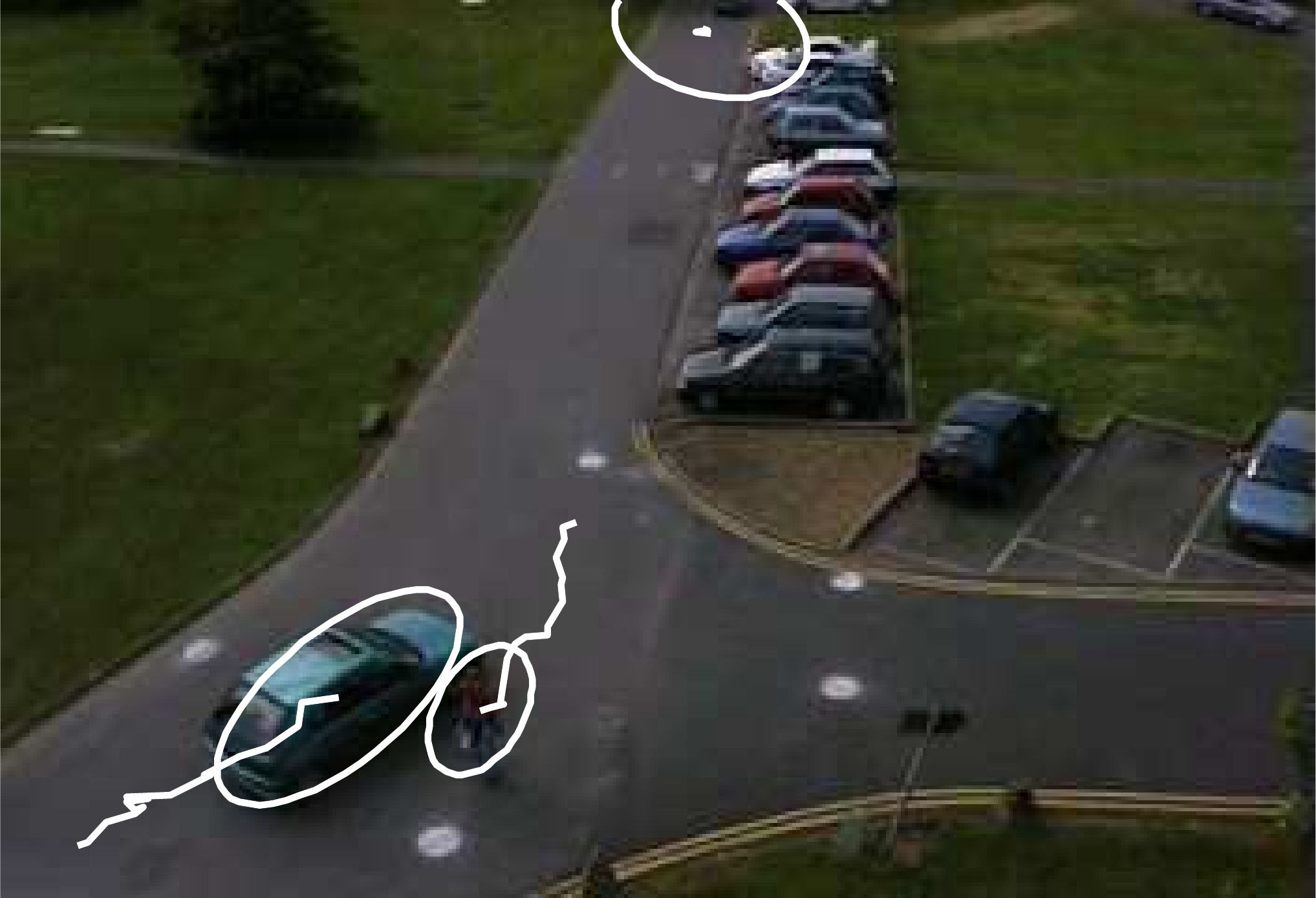}} \hspace{1pt}
  \subfloat[]{\includegraphics[width=0.24\textwidth]{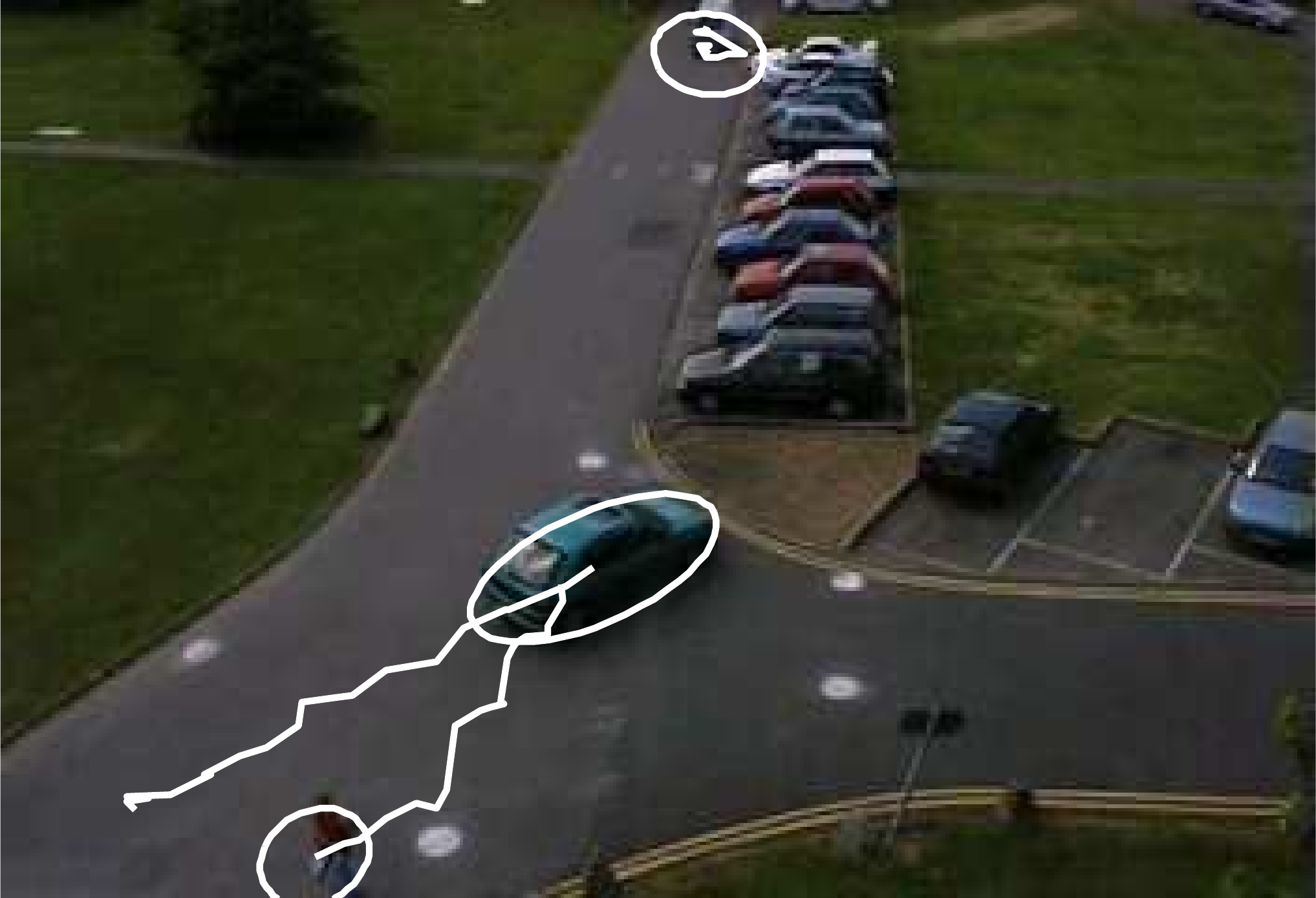}} \hspace{1pt}
  \subfloat[]{\includegraphics[width=0.24\textwidth]{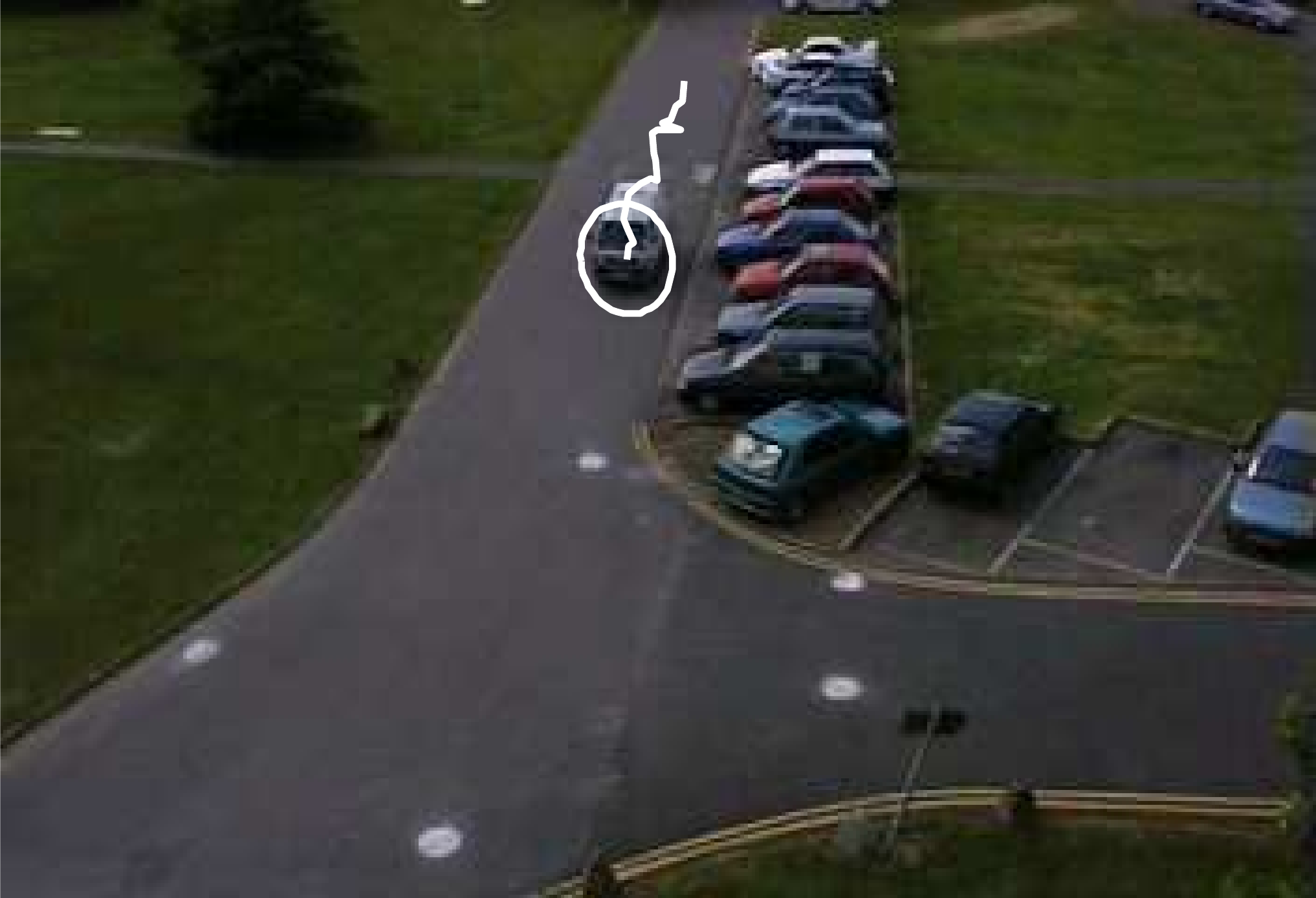}} \\
  \caption{Results from the PETS2000 (a) and PETS2001 (b) dataset. Both plots show a sample from the posterior distribution of the state, where the vertical axis denotes time, the horizontal axes represent spatial position, color represents assignment, and the mean and standard deviation are shown. Below are four frames from the PETS2000 (c-f) and PETS2001 (g-j) sequence with one posterior sample mean and covariance matrix representation shown for each frame (and one sample mean shown for the previous 20 frames).}
  \label{fig:pets2001_overlay}
\end{figure*}

\subsection{Benchmark Video Datasets}
\label{sec:benchmarkvideodatasets}

Benchmark video datasets for object tracking and detection have been produced to provide standard scenes on which researchers can compare detection and tracking results. These videos have been primarily produced for surveillance-related workshops---notably, for the International Workshop on Performance Evaluation of Tracking and Surveillance (PETS)---which provides researchers with video datasets and algorithmic goals on which to focus. Three commonly used benchmark videos from PETS workshops---one used in PETS2000, one in PETS2001, and one used both in PETS2009 and PETS2010---were chosen to demonstrate the method presented in this study. The performance metrics and benchmark datasets allow the methods developed in this paper to be quantitatively compared against other detection and tracking algorithms.

\begin{figure*}[!]
  \centering             
  \subfloat[]{\includegraphics[width=0.49\textwidth]{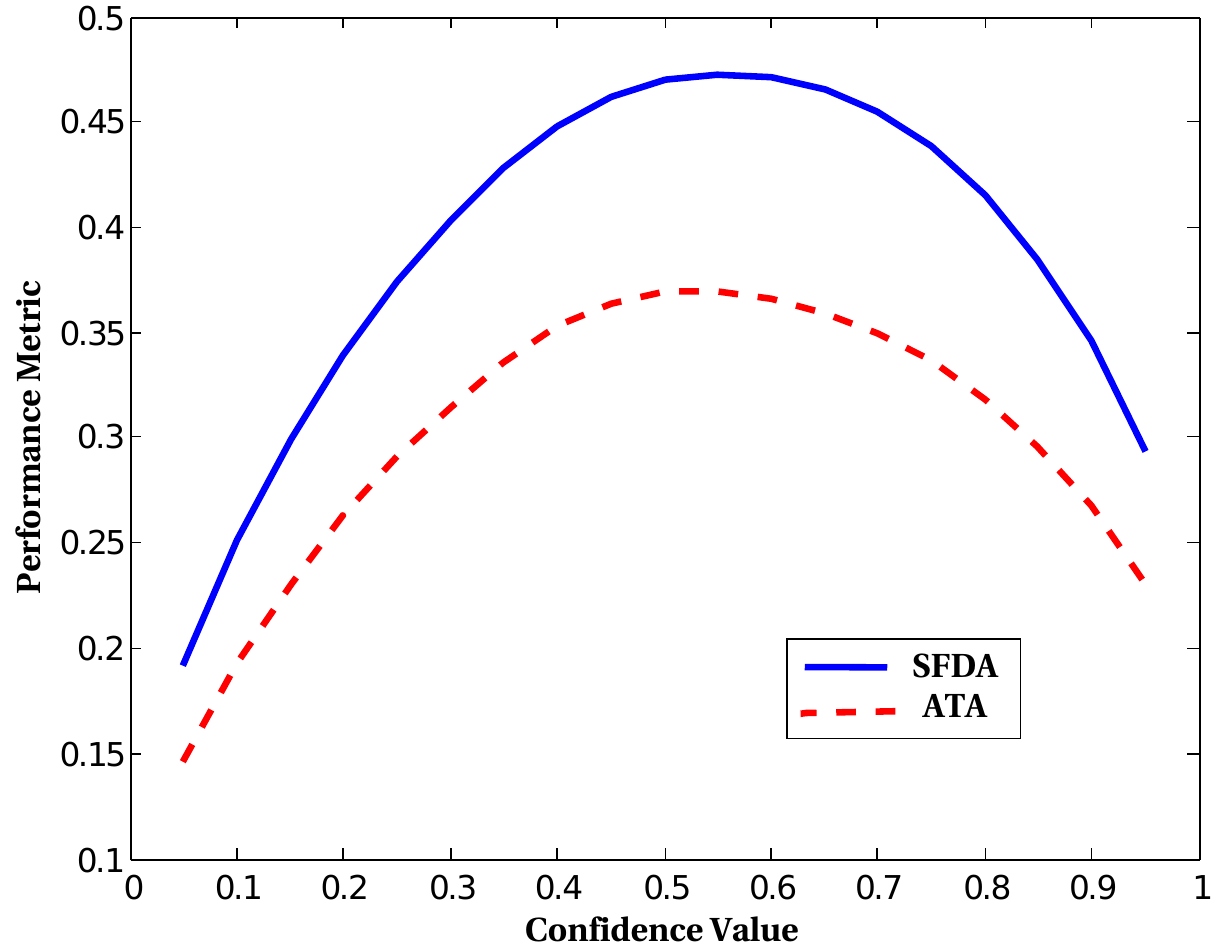}} \hspace{1mm}
  \subfloat[]{\includegraphics[width=0.49\textwidth]{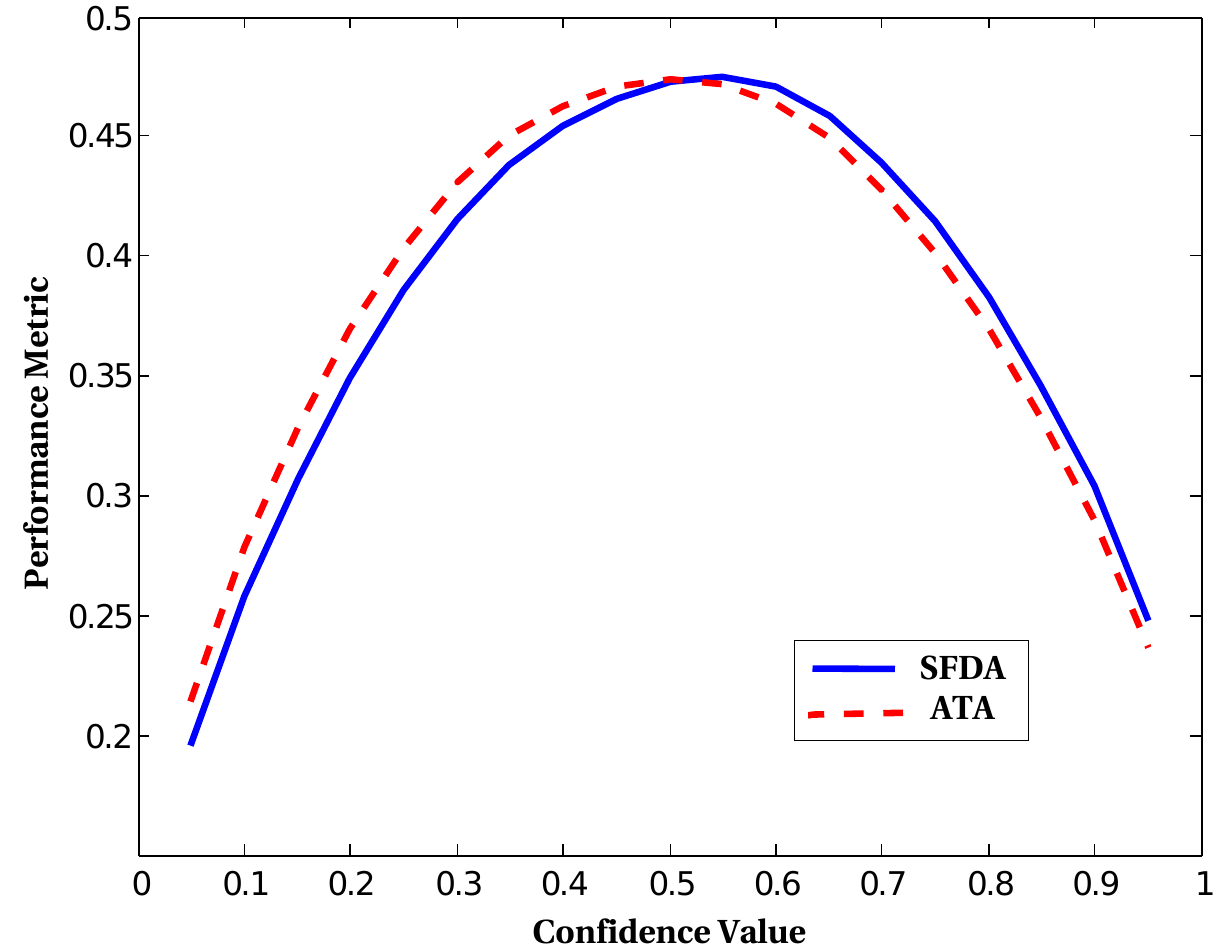}}
  \caption{The SFDA (blue solid line) and ATA (red dashed line) vs  confidence values from which an object's oval region is computed for (a) PETS2000 and (b) PETS2001 video datasets.}
  \label{fig:pm_conf}
\end{figure*}

\subsubsection{PETS2000 and PETS2001}
\label{sec:pets2000and2001}

The PETS2000 and PETS2001 video datasets both consist of a small number of humans and vehicles traveling across a parking lot, with video taken from above, emulating what might be recorded by standard outdoor surveillance equipment. The ``Test Sequence'', a set of images from a monocular, stationary camera, was used from the PETS2000 workshop, and ``View Two of Dataset 1", also taken via a monocular, stationary camera, was used from the PETS2001 workshop. The MCMC algorithm (described in Section~\ref{sec:MCMC}) was used for inference in these experiments.

Due to the computation required for the MCMC batch inference method (discussed further in Section~\ref{sec:conclusion}), only the final 1000 frames of the video were used from both datasets. Extraction was performed with frame differencing as described in Section~\ref{sec:dataextractioninexperiments}, using $L=3$. Parameter values were set to the same values as in the synthetic experiments ($\alpha = 0.1, \rho = 0.3, M = 10, \boldsymbol{\mu}_{0} = (0,0), \kappa_{0} = 0.05, \nu_{0} = 5, \Lambda_{0} = \left( \begin{smallmatrix} 1&0\\ 0&1 \end{smallmatrix} \right), \text{ and } \bold{q}_{0}$ $=$ $(5,$ $\ldots,5)$). The MCMC sampler was successful for both benchmark videos; each object was detected, tracked, and its shape estimated in manner very consistent with the ground-truth. The results for the PETS2000 dataset are displayed in Figure~\ref{fig:pets2000_results} and for the PETS2001 dataset in Figure~\ref{fig:pets2001_results}; in these figures, a sample from the posterior distribution of the cluster parameters is overlayed on the extracted data over a sequence of frames, where the assignment of each data point is represented by color and marker type.

To calculate performance metrics (both SFDA and ATA), one must specify a confidence value that allows the oval representing the region occupied by an object to be computed from the inferred covariance matrix of each cluster (as discussed in Section~\ref{sec:inferencetotrackingresults}). The performance metrics were found for a range of confidence intervals, and the resulting curves for both the PETS2000 and PETS2001 video are shown in Figure~\ref{fig:pm_conf}.

\subsubsection{PETS2009/2010}
\label{sec:pets20092010}

\begin{figure*}[!]
  \centering
  \subfloat[]{\includegraphics[width=0.43\textwidth]{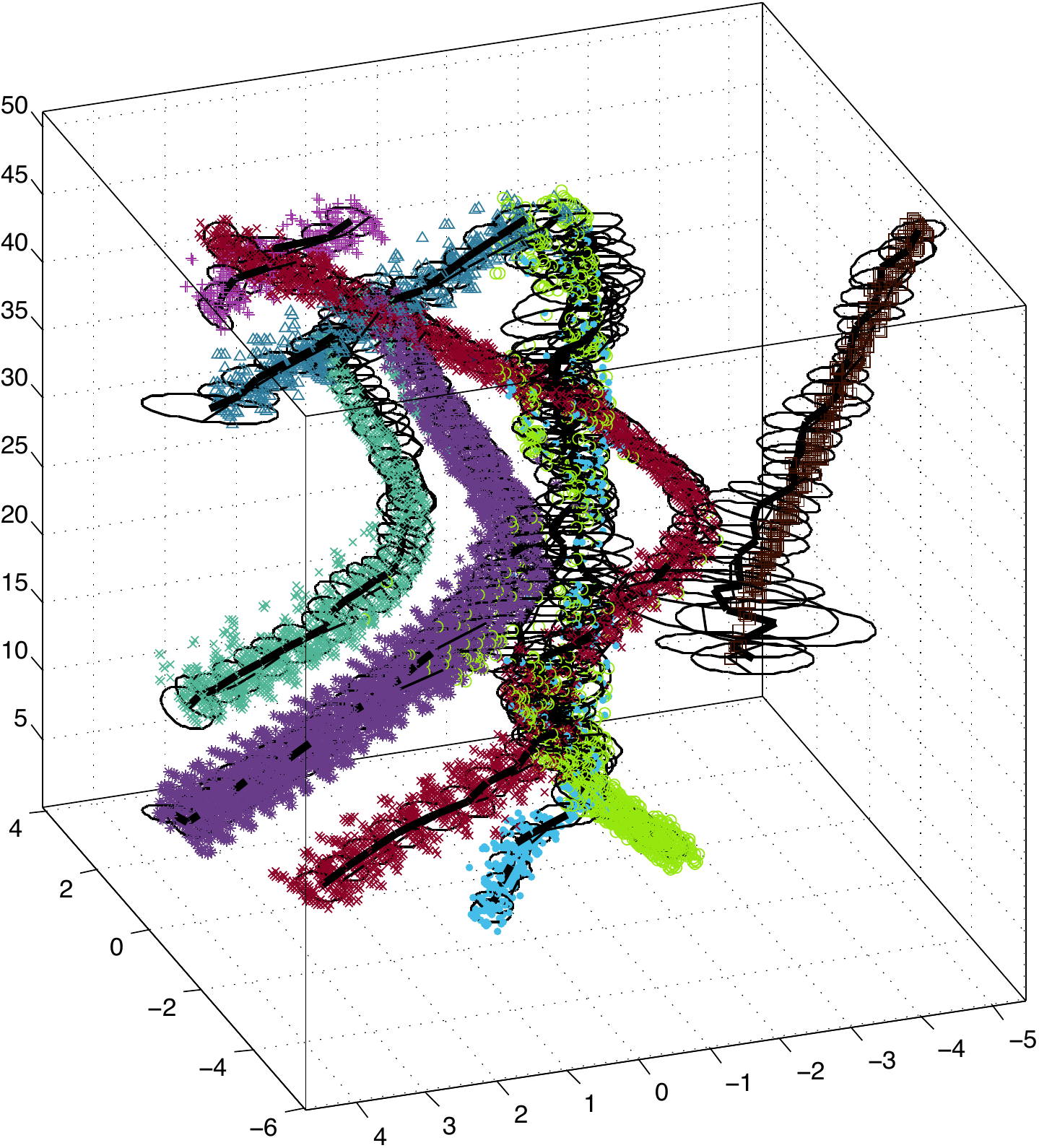}}
  \hspace{4mm}
  \subfloat[]{\includegraphics[width=0.53\textwidth]{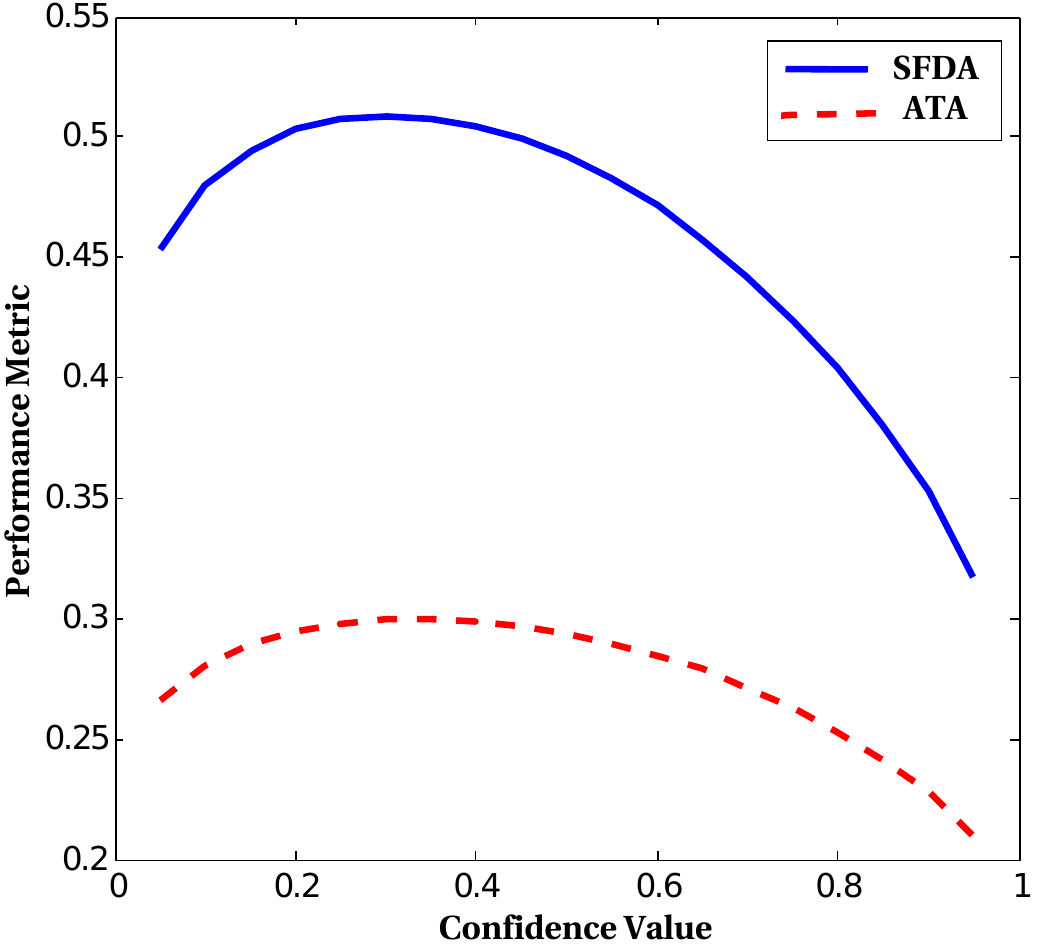}} \\
  \subfloat[]{\includegraphics[width=0.24\textwidth]{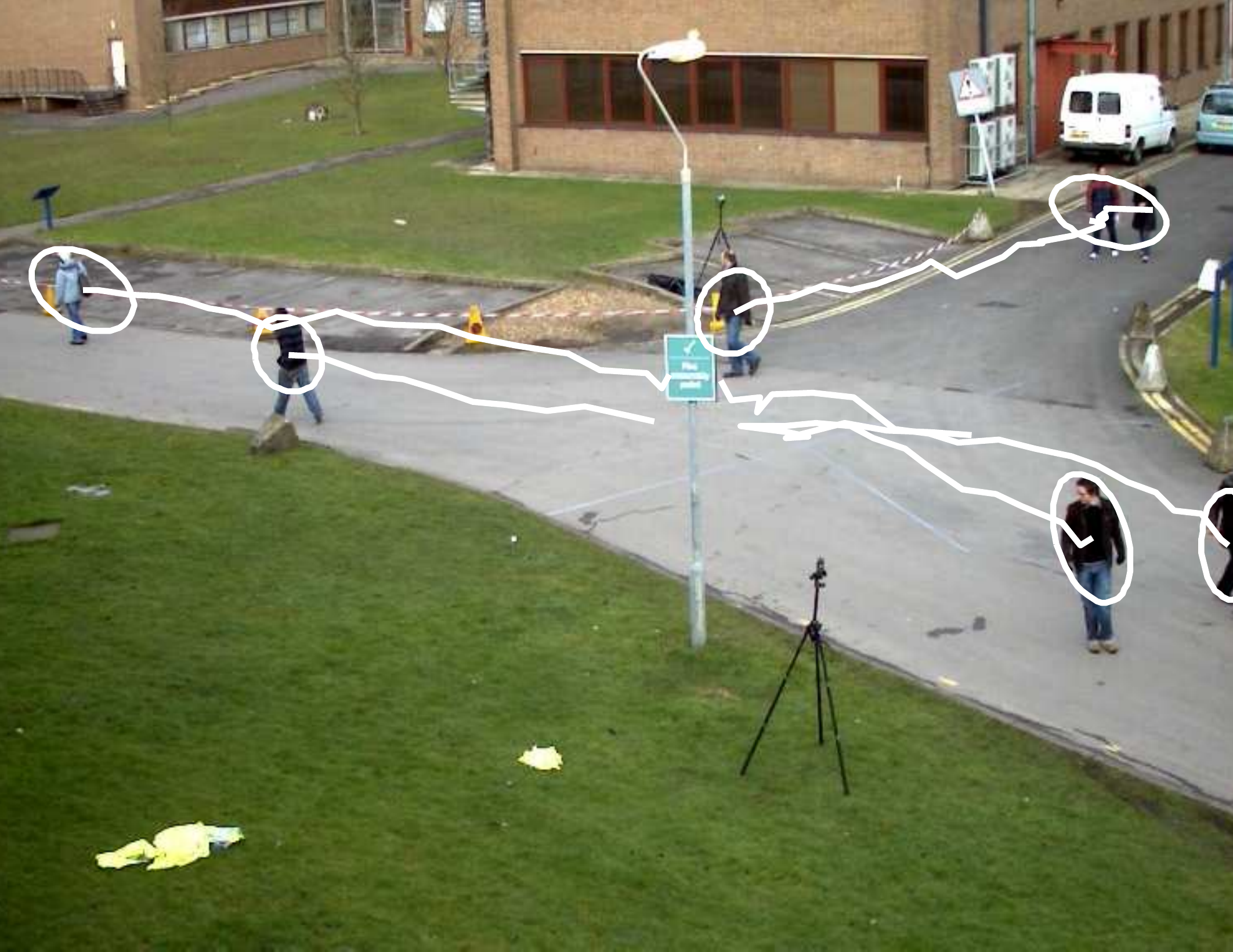}} \hspace{1pt}
  \subfloat[]{\includegraphics[width=0.24\textwidth]{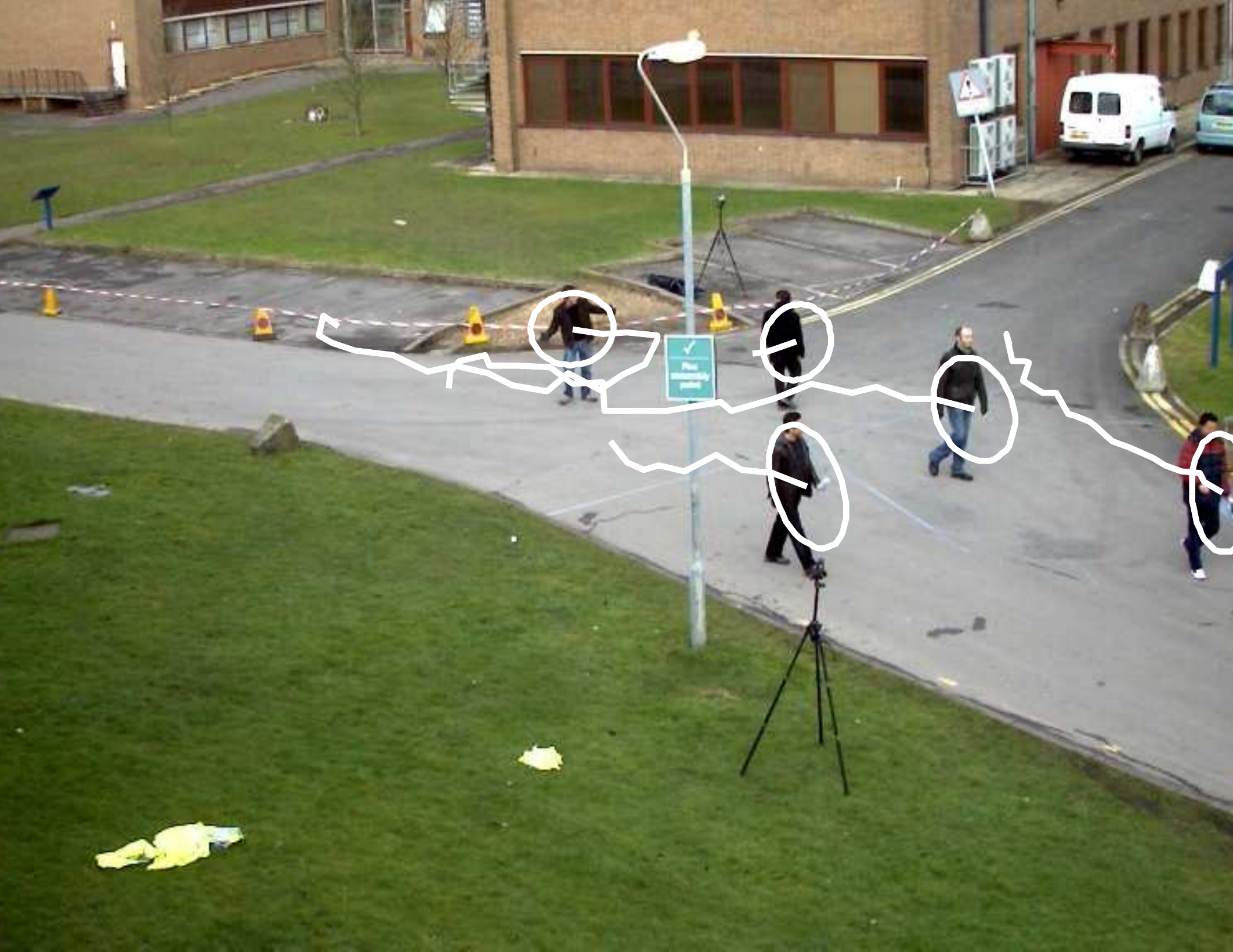}} \hspace{1pt}
  \subfloat[]{\includegraphics[width=0.24\textwidth]{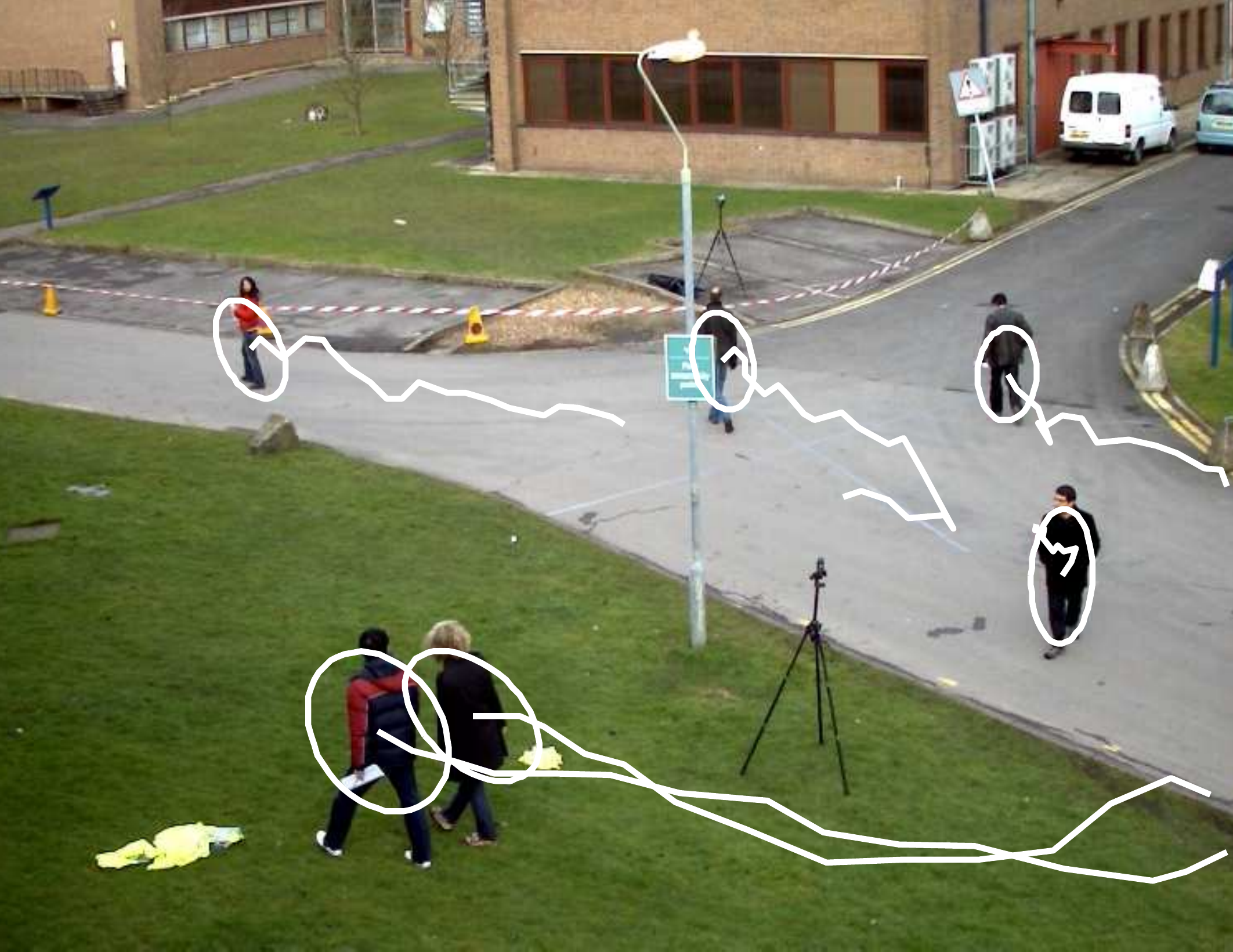}} \hspace{1pt}
  \subfloat[]{\includegraphics[width=0.24\textwidth]{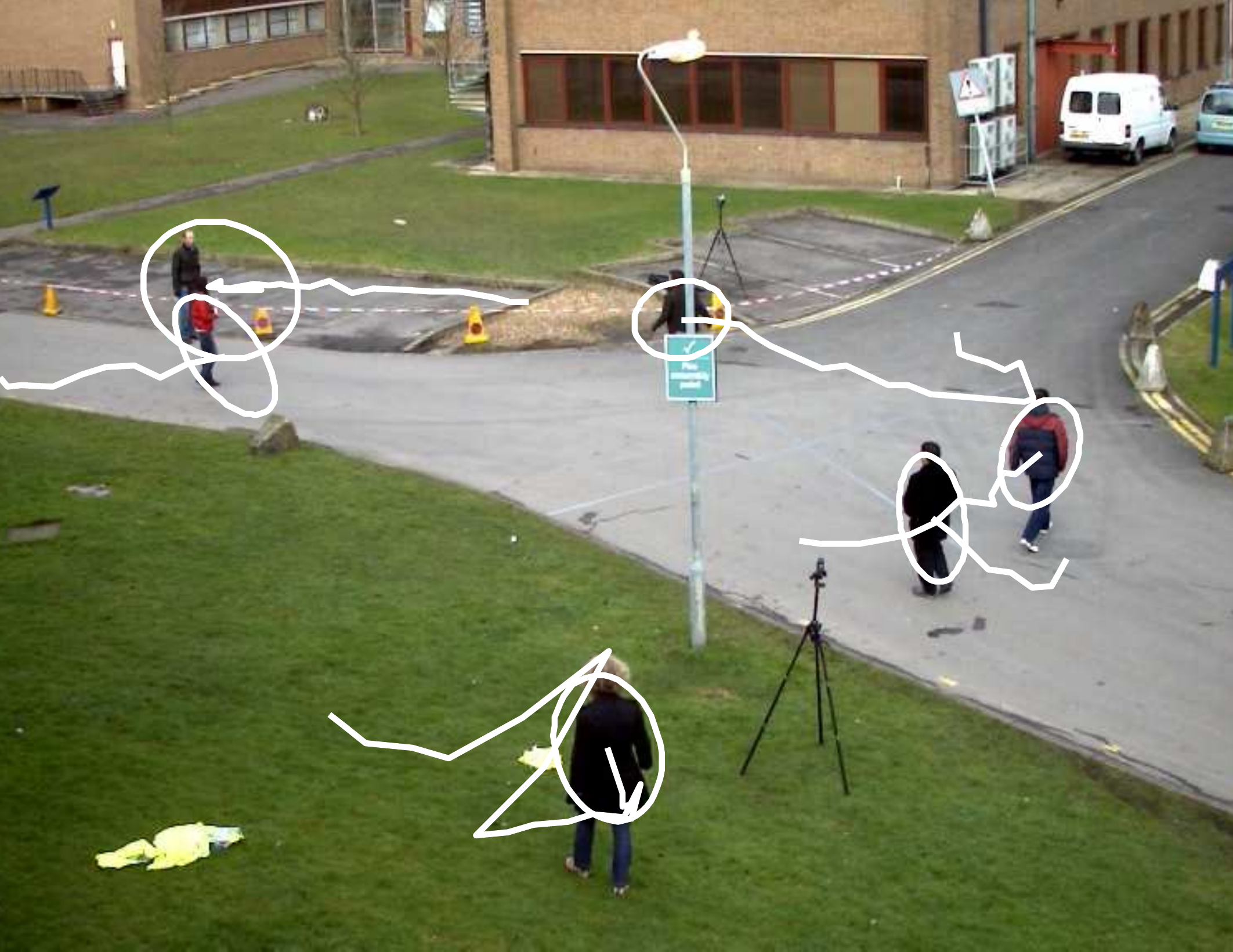}}
  \caption{ (a) Results for the PETS2009/2010 dataset, showing a sample from the posterior distribution of the state for frames 1-50, where the vertical axis denotes time, the horizontal axes represent spatial position, color represents assignment, and the mean and standard deviation are shown. (b) performance metrics vs. covariance confidence interval threshold. Below (c-f) are four frames with one posterior sample mean and covariance matrix representation shown for each frame (and one sample mean shown for the previous 20 frames).}
  \label{fig:pets2009_results}
\end{figure*}

A video dataset used in both the PETS2009 and PETS-\\2010 conferences, called ``S2.L1 at time sequence 12.34'' was chosen for experimentation due to its prominence in a number of studies \cite{ellis_2010,arsic2009multi,berclaz2009multiple,conte2010performance,bolme2009simple,breitenstein2009markovian,ge2009evaluation,alahi2009sparsity,yang2009probabilistic}. This dataset consists of a monocular, stationary camera, 794 frame video sequence. The entire video sequence was used in this experiment. 

Due to the large number of frames and objects in this video, the SMC algorithm (described in Section~\ref{sec:SMC}) was used for inference. This method of sequential inference was observed, on this dataset, to converge to a better sample in a shorter period of time in comparison with the MCMC algorithm.

Extraction was performed with frame differencing as described in Section~\ref{sec:dataextractioninexperiments}, using $L=3$, and parameters for the model were chosen to be $\alpha = 0.1, \rho = 0.8, M = 10, \boldsymbol{\mu}_{0} = (0,0), \kappa_{0} = 0.05, \nu_{0} = 6, \Lambda_{0} = \left( \begin{smallmatrix} 1&0\\ 0&1 \end{smallmatrix} \right), \text{ and } \bold{q}_{0} = (3, \ldots, 3)$. Additionally, as with the other video datasets, ground-truth bounding boxes around each object were authored using the ViPER tool.

The SMC inference algorithm yielded an estimate of the posterior distribution of the model, from which the object detection and tracking results were obtained (as described in Section~\ref{sec:inferencetotrackingresults}). In Figure \ref{fig:pets2009_results}, the MAP sample from the posterior distribution over the cluster parameters is overlayed on the extracted data over a sequence of frames, where the assignment of each data point is represented by color and marker type.

\subsubsection{Comparison with Other Methods}
\label{sec:comparisonwithothermethods}

In \cite{ellis_2010}, performance metrics (including the SFDA and ATA) were computed for a number of studies that carried out object detection and tracking for the PETS20-\\09/2010 dataset. As this dataset consists solely of humans, all ten of the algorithms presented for comparison were developed for the specific purpose of people tracking (i.e. not for general detection and tracking of arbitrary objects). As a consequence, many of these studies use externally developed (and trained) state-of-the-art human detectors, exploit the orientation of the humans in this specific dataset, or apply motion models based on assumptions about human motion. In particular, Breitenstein et al. \cite{breitenstein2009markovian} base their tracking on ouput from an externally trained human-specific detector; Yang et al. \cite{yang2009probabilistic} assume they are tracking an upright person, and perform feet and head detection; Conte et al. \cite{conte2010performance} group foreground fragments based on geometry of the human shape to be recognized and look for shadows often present in human surveillance scenarios; Berclaz et al. \cite{berclaz2009multiple} use an external detector that makes use of multiple camera views and models each human as a cylinder; Alahi et al. \cite{alahi2009sparsity} base their method on modelling the silhouettes of humans; Bolme et al. \cite{bolme2009simple} train a human specific detector; Ge et al. \cite{ge2009evaluation} provide their algorithm with estimates of typical human size and orientation; and Arsic et al. \cite{arsic2009multi} localize human feet positions.

We compare SFDA and ATA results of our strategy with these methods to show that our arbitrary object framework can yield comparable results even when compared with object-specific trackers. Table~\ref{benchmark_results_table} shows performance metric results for comparison (data published with permission from the authors of \cite{ellis_2010}). Our method achieves the fourth best SFDA and third best ATA.

\begin{table}
\begin{tabular}[!] {| l | l | l |}
  \hline
  \textsc{Method Name} & \textsc{SFDA}  & \textsc{ATA} \\ \hline \hline
  Breitenstein \cite{breitenstein2009markovian} & $ 0.57 $ & $0.30$ \\ \hline
  Yang \cite{yang2009probabilistic} & $ 0.55 $ & $0.45$ \\ \hline
  Conte \cite{conte2010performance} & $ 0.53  $ & $0.06$ \\ \hline
  \textbf{GPUDDPM} & $ \bold{0.51} $ & $\bold{0.30}$ \\ \hline
  Berclaz \cite{berclaz2009multiple} & $ 0.48 $ & $0.15$ \\ \hline
  Alahi 1 \cite{alahi2009sparsity} & $ 0.43 $ & $0.04$ \\ \hline
  Alahi 2 \cite{alahi2009sparsity} & $ 0.42 $ & $0.05$ \\ \hline
  Bolme 1 \cite{bolme2009simple} & $ 0.41 $ & NA \\ \hline
  Ge \cite{ge2009evaluation} & $ 0.38 $ & $0.04$ \\ \hline
  Bolme 2 \cite{bolme2009simple} & $ 0.34 $ & NA \\ \hline
  Arsic \cite{arsic2009multi} & $ 0.18 $ & $0.02$ \\
  \hline
\end{tabular}
\caption{SFDA and ATA performance metric results are shown for our method (in bold) and for ten other algorithms on the PETS2009/2010 benchmark dataset. Results are listed in descending order of the SFDA value. The results were provided by the authors of \cite{ellis_2010}.}
\label{benchmark_results_table}
\end{table}

\subsubsection{Sensitivity Analysis}
\label{sec:sensitivityanalysis}

SMC inference on the PETS2009/2010 video dataset was carried out for a range of the Generalized Polya Urn parameter values, $\alpha$ and $\rho$. The performance metric measures, SFDA and ATA, were computed for each combination of these two parameters. This sensitivity investigation focused on these parameters due to their potential to have a large effect on object detection accuracies. The $\alpha$ values tested included $\{ 0.01, 0.1, 1, 10, 100 \}$, and the $\rho$ values tested included $\{ 0.7, 0.75, 0.8, 0.85, 0.9 \}$.

The SFDA and ATA achieve their maximal values at different $\alpha$ and $\rho$ parameters, though both achieve a reasonably optimal value at the intermediate parameter values $\alpha = 10$ and $\rho = 0.85$. Detection and tracking performance was also shown to be fairly robust to minor variations in these parameter values.


\section{Conclusion}
\label{sec:conclusion}

We have presented a new model for the unsupervised detection and tracking of arbitrary objects in videos. The primary intention of this technique is to reduce the need for detection or localization methods tailored to specific object types and serve as a general framework applicable to videos with varied objects, backgrounds, and film qualities. The GPUDDPM, a time-dependent Dirichlet process mixture, has been introduced, and we have shown how inference on this model allows us to achieve detection and tracking results. Furthermore, we have demonstrated a specific implementation of the model using spatial and color pixel data extracted via frame differencing and provided two algorithms for performing Bayesian inference on the model to accomplish detection and tracking. Both algorithms were carried out on multiple synthetic and benchmark multi-object video datasets in order to demonstrate an ability to accomplish unsupervised detection and tracking of arbitrary objects in both manufactured and real world settings. We have described and computed standard performance metrics for our technique's detection and tracking results, and found it to be comparable with state-of-the-art object-specific detection and tracking methods designed for people tracking in the PETS2009/2010 video dataset. Results from the synthetic and benchmark video datasets illustrate the ability of the technique described in this paper to, without modification, perform completely unsupervised detection and tracking of objects with diverse physical characteristics moving over non-uniform backgrounds and through occlusion.

\begin{small}
\bibliographystyle{plainnat}
\bibliography{paper_refs} 

\begin{thebibliography}{67}
\providecommand{\natexlab}[1]{#1}
\providecommand{\url}[1]{\texttt{#1}}
\expandafter\ifx\csname urlstyle\endcsname\relax
  \providecommand{\doi}[1]{doi: #1}\else
  \providecommand{\doi}{doi: \begingroup \urlstyle{rm}\Url}\fi

\bibitem[Alahi et~al.(2009)Alahi, Jacques, Boursier, and
  Vandergheynst]{alahi2009sparsity}
A.~Alahi, L.~Jacques, Y.~Boursier, and P.~Vandergheynst.
\newblock Sparsity-driven people localization algorithm: Evaluation in crowded
  scenes environments.
\newblock In \emph{Performance Evaluation of Tracking and Surveillance
  (PETS-Winter), 2009 Twelfth IEEE International Workshop on}, pages 1--8.
  IEEE, 2009.

\bibitem[Arsic et~al.(2009)Arsic, Lyutskanov, Rigoll, and
  Kwolek]{arsic2009multi}
D.~Arsic, A.~Lyutskanov, G.~Rigoll, and B.~Kwolek.
\newblock Multi camera person tracking applying a graph-cuts based foreground
  segmentation in a homography framework.
\newblock In \emph{Performance Evaluation of Tracking and Surveillance
  (PETS-Winter), 2009 Twelfth IEEE International Workshop on}, pages 1--8.
  IEEE, 2009.

\bibitem[Beleznai et~al.(2006)Beleznai, Fruhstuck, and Bischof]{beleznai_2006}
C.~Beleznai, B.~Fruhstuck, and H.~Bischof.
\newblock Human tracking by fast mean shift mode seeking.
\newblock \emph{Journal of Multimedia}, 1\penalty0 (1):\penalty0 1--8, 2006.

\bibitem[Berclaz et~al.(2009)Berclaz, Fleuret, and Fua]{berclaz2009multiple}
J.~Berclaz, F.~Fleuret, and P.~Fua.
\newblock Multiple object tracking using flow linear programming.
\newblock In \emph{Performance Evaluation of Tracking and Surveillance
  (PETS-Winter), 2009 Twelfth IEEE International Workshop on}, pages 1--8.
  IEEE, 2009.

\bibitem[Bolme et~al.(2009)Bolme, Lui, Draper, and Beveridge]{bolme2009simple}
D.S. Bolme, Y.M. Lui, BA~Draper, and JR~Beveridge.
\newblock Simple real-time human detection using a single correlation filter.
\newblock In \emph{Performance Evaluation of Tracking and Surveillance
  (PETS-Winter), 2009 Twelfth IEEE International Workshop on}, pages 1--8.
  IEEE, 2009.

\bibitem[Breitenstein et~al.(2009{\natexlab{a}})Breitenstein, Reichlin, Leibe,
  Koller-Meier, and Van~Gool]{breitenstein2009markovian}
M.D. Breitenstein, F.~Reichlin, B.~Leibe, E.~Koller-Meier, and L.~Van~Gool.
\newblock Markovian tracking-by-detection from a single, uncalibrated camera.
\newblock \emph{Work}, 2009{\natexlab{a}}.

\bibitem[Breitenstein et~al.(2009{\natexlab{b}})Breitenstein, Reichlin, Leibe,
  Koller-Meier, and Gool]{eth_biwi_00633}
Michael~D. Breitenstein, Fabian Reichlin, Bastian Leibe, Esther Koller-Meier,
  and Luc~Van Gool.
\newblock Robust tracking-by-detection using a detector confidence particle
  filter.
\newblock In \emph{IEEE International Conference on Computer Vision}, October
  2009{\natexlab{b}}.

\bibitem[Brostow and Cipolla(2006)]{brostow2006unsupervised}
G.J. Brostow and R.~Cipolla.
\newblock Unsupervised {B}ayesian detection of independent motion in crowds.
\newblock In \emph{Computer Vision and Pattern Recognition, 2006 IEEE Computer
  Society Conference on}, volume~1, pages 594--601. IEEE, 2006.

\bibitem[Brox and Malik(2010)]{brox2010object}
T.~Brox and J.~Malik.
\newblock Object segmentation by long term analysis of point trajectories.
\newblock \emph{Computer Vision--ECCV 2010}, pages 282--295, 2010.

\bibitem[Brox et~al.(2003)Brox, Rousson, Deriche, and
  Weickert]{brox2003unsupervised}
T.~Brox, M.~Rousson, R.~Deriche, and J.~Weickert.
\newblock Unsupervised segmentation incorporating colour, texture, and motion.
\newblock In \emph{Computer Analysis of Images and Patterns}, pages 353--360.
  Springer, 2003.

\bibitem[Caron et~al.(2007)Caron, Davy, and Doucet]{caron_2007}
F.~Caron, M.~Davy, and A.~Doucet.
\newblock Generalized {P}olya urn for time-varying {D}irichlet process
  mixtures.
\newblock In \emph{23rd Conference on Uncertainty in Artificial Intelligence
  (UAI'2007), Vancouver, Canada, July 2007}, 2007.

\bibitem[Chien et~al.(2002)Chien, Ma, and Chen]{chien2002efficient}
S.Y. Chien, S.Y. Ma, and L.G. Chen.
\newblock Efficient moving object segmentation algorithm using background
  registration technique.
\newblock \emph{Circuits and Systems for Video Technology, IEEE Transactions
  on}, 12\penalty0 (7):\penalty0 577--586, 2002.

\bibitem[Choo and Fleet(2001)]{choo2001people}
Kiam Choo and D.J. Fleet.
\newblock People tracking using hybrid monte carlo filtering.
\newblock In \emph{Computer Vision, 2001. ICCV 2001. Proceedings. Eighth IEEE
  International Conference on}, volume~2, pages 321 --328 vol.2, 2001.

\bibitem[Chu et~al.(2007)Chu, Ye, Guo, and Liu]{chu_2007}
H.~Chu, S.~Ye, Q.~Guo, and X.~Liu.
\newblock Object tracking algorithm based on camshift algorithm combinating
  with difference in frame.
\newblock In \emph{Automation and Logistics, 2007 IEEE International Conference
  on}, pages 51--55. IEEE, 2007.

\bibitem[Collins(2003)]{collins2003mean}
R.T. Collins.
\newblock Mean-shift blob tracking through scale space.
\newblock In \emph{Computer Vision and Pattern Recognition, 2003. Proceedings.
  2003 IEEE Computer Society Conference on}, volume~2, pages II--234. IEEE,
  2003.

\bibitem[Comaniciu et~al.(2003)Comaniciu, Ramesh, and Meer]{comaniciu_2003}
D.~Comaniciu, V.~Ramesh, and P.~Meer.
\newblock Kernel-based object tracking.
\newblock \emph{Pattern Analysis and Machine Intelligence, IEEE Transactions
  on}, 25\penalty0 (5):\penalty0 564 -- 577, may 2003.

\bibitem[Conte et~al.(2010)Conte, Foggia, Percannella, and
  Vento]{conte2010performance}
D.~Conte, P.~Foggia, G.~Percannella, and M.~Vento.
\newblock Performance evaluation of a people tracking system on pets2009
  database.
\newblock In \emph{Advanced Video and Signal Based Surveillance (AVSS), 2010
  Seventh IEEE International Conference on}, pages 119--126. IEEE, 2010.

\bibitem[Cucchiara et~al.(2004)Cucchiara, Grana, Tardini, and
  Vezzani]{cucchiara2004probabilistic}
R.~Cucchiara, C.~Grana, G.~Tardini, and R.~Vezzani.
\newblock Probabilistic people tracking for occlusion handling.
\newblock In \emph{Pattern Recognition, 2004. ICPR 2004. Proceedings of the
  17th International Conference on}, volume~1, pages 132--135. IEEE, 2004.

\bibitem[Dockstader and Tekalp(2001)]{dockstader2001multiple}
S.L. Dockstader and A.M. Tekalp.
\newblock Multiple camera tracking of interacting and occluded human motion.
\newblock \emph{Proceedings of the IEEE}, 89\penalty0 (10):\penalty0
  1441--1455, 2001.

\bibitem[Doermann and Mihalcik(2000)]{doermann_2000}
D.~Doermann and D.~Mihalcik.
\newblock Tools and techniques for video performance evaluation.
\newblock In \emph{Pattern Recognition, 2000. Proceedings. 15th International
  Conference on}, volume~4, pages 167--170. IEEE, 2000.

\bibitem[Douc and Capp{\'e}(2005)]{douc2005comparison}
R.~Douc and O.~Capp{\'e}.
\newblock Comparison of resampling schemes for particle filtering.
\newblock In \emph{Image and Signal Processing and Analysis, 2005. ISPA 2005.
  Proceedings of the 4th International Symposium on}, pages 64--69. IEEE, 2005.

\bibitem[Ellis and Ferryman(2010)]{ellis_2010}
A.~Ellis and J.~Ferryman.
\newblock Pets2010 and pets2009 evaluation of results using individual ground
  truthed single views.
\newblock In \emph{Advanced Video and Signal Based Surveillance (AVSS), 2010
  Seventh IEEE International Conference on}, pages 135--142. IEEE, 2010.

\bibitem[Fei-Fei and Perona(2005)]{fei2005bayesian}
L.~Fei-Fei and P.~Perona.
\newblock A {B}ayesian hierarchical model for learning natural scene
  categories.
\newblock In \emph{Computer Vision and Pattern Recognition, 2005. CVPR 2005.
  IEEE Computer Society Conference on}, volume~2, pages 524--531. IEEE, 2005.

\bibitem[Ferguson(1973)]{ferguson_1973}
T.S. Ferguson.
\newblock A {B}ayesian analysis of some nonparametric problems.
\newblock \emph{The annals of statistics}, pages 209--230, 1973.

\bibitem[Fragkiadaki and Shi(2011)]{fragkiadaki2011detection}
K.~Fragkiadaki and J.~Shi.
\newblock Detection free tracking: Exploiting motion and topology for
  segmenting and tracking under entanglement.
\newblock In \emph{Computer Vision and Pattern Recognition (CVPR), 2011 IEEE
  Conference on}, pages 2073--2080. IEEE, 2011.

\bibitem[Francois(2004)]{francois2004real}
A.R. Francois.
\newblock Real-time multi-resolution blob tracking.
\newblock Technical report, DTIC Document, 2004.

\bibitem[Gasthaus et~al.(2008)Gasthaus, Wood, G\"or\"ur, and
  Teh]{gasthaus_2008}
J.~Gasthaus, F.~Wood, D.~G\"or\"ur, and Y.~W. Teh.
\newblock Dependent {D}irichlet process spike sorting.
\newblock In \emph{Advances in Neural Informations Processing Systems 22},
  2008.

\bibitem[Gasthaus(2008)]{gasthaus_thesis}
Jan Gasthaus.
\newblock Spike sorting using time-varying {D}irichlet process mixture models,
  2008.

\bibitem[Ge and Collins(2009)]{ge2009evaluation}
W.~Ge and R.T. Collins.
\newblock Evaluation of sampling-based pedestrian detection for crowd counting.
\newblock In \emph{Performance Evaluation of Tracking and Surveillance
  (PETS-Winter), 2009 Twelfth IEEE International Workshop on}, pages 1--7.
  IEEE, 2009.

\bibitem[Gelman(2004)]{gelman2004bayesian}
A.~Gelman.
\newblock \emph{Bayesian data analysis}.
\newblock CRC press, 2004.

\bibitem[Griffin and Steel(2006)]{griffin2006order}
J.E. Griffin and M.F.J. Steel.
\newblock Order-based dependent {D}irichlet processes.
\newblock \emph{Journal of the American statistical Association}, 101\penalty0
  (473):\penalty0 179--194, 2006.

\bibitem[Han et~al.(2004)Han, Sethi, Hua, and Gong]{han_2004}
M.~Han, A.~Sethi, W.~Hua, and Y.~Gong.
\newblock A detection-based multiple object tracking method.
\newblock In \emph{Image Processing, 2004. ICIP'04. 2004 International
  Conference on}, volume~5, pages 3065--3068. IEEE, 2004.

\bibitem[Hong et~al.(2007)Hong, Lee, and Chang]{hong2007real}
W.D. Hong, T.H. Lee, and P.C. Chang.
\newblock Real-time foreground segmentation for the moving camera based on h.
  264 video coding information.
\newblock In \emph{Future Generation Communication and Networking (FGCN 2007)},
  volume~1, pages 385--390. IEEE, 2007.

\bibitem[Isard and MacCormick(2001)]{isard_2001}
M.~Isard and J.~MacCormick.
\newblock Bramble: A {B}ayesian multiple-blob tracker.
\newblock In \emph{Computer Vision, 2001. ICCV 2001. Proceedings. Eighth IEEE
  International Conference on}, volume~2, pages 34--41. IEEE, 2001.

\bibitem[Jain et~al.(1997)Jain, Ratha, and Lakshmanan]{jain1997object}
A.K. Jain, N.K. Ratha, and S.~Lakshmanan.
\newblock Object detection using {G}abor filters.
\newblock \emph{Pattern Recognition}, 30\penalty0 (2):\penalty0 295--309, 1997.

\bibitem[Jepson et~al.(2003)Jepson, Fleet, and El-Maraghi]{jepson_2003}
A.D. Jepson, D.J. Fleet, and T.F. El-Maraghi.
\newblock Robust online appearance models for visual tracking.
\newblock \emph{IEEE Transactions on Pattern Analysis and Machine
  Intelligence}, pages 1296--1311, 2003.

\bibitem[Kasturi et~al.(2008)Kasturi, Goldgof, Soundararajan, Manohar,
  Garofolo, Bowers, Boonstra, Korzhova, and Zhang]{kasturi_2008}
R.~Kasturi, D.~Goldgof, P.~Soundararajan, V.~Manohar, J.~Garofolo, R.~Bowers,
  M.~Boonstra, V.~Korzhova, and J.~Zhang.
\newblock Framework for performance evaluation of face, text, and vehicle
  detection and tracking in video: Data, metrics, and protocol.
\newblock \emph{IEEE Transactions on Pattern Analysis and Machine
  Intelligence}, pages 319--336, 2008.

\bibitem[Khan et~al.(2004)Khan, Balch, and Dellaert]{khan_2004}
Z.~Khan, T.~Balch, and F.~Dellaert.
\newblock An {MCMC}-based particle filter for tracking multiple interacting
  targets.
\newblock \emph{Computer Vision-ECCV 2004}, pages 279--290, 2004.

\bibitem[Kim and Hwang(2002)]{kim2002fast}
C.~Kim and J.N. Hwang.
\newblock Fast and automatic video object segmentation and tracking for
  content-based applications.
\newblock \emph{Circuits and Systems for Video Technology, IEEE Transactions
  on}, 12\penalty0 (2):\penalty0 122--129, 2002.

\bibitem[Lee and Nevatia(2009)]{lee_2009}
S.~Lee and R.~Nevatia.
\newblock Speed performance improvement of vehicle blob tracking system.
\newblock \emph{Multimodal Technologies for Perception of Humans}, pages
  197--202, 2009.

\bibitem[Leibe et~al.(2008)Leibe, Schindler, Cornelis, and
  Van~Gool]{leibe2008coupled}
B.~Leibe, K.~Schindler, N.~Cornelis, and L.~Van~Gool.
\newblock Coupled object detection and tracking from static cameras and moving
  vehicles.
\newblock \emph{Pattern Analysis and Machine Intelligence, IEEE Transactions
  on}, 30\penalty0 (10):\penalty0 1683 --1698, Oct. 2008.

\bibitem[MacCormick and Blake(1999)]{maccormick1999probabilistic}
J.~MacCormick and A.~Blake.
\newblock A probabilistic exclusion principle for tracking multiple objects.
\newblock In \emph{Computer Vision, 1999. The Proceedings of the Seventh IEEE
  International Conference on}, volume~1, pages 572--578. IEEE, 1999.

\bibitem[McKenna et~al.(1999)McKenna, Raja, and Gong]{mckenna_1999}
S.J. McKenna, Y.~Raja, and S.~Gong.
\newblock Tracking colour objects using adaptive mixture models.
\newblock \emph{Image and Vision Computing}, 17\penalty0 (3-4):\penalty0
  225--231, 1999.

\bibitem[McKenna et~al.(2000)McKenna, Jabri, Duric, and
  Wechsler]{mckenna2000tracking}
S.J. McKenna, S.~Jabri, Z.~Duric, and H.~Wechsler.
\newblock Tracking interacting people.
\newblock In \emph{Automatic Face and Gesture Recognition, 2000. Proceedings.
  Fourth IEEE International Conference on}, pages 348--353. IEEE, 2000.

\bibitem[Munkres(1957)]{munkres_1957}
J.~Munkres.
\newblock Algorithms for the assignment and transportation problems.
\newblock \emph{Journal of the Society for Industrial and Applied Mathematics},
  pages 32--38, 1957.

\bibitem[Nedrich and Davis(2010)]{nedrich2010learning}
M.~Nedrich and J.~Davis.
\newblock Learning scene entries and exits using coherent motion regions.
\newblock \emph{Advances in Visual Computing}, pages 120--131, 2010.

\bibitem[Okuma et~al.(2004)Okuma, Taleghani, Freitas, Freitas, Little, and
  Lowe]{Okuma04aboosted}
Kenji Okuma, Ali Taleghani, Nando~De Freitas, O~De Freitas, James~J. Little,
  and David~G. Lowe.
\newblock A boosted particle filter: Multitarget detection and tracking.
\newblock In \emph{In ECCV}, pages 28--39, 2004.

\bibitem[Paragios and Deriche(2000)]{paragios2000geodesic}
N.~Paragios and R.~Deriche.
\newblock Geodesic active contours and level sets for the detection and
  tracking of moving objects.
\newblock \emph{Pattern Analysis and Machine Intelligence, IEEE Transactions
  on}, 22\penalty0 (3):\penalty0 266 --280, Mar 2000.

\bibitem[Pece(2002)]{pece_2002}
A.E.C. Pece.
\newblock Generative-model-based tracking by cluster analysis of image
  differences.
\newblock \emph{Robotics and Autonomous Systems}, 39\penalty0 (3-4):\penalty0
  181--194, 2002.

\bibitem[P{\'e}rez et~al.(2002)P{\'e}rez, Hue, Vermaak, and
  Gangnet]{perez_2002}
P.~P{\'e}rez, C.~Hue, J.~Vermaak, and M.~Gangnet.
\newblock Color-based probabilistic tracking.
\newblock \emph{Computer Vision-ECCV 2002}, pages 661--675, 2002.

\bibitem[Piccardi(2004)]{piccardi2004background}
M.~Piccardi.
\newblock Background subtraction techniques: a review.
\newblock In \emph{Systems, Man and Cybernetics, 2004 IEEE International
  Conference on}, volume~4, pages 3099--3104. IEEE, 2004.

\bibitem[Raja et~al.(1998)Raja, McKenna, and Gong]{raja_1998}
Y.~Raja, S.J. McKenna, and S.~Gong.
\newblock Tracking and segmenting people in varying lighting conditions using
  colour.
\newblock In \emph{Automatic Face and Gesture Recognition, 1998. Proceedings.
  Third IEEE International Conference on}, pages 228--233. IEEE, 1998.

\bibitem[Reilly et~al.(2010)Reilly, Idrees, and Shah]{reilly2010detection}
V.~Reilly, H.~Idrees, and M.~Shah.
\newblock Detection and tracking of large number of targets in wide area
  surveillance.
\newblock \emph{Computer Vision--ECCV 2010}, pages 186--199, 2010.

\bibitem[Senior et~al.(2006)Senior, Hampapur, Tian, Brown, Pankanti, and
  Bolle]{senior2006appearance}
A.~Senior, A.~Hampapur, Y.L. Tian, L.~Brown, S.~Pankanti, and R.~Bolle.
\newblock Appearance models for occlusion handling.
\newblock \emph{Image and Vision Computing}, 24\penalty0 (11):\penalty0
  1233--1243, 2006.

\bibitem[Sista and Kashyap(2000)]{sista2000unsupervised}
S.~Sista and R.L. Kashyap.
\newblock Unsupervised video segmentation and object tracking.
\newblock \emph{Computers in Industry}, 42\penalty0 (2-3):\penalty0 127--146,
  2000.

\bibitem[Sivic et~al.(2005)Sivic, Russell, Efros, Zisserman, and
  Freeman]{sivic2005discovering}
J.~Sivic, B.C. Russell, A.A. Efros, A.~Zisserman, and W.T. Freeman.
\newblock Discovering objects and their location in images.
\newblock In \emph{Computer Vision, 2005. ICCV 2005. Tenth IEEE International
  Conference on}, volume~1, pages 370--377. IEEE, 2005.

\bibitem[Song and Nevatia(2005)]{song2005model}
X.~Song and R.~Nevatia.
\newblock A model-based vehicle segmentation method for tracking.
\newblock In \emph{Computer Vision, 2005. ICCV 2005. Tenth IEEE International
  Conference on}, volume~2, pages 1124--1131. IEEE, 2005.

\bibitem[Stauffer(2003)]{stauffer2003estimating}
C.~Stauffer.
\newblock Estimating tracking sources and sinks.
\newblock In \emph{Computer Vision and Pattern Recognition Workshop, 2003.
  CVPRW'03. Conference on}, volume~4, pages 35--35. IEEE, 2003.

\bibitem[Taj et~al.(2007)Taj, Maggio, and Cavallaro]{taj_2007}
M.~Taj, E.~Maggio, and A.~Cavallaro.
\newblock Multi-feature graph-based object tracking.
\newblock \emph{Multimodal Technologies for Perception of Humans}, pages
  190--199, 2007.

\bibitem[Vermaak et~al.(2003)Vermaak, Doucet, and P{\'e}rez]{vermaak_2003}
J.~Vermaak, A.~Doucet, and P.~P{\'e}rez.
\newblock Maintaining multimodality through mixture tracking.
\newblock In \emph{Computer Vision, 2003. Proceedings. Ninth IEEE International
  Conference on}, pages 1110--1116. IEEE, 2003.

\bibitem[Wang(1998)]{wang1998unsupervised}
D.~Wang.
\newblock Unsupervised video segmentation based on watersheds and temporal
  tracking.
\newblock \emph{Circuits and Systems for Video Technology, IEEE Transactions
  on}, 8\penalty0 (5):\penalty0 539--546, 1998.

\bibitem[Wood and Black(2008)]{Wood2008nonparametric}
Frank Wood and Michael~J. Black.
\newblock A nonparametric bayesian alternative to spike sorting.
\newblock \emph{Journal of Neuroscience Methods}, 173\penalty0 (1):\penalty0 1
  -- 12, 2008.

\bibitem[Yang et~al.(2009)Yang, Vela, Shi, and Teizer]{yang2009probabilistic}
J.~Yang, PA~Vela, Z.~Shi, and J.~Teizer.
\newblock Probabilistic multiple people tracking through complex situations.
\newblock In \emph{11th IEEE International Workshop on Performance Evaluation
  of Tracking and Surveillance}, 2009.

\bibitem[Zhang and Lu(2001)]{zhang2001segmentation}
D.~Zhang and G.~Lu.
\newblock Segmentation of moving objects in image sequence: A review.
\newblock \emph{Circuits, systems, and signal processing}, 20\penalty0
  (2):\penalty0 143--183, 2001.

\bibitem[Zhang et~al.(2007)Zhang, Jia, Xiong, Wong, Heng, and
  Bao]{zhang2007moving}
G.~Zhang, J.~Jia, W.~Xiong, T.T. Wong, P.A. Heng, and H.~Bao.
\newblock Moving object extraction with a hand-held camera.
\newblock In \emph{Computer Vision, 2007. ICCV 2007. IEEE 11th International
  Conference on}, pages 1--8. IEEE, 2007.

\bibitem[Zhou and Tao(2003)]{zhou2003background}
Y.~Zhou and H.~Tao.
\newblock A background layer model for object tracking through occlusion.
\newblock In \emph{Computer Vision, 2003. Proceedings. Ninth IEEE International
  Conference on}, pages 1079--1085. IEEE, 2003.

\bibitem[Zhu et~al.(2006)Zhu, Xu, Huang, and Gao]{4036928}
Guangyu Zhu, Changsheng Xu, Qingming Huang, and Wen Gao.
\newblock Automatic multi-player detection and tracking in broadcast sports
  video using support vector machine and particle filter.
\newblock In \emph{Multimedia and Expo, 2006 IEEE International Conference on},
  pages 1629 --1632, july 2006.

\end{thebibliography}
\end{small}


\appendix

\newpage

\section{-  Appendix: Model Background}
\label{sec:modelbackground}

We give background on mixture models, Bayesian mixture models, the Dirichlet process, Dirichlet process (infinite) mixture models, and dependent Dirichlet process mixture models.

\subsection{Finite Mixture Model}
\label{sec:finitemixture}

A finite mixture model can be thought of as a probability distribution for an observation $x_{i}$ formulated as a linear combination of K mixture components (which we also refer to as `clusters'), where each mixture component is a probability distribution for $x_{i}$ with some parametric form, and the coefficients of the linear combination sum to one. The finite mixture model can be written as
\begin{equation}
P(x_{i}) = \sum_{k=1}^{K} P(c_{i} = k)P(x_{i}|\theta_{k})
\end{equation}
$\forall i \in \{ 1, \ldots, N \}$, where $c_{i} \in \{ 1, \ldots, K \}$ denotes the assignment of $x_{i}$ to a given mixture and $\theta_{k}$ denotes the parameters of the $k^{\text{th}}$ mixture component. Note that by choosing $P(c_{i} = k)$ as coefficients of the linear combination, it is ensured that these coefficients sum to one. We also define $p_{k} := P(c_{i} = k)$ for $k \in \{ 1, \ldots, K \} $. We can therefore write this model generatively as
\begin{align}
\begin{split}
  c_{i}|p_{1}, \ldots, p_{K}  &\sim  \text{Discrete}(p_{1}, \ldots, p_{K}) \\
  x_{i}|c_{i}, \theta_{c_{i}}  &\sim  \text{F}(\theta_{c_{i}})
\end{split}
\end{align}
$\forall i \in \{ 1, \ldots, N \}$, where the $x_{i}$ are observations, the $c_{i}$ are the mixture component assignments associated with each observation, the $\theta_{c_{i}}$ are parameters defining the $c_{i}^{\text{th}}$ mixture component (i.e. the distribution to be mixed, $\text{F}(\theta_{c_{i}})$), and the ``Discrete'' distribution refers to a multinomial distribution whose parameters are a 1-of-K vector (i.e. a vector of counts that sums to one).

\subsection{Bayesian (Finite) Mixture Model}
\label{sec:bayesianmixture}

The finite mixture model of Section~\ref{sec:finitemixture} can be extended to a Bayesian mixture model by viewing parameters that were previously point values, $\theta_{c_{i}}$ (the mixture component parameters) and $p_{1}, \ldots, p_{K}$ (the mixture component assignment weights), as random variables and providing each with a prior distribution. In this case, the prior distribution $\mathbb{G}_{0}$ is placed on the mixture component parameters, and the prior distribution $\text{Dir}(\alpha/K, \ldots, \alpha/K)$ is placed on the mixture component assignment weights. The resulting Bayesian mixture model can be formulated generatively as
\begin{align}
\begin{split}
\label{bayesian_mixture_model}
  p_{1}, \ldots, p_{K}  &\sim  \text{Dir}(\alpha/K, \ldots, \alpha/K)\\
  \theta_{1}, \ldots, \theta_{K}  &\sim  \mathbb{G}_{0} \\
  c_{i}|p_{1}, \ldots, p_{K}  &\sim  \text{Discrete}(p_{1}, \ldots, p_{K}) \\
  x_{i}|c_{i}, \theta_{c_{i}}  &\sim  \text{F}(\theta_{c_{i}})
\end{split}
\end{align}
$\forall i \in \{ 1, \ldots, N \}$, where the $x_{i}$ are observations, the $c_{i}$ are the mixture component assignments associated with each observation, the $\theta_{k}$ are parameters defining the $k^{\text{th}}$ mixture component (i.e. the distribution to be mixed, $\text{F}(\theta_{k})$), the $\theta_{k}$ are drawn from a prior distribution $\mathbb{G}_{0}$, and $p_{1}, \ldots, p_{K}$ are drawn from a Dirichlet prior parameterized by $\alpha/K, \ldots, \alpha/K$.

\subsection{Dirichlet Process}
\label{sec:dirichletprocess}

The Dirichlet process (DP), first introduced by \cite{ferguson_1973} in 1973, may be intuitively viewed as a probability distribution over discrete probability distributions. Accordingly, draws from a DP are probability mass functions (PMFs). A DP is parameterized by a base distribution $\mathbb{G}_{0}$, which is a probability distribution over a set $\Theta$, and a concentration parameter $\alpha \in \mathbb{R}_{+}$. We say that $G$ is a random PMF distributed according to a DP, written $G \sim \text{DP}(\alpha, \mathbb{G}_{0})$, if the following holds for all finite partitions $A_{1}, \ldots, A_{p}$ of $\Theta$:
\begin{equation}
(G(A_{1}), \ldots, G(A_{p})) \sim \text{Dir}(\alpha \mathbb{G}_{0}(A_{1}), \ldots, \alpha \mathbb{G}_{0}(A_{p}))
\end{equation}
Where `Dir' denotes a Dirichlet distribution. The parameters $\mathbb{G}_{0}$ and $\alpha$ may be intuitively viewed as the mean and precision of the DP. This is due to the fact that if the base distribution $\mathbb{G}_{0}$ is a distribution over $\Theta$, $A \subset \Theta$, and $G \sim \text{DP}(\alpha, \mathbb{G}_{0})$, then the following holds:
\begin{equation}
\mathbb{E}[G(A)] = \mathbb{G}_{0}(A)
\end{equation}
\begin{equation}
\text{Var}[G(A)] = \mathbb{G}_{0}(A) (1 - \mathbb{G}_{0}(A)) / (\alpha + 1)
\end{equation}
Hence, the expectation of $G(A)$ is $\mathbb{G}_{0}$, the variance of $G(A) \rightarrow 0$ as $\alpha \rightarrow \infty$, and $G$ converges pointwise to $\mathbb{G}_{0}$ when $\alpha$ is unbounded.

\subsection{Dirichlet Process (Infinite) Mixture Model}
\label{sec:dpmixture}

A DPM model, also refered to as an infinite mixture model, is an extension of the Bayesian mixture model described in Section~\ref{sec:bayesianmixture}. When using a DP as a prior in a Bayesian mixture model, $\Theta$ represents the set of parameters of the component mixture distributions. A DPM may be viewed as allowing the prior distribution over the mixture component parameters in a standard mixture model to be distributed according to a DP; this allows for modeling data where the true number of latent mixture components is unknown and arbitrarily large by letting the number of components remain unbounded (note that only a finite number of these components are assigned to the data). In particular, the DPM can be defined generatively as
\begin{align}
\begin{split}
  \mathbb{G} | \alpha, \mathbb{G}_{0}  &\sim  \text{DP}(\alpha, \mathbb{G}_{0}) \\
  \phi_{i} | \mathbb{G}  &\sim  \mathbb{G} \\
  x_{i}|\phi_{i} &\sim \text{F}(\phi_{i})
\end{split}
\end{align}
$\forall i \in \{ 1, \ldots, N \}$, where the $x_{i}$ are observations, the $\phi_{i}$ are parameters defining the mixture component from which the $i_{th}$ observation is drawn (i.e. the distribution to be mixed, $\text{F}(\phi_{i})$), and the $\phi_{i}$ are drawn from a prior distribution $\mathbb{G}$, which is in turn drawn from a DP with base distribution $\mathbb{G}_{0}$ and parameter $\alpha$. See \cite{gasthaus_2008} and \cite{gasthaus_thesis} for more details on this formulation. Note the difference between the indexing of the clusters in this model and the indexing in the previous two models. This formulation can be shown to be equivalent to the Bayesian mixture model defined in \eqref{bayesian_mixture_model}, when $K$ is taken to be unbounded; as a result, this model is sometimes called an infinite mixture model. If we let $K$ be the number of distinct mixture components assigned to observations using the above model, we can write the mixture components as $\theta_{1}, \ldots, \theta_{K}$. We also let $c_{1}, \ldots, c_{N}$ (where $c_{i} \in \{1, \ldots, K \}$) be class assignment variables that indicate the cluster to which observation $x_{i}$ is assigned.

\subsection{Dependent Dirichlet Process Mixture Model}
\label{sec:ddpmixture}

The goal of DDPM models is to allow modeling of data that is not independent and identically distributed but instead has some underlying dependencies. For example, data generated during video extraction procedures have some associated temporal dependencies, since there exist similarities between features (such as those that encode the spatial positions or appearances of objects) of data at nearby time steps.

To account for the dependent behavior of data, there has been research into models involving a sequence of DPMs, where components of the mixtures are dependent upon (or are sometimes said to be ``tied to'') corresponding components at neighboring positions in the sequence. For example, if the data shows temporal dependence, the goal might be to create a sequence of DPMs, one for each time-step, where the components of the mixture at each step are dependent upon corresponding components in the both the following and previous time steps.

More rigorously, we take the definition of a DDPM to be a stochastic process defined on the space of probability distributions over a domain, which are indexed by time, space, or a selection of other covariates in such a way that the marginal distribution at any point in the domain follows a Dirichlet process (adapted from definitions found in \cite{gasthaus_thesis} and \cite{griffin2006order}). Hence, a time-dependent DDPM is a model which remains a Dirichlet process, marginally, at each time step, yet allows cluster parameters at a given time step to vary from (and remain dependent upon) the parameters in neighboring time steps.

\section{-  Appendix: Performance Metric Details}
\label{sec:performancemetricdetails}
This section provides details on the definition and calculation of the performance evaluation metrics, SFDA and ATA, used to quantify detection and tracking results in this study.

\subsection{Mapping Ground-Truth to Output}
\label{sec:mappingtoground}

The problem of finding a mapping between a video's ground-truth tracks and an algorithm's output tracks is nontrivial, though necesary to solve, in order to compute the performance evaluation metrics used in this study. In short, the typical solution to this problem involves first specifying a performance metric and then choosing the mapping from ground-truth tracks to output tracks which yields the most favorable performance metric value. This process is described in detail by Kasturi et al. \cite{kasturi_2008}; we follow the method outlined in this paper to find an optimal mapping. Similiar to descriptions in \cite{kasturi_2008}, we implement the Hungarian algorithm \cite{munkres_1957} as a polynomial-time ($O(n^{3})$) solution to the problem of optimally mapping two sets of tracks once the similarity between any two tracks given some specified metric is established. Additionally, the method employed in \cite{kasturi_2008} allows erroneous and undetected tracks to be left unmapped, which is both desired and necessary in the case where there is a different number of ground-truth and output tracks. Note that once a mapping from a collection of ground-truth tracks to a collection of result tracks has been established, one can determine which result tracks are false positives (the result tracks to which no ground-truth track is assigned) and which ground-truth tracks are true negatives (the ground-truth tracks that are not assigned to a result track). The numbers of tracks displaying both of these failures are factors in the performance metrics used in this study.

\subsection{SFDA and ATA}

The two metrics used to quantify performance in this study are known as the the Sequence Frame Detection Accuracy (SFDA) and the Average Tracking Accuracy (ATA). These metrics were developed during VACE Phase II to provide a single, comprehensive metric to describe detection, and one to describe tracking. The following are used in the definitions of the performance metrics:
\begin{itemize}
\renewcommand{\labelitemi}{$\bullet$}
\item $G_{i}$ denotes the spatiotemporal region occupied by the $i$th ground-truth object in a video, and $G_{i}^{(t)}$ denotes the region occupied by the $i$th ground-truth object in frame $t$.
\vspace{2mm}
\item $D_{i}$ denotes the spatiotemporal region occupied by the $i$th detected object in a video, and $D_{i}^{(t)}$ denotes the region occupied by the $i$th detected object in frame $t$.
\vspace{2mm}
\item $N_{G}$ denotes the total number of unique ground-truth objects in a video, and $N_{G}^{(t)}$ denotes the number of unique ground-truth objects present at frame $t$.
\vspace{2mm}
\item $N_{D}$ denotes the total number of unique detected objects in a video, and $N_{D}^{(t)}$ denotes the number of unique detected objects present at frame $t$.
\vspace{2mm}
\item $N_{\text{frames}}$ denotes the total number of frames in a video, and $N_{\text{frames}}^{(i)}$ denotes the number of frames in which an object $i$ (which can be a ground-truth or detected object, depending on the context) is present in a video.
\vspace{2mm}
\item $N_{\text{mapped}}$ denotes the number of mapped ground-truth/ detect pairs in a video, and $N_{\text{mapped}}^{(t)}$ denotes the number of mapped ground-truth/detect pairs present at frame $t$.
\vspace{1mm}
\end{itemize}

The SFDA metric quantifies the performance of an object detection algorithm as a function of the number of correct detects, false positive detects, missed (true negative) detects, and spatial allignment of detects relative to the ground-truth. The SFDA is calculated by computing the Frame Detection Accuracy at frame $t$ ($\text{FDA}^{(t)}$) for each frame in a video sequence. The FDA provides a measure of the allignment between ground-truth and detected objects in a given frame via the overlap ratio of a ground-truth/detect pair, defined to be the ratio of the intersection of ground-truth and detect regions to the union of ground-truth and detect regions. Formally, we can write
\begin{equation}
\text{FDA}^{(t)} = \frac{\text{Overlap Ratio}}{\left(\frac{N_{G}^{(t)} + N_{D}^{(t)}}{2}\right)}
\end{equation}
where
\begin{equation}
\text{Overlap Ratio} = \sum_{i = 1}^{N_{\text{mapped}}^{(t)}} \frac{\left|G_{i}^{(t)} \cap D_{i}^{(t)}\right|}{\left|G_{i}^{(t)} \cup D_{i}^{(t)}\right|}
\end{equation}
The term $N_{\text{mapped}}^{(t)}$ refers to an optimal mapping between ground-truth and detects at frame $t$ as specified in section \ref{sec:mappingtoground} using the $\text{FDA}^{(t)}$ as the relevant metric. Given the $\text{FDA}^{(t)}$ at each frame, the SFDA can be computed; this metric may be viewed as the average FDA over all frames of a video sequence. We define
\begin{equation}
\text{SFDA} = \frac{\sum_{t=1}^{N_{\text{frames}}} \text{FDA}^{(t)}}{\sum_{t=1}^{N_{\text{frames}}} \exists \left( N_{G}^{(t)} \vee N_{D}^{(t)} \right)}
\end{equation}
where $\exists \left( N_{G}^{(t)} \vee N_{D}^{(t)} \right)$ yields a 1 if either a detected or ground-truth object is present in frame $t$ and a 0 otherwise.

The ATA metric quantifies the performance of an object tracking algorithm as a function of the spatial overlap of a mapped set of sequences of detected object positions to a set of sequences of groundtruth object positions. The ATA is calculated by first computing the Sequence Track Detection Accuracy (STDA), which can be viewed as a tracking performance measure unnormalied in terms of the number of objects. We can write the STDA as
\begin{equation}
\text{STDA} = \sum_{i=1}^{N_{\text{mapped}}} \frac{\sum_{t=1}^{N_{\text{frames}}} \left( \frac{\left|G_{i}^{(t)} \cap D_{i}^{(t)}\right|}{\left|G_{i}^{(t)} \cup D_{i}^{(t)}\right|}  \right) }{ N_{(G_{i} \cup D_{i} \neq \emptyset)} }
\end{equation}
where $N_{\text{mapped}}$ refers to an optimal mapping between ground-truth and detected objects as specified in section \ref{sec:mappingtoground} using the STDA as the relevant metric, and $N_{(G_{i} \cup D_{i} \neq \emptyset)}$ denotes the number of frames in which a given tracked object, the ground truth object to which it is mapped, or both, are present.

Given the STDA for a video sequence, the ATA can be computed by the formula
\begin{equation}
\text{ATA} = \frac{\text{STDA}}{\left( \frac{N_{G} + N_{D}}{2} \right)}
\end{equation}

\end{document}